\RequirePackage{fix-cm}
\documentclass[Afour,sageh,times]{sagej}

\usepackage{moreverb,url}
\usepackage[colorlinks,bookmarksopen,bookmarksnumbered,citecolor=red,urlcolor=red]{hyperref}

\usepackage{math}
\usepackage{amsmath}
\usepackage{balance} 
\usepackage{tabularx} 
\usepackage{multirow}

\usepackage{natbib}
\usepackage{booktabs}
\usepackage{tabularx}
\usepackage{graphicx}
\usepackage{placeins}

\usepackage{tcolorbox}
\definecolor{mycolor}{rgb}{0.811, 0.882, 0.952}
\newtcolorbox{mybox}{colback=gray!10!white,colframe=black!50!black, boxrule=0.4pt,arc=6pt, left=6pt,right=6pt,top=5pt,bottom=5pt}

\newcolumntype{Y}{>{\raggedright\arraybackslash}m{\hsize}}
\newcommand\BibTeX{{\rmfamily B\kern-.05em \textsc{i\kern-.025em b}\kern-.08em
T\kern-.1667em\lower.7ex\hbox{E}\kern-.125emX}}

\def\volumeyear{2026}

\begin{document}

\runninghead{Liang et al.}

\title{MOSU2: Long-range Navigation System with Multi-modal Scene Understanding
}

\author{Jing Liang*\affilnum{1}, Zhenyu Hou*\affilnum{2}, Senthurbavan Kirubaharan\affilnum{2}, Md Safwan Mondal\affilnum{2},  Ashok Gudivada\affilnum{2}, Xuesu Xiao\affilnum{2}, Dinesh Manocha\affilnum{3} 
}

\affiliation{\affilnum{1}Stanford University, CA, USA\\
\affilnum{2}George Mason University, VA, USA\\
\affilnum{3}University of Maryland, College Park, MD, USA \\
* indicates equal contribution
}

\corrauth{Jing Liang, Stanford Robotics Center, Computer Science Department, Stanford University, 350 Jane Stanford Way, RM122, Stanford, CA 94305, USA.}

\email{jinglms@stanford.edu}

\begin{abstract}
Robots that travel long distances through public outdoor spaces must do more than avoid obstacles: they must remain on terrain suited to their mobility and decide when to proceed, yield, or detour around pedestrians, cyclists, vehicles, traffic signals, and temporary restrictions. These decisions are difficult because their correctness often depends on how a scene evolves, such as whether an approaching car is slowing or a cyclist will cross the robot's path. Vision-language models (VLMs) can interpret such cues, but they reason over images that lack metric geometry and motion. Supplying this evidence at every step is too slow for real-time navigation, and hand-crafted triggers cannot anticipate the open-ended variety of outdoor situations. We present MOSU2, a long-range navigation system with two-stage trajectory selection, in which first-stage contextual reasoning determines when additional evidence is needed and what to retrieve. Only when an interaction remains unresolved does the first stage trigger the second stage to retrieve geometric and dynamic information and align it with scene context to make navigation decisions. Building on MOSU, MOSU2 follows routes from GPS-based route planning and filters candidate trajectories through multi-modal traversability analysis before selecting a trajectory to move the robot. In real-world experiments spanning stop gestures, construction sites, crosswalks, and roads shared with vehicles, cyclists, and pedestrians, MOSU2 raises mean navigation success from 61.8\% to 94.2\% and condition-detection accuracy from 72.2\% to 91.4\% over its single-image predecessor, at about 2.2 times its mean decision latency.
Project website: \url{https://jinggm.github.io/mosu2/}
\end{abstract}


\maketitle

\section{Introduction}
\begin{figure*}
    \includegraphics[width=\linewidth]{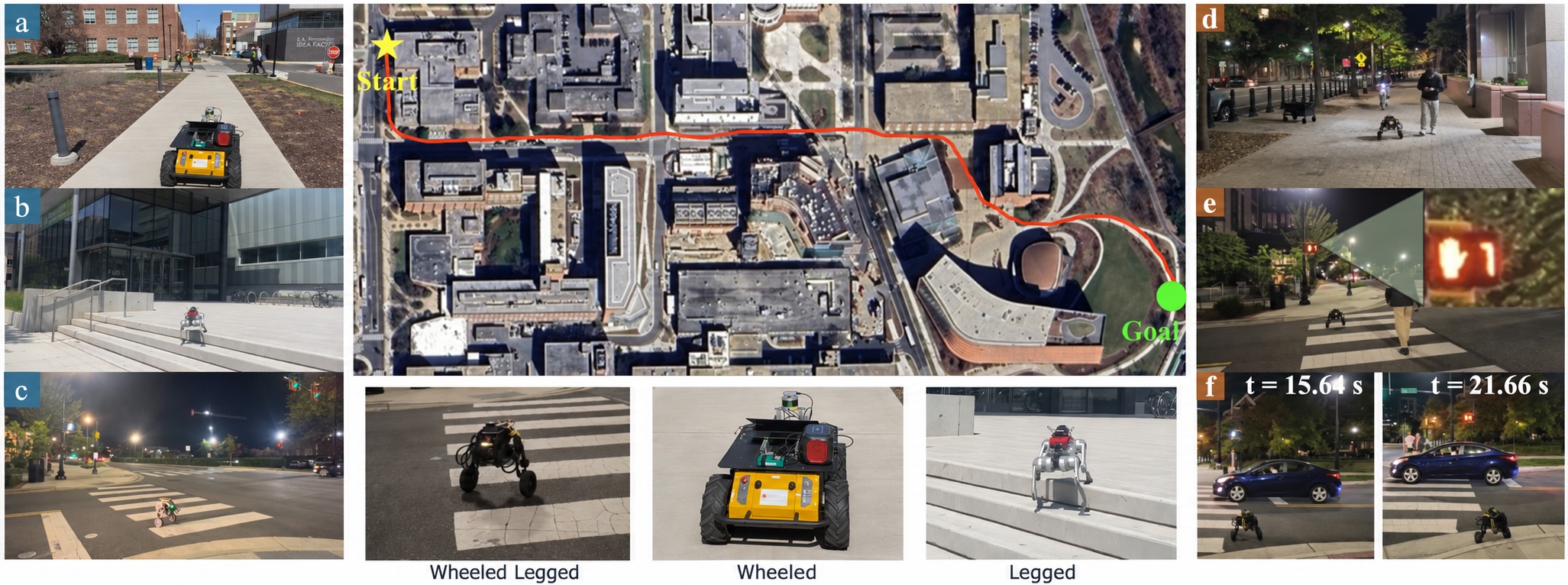}
    \caption{MOSU2 supports long-range outdoor navigation across wheeled-legged, wheeled, and legged robot platforms (bottom row). The satellite view illustrates a route from the start (yellow star) to the goal (green circle). The left-side examples highlight traversability challenges: (a) distinguishing sidewalks from off-road terrain, (b) accounting for differences in robot mobility, and (c) identifying crosswalks for street crossing. The right-side examples highlight social and traffic challenges: (d) interacting with moving agents, such as pedestrians, (e) interpreting traffic signals, including pedestrian signals with countdown displays, and (f) resolving conflicting traffic cues when vehicles violate traffic rules and obstruct the robot's path. MOSU2 combines global route guidance and robot-specific traversability assessment with two-stage reasoning that retrieves local geometric and dynamic information to interpret these conditions and select appropriate motion.
     }
    \label{fig:front}
\end{figure*}


Long-range robot navigation has advanced significantly in recent years~\citep{gao2019global, liang2025robot}. This task requires mobile robots to perceive and reason about heterogeneous environments to reach distant destinations safely, efficiently, and in a socially compliant manner~\citep{gul2019comprehensive, desouza2002vision, mavrogiannis2023core}. It plays a crucial role in applications such as logistics~\citep{alverhed2024autonomous, chen2021adoption}, service robotics~\citep{zemke2020build, berezina2019robots}, and search and rescue~\citep{davids2002urban, kalita2020exploration}.

Despite these advances, long-range navigation remains challenging in diverse and unstructured environments, where robots with different mobility capabilities~\citep{gnd, eder2023traversability, wellhausen2021rough} encounter varying terrain~\citep{graspe, terrapn}, irregular obstacle motion~\citep{wijayathunga2023challenges}, and complex traffic and social interactions~\citep{liang2025mosu, payandeh2025social, kong2025autospatial}. For example, a robot must identify terrain that is compatible with its physical capabilities and, before entering the crosswalk, determine whether an approaching vehicle is yielding. These situations require the robot to coordinate global route guidance, local traversability assessment, and contextual reasoning while accounting for the evolving behavior of surrounding road users.

Long-range navigation requires both global guidance to a distant goal and accurate, timely local scene understanding so that the robot can move safely and efficiently, as illustrated in Fig.~\ref{fig:front}. Existing navigation systems use global SLAM-based maps to support localization and path planning~\citep{gasparetto2015path, ozturk2022review}, while approaches such as ViNT combine learned visual navigation with topological graph-based planning~\citep{shah2023vint}. These representations provide useful global structure, but they must be built from prior mapping or traversal of the environment, maintaining them over large environments can be costly, and environmental changes can invalidate previously traversable routes. Moreover, global guidance does not necessarily capture the terrain conditions or temporary obstructions encountered during execution. The challenge is therefore to maintain progress toward a distant goal while reconciling global guidance with limited local observations. Our framework uses publicly available route information and GPS localization for coarse guidance, while onboard perception determines locally traversable motion without requiring a detailed prebuilt metric map.

Another challenge is to assess traversability according to the robot's morphology and motion capabilities, as illustrated by the terrain variations in Fig.~\ref{fig:front} (a-c). Geometric methods maintain local elevation maps to represent terrain shape and obstacles~\citep{elevationmap}, while semantic methods estimate traversability to distinguish areas that have similar geometry but different implications for navigation~\citep{kim2024learning}. Learning-based approaches generate local actions or candidate trajectories directly from onboard observations~\citep{liang2024mtg, shah2023vint, sridhar2024nomad}. Nevertheless, local-map reconstruction remains limited by sensor coverage and occlusion, particularly around corners or in crowded environments, while learned semantic predictions are constrained by the training distributions~\citep{kim2024learning}. Our framework combines learned trajectory generation with geometric and robot-specific semantic traversability constraints to evaluate candidate motions.

Beyond physical traversability, robots operating in shared public spaces must understand social cues and traffic conditions, as illustrated in Fig.~\ref{fig:front} (d-e). Human gestures, pedestrian signals, countdown displays, and vehicle stopping behavior convey information whose meaning depends on the surrounding situation and its evolution over time. Recent approaches use vision-language models (VLMs) to incorporate contextual knowledge into social navigation~\citep{song2024vlm, payandeh2025social}. However, methods interpreting individual images can miss the temporal evidence needed to analyze ongoing motion from changes in behavior, and qualitative descriptions do not necessarily provide reliable metric information for planning~\citep{kong2025autospatial}. Our approach grounds visual-context reasoning in geometric and dynamic information retrieved from a local map.

Another challenge is translating these complementary, and sometimes conflicting, observations into an executable navigation decision, as illustrated in Fig.~\ref{fig:front} (f). For instance, if the pedestrian signal indicates that crossing is permitted but a vehicle is slowly approaching without yielding, the robot must be able to prioritize these conflicting cues to make a safe navigational decision. Existing approaches integrate contextual information through behavioral cost maps~\citep{weerakoon2025behav} or VLM-based selection among paths generated using geometric and human-motion information~\citep{fang2026obstacles}. Nevertheless, ambiguity among concurrent activities and missing temporal context can still lead to inappropriate prioritization of social cues~\citep{fang2026obstacles}. The remaining challenge is to identify which observations matter for the current decision and reconcile their implications with physical feasibility and progress toward the goal. Our approach organizes contextual interpretation, geometric and dynamic analysis, and trajectory selection into two reasoning stages that pass along the information relevant to the navigation decision.

Our previous work, MOSU~\citep{liang2025mosu}, established a long-range navigation framework that combines coarse GPS-based routing with multimodal local scene understanding. It generates candidate trajectories from LiDAR and robot-motion observations, applies semantic traversability constraints using RGB images, and uses a VLM to assess the social appropriateness of candidate trajectories for various robot types. However, although its trajectory generator incorporates temporal sensor inputs, its VLM-based social reasoning does not explicitly associate visual cues with metric geometry and motion history. Consequently, the system has limited understanding of critical interactions that emerge over time, such as whether a vehicle or a bicycle is approaching, stopping, or yielding. In addition, its predefined prompting and trajectory-scoring pipeline lacks an explicit mechanism for selecting and retrieving additional scene information when the current semantic and geometric information is insufficient to interpret an interaction. These limitations motivate a tighter connection between contextual reasoning, measured road-user states, and navigation decisions.

To address these limitations, we present MOSU2, an extension of MOSU that combines information retrieval and two-stage reasoning for long-range navigation. MOSU2 allows VLMs to retrieve geometric and dynamic information from local maps and interpret it together with social and traffic cues from RGB observations. We also propose a two-stage reasoning architecture that coordinates this process by (1) identifying the information relevant to the current navigation decision and (2) retrieving and interpreting supporting evidence to select trajectories within MOSU's routing and traversability framework. The design leverages a key distinction between interaction detection and interpretation: a single image can flag an approaching vehicle or gesture, but deciding whether to yield requires measured trajectory and speed. MOSU2 thus resolves routine paths in a fast first stage and selectively triggers a second, deliberative stage only when metric geometric and dynamic evidence is needed. Our contributions beyond MOSU are:
\begin{itemize}
    \item We introduce a method that connects VLM-based contextual understanding with geometric and dynamic information represented in local maps. The method enables VLMs to retrieve relevant map information and relate surrounding geometry and motion to observed social and traffic cues.
    \item We develop a two-stage architecture that coordinates information gathering and navigation decisions. The first stage determines whether more information is needed for the current situation and, if so, what information is needed and where; given the request from the first stage, the second stage retrieves geometric and dynamic information about the scene according to the context and combines it with traversability assessments and route guidance to select an appropriate trajectory.
\end{itemize}

\section{Related Work}

\subsection{Long-range Outdoor Navigation and Traversability}

Long-range outdoor navigation requires robots to reconcile guidance toward a distant destination with locally feasible motion. Several approaches address this problem without requiring a detailed global metric map. ViNT~\citep{shah2023vint} learns a general-purpose visual navigation policy from heterogeneous robot datasets and combines it with topological planning for extended navigation. NoMaD~\citep{sridhar2024nomad} further unifies goal-directed navigation and exploration through a goal-conditioned diffusion policy. These approaches demonstrate that coarse global information and learned local navigation can support long-range autonomy without reconstructing the entire environment in metric detail.

Recent methods improve outdoor navigation through broader training data and longer planning horizons. CityWalker~\citep{liu2025citywalker} learns urban navigation policies from large-scale walking and driving videos, using observation and trajectory histories to predict future motion. Long Range Navigator~\citep{schmittle2025lrn} predicts promising frontier directions from egocentric images, extending the effective horizon of local navigation systems beyond their metric maps. These methods address complementary challenges in generalization and goal-directed progress. However, their navigation interfaces do not explicitly request additional metric evidence to resolve a particular social or traffic ambiguity. This distinction motivates our first reasoning stage, which uses contextual reasoning to identify when route-following requires further investigation of the surrounding scene.

Local traversability provides another essential component of outdoor navigation. ViPlanner~\citep{roth2024viplanner} combines depth and semantic observations to generate paths that account for terrain-dependent traversability, while MTG~\citep{liang2024mtg} generates diverse candidate trajectories with objectives that encourage traversability and coverage. However, physical traversability alone does not determine whether a motion is appropriate. For example, a traversable crossing may require waiting for traffic, and an accessible pathway may be temporarily restricted. Our previous MOSU framework~\citep{liang2025mosu} integrates candidate generation, geometric and semantic assessment, and visual-context evaluation with GPS-based route guidance. MOSU2 retains this navigation foundation while introducing additional reasoning over social, traffic, and temporary environmental conditions.

\subsection{Social and Traffic-aware Navigation}

Social navigation extends collision avoidance by considering how robot motion affects nearby people and whether it conforms to contextual expectations. ORCA~\citep{orca} computes reciprocal collision-avoidance actions using the relative motion of neighboring agents. Learning-based approaches, such as the attention-based crowd-navigation policy of CrowdSteer~\citep{liang2021crowd}, additionally model human-robot and human-human interactions to anticipate crowd behavior. These methods explicitly account for motion, but their state-based formulations do not directly represent the meaning of gestures, traffic signals, or temporary restrictions. 

Vision-language models provide a complementary means of interpreting such contextual information. VLM-Social-Nav~\citep{song2024vlm} uses visual-language reasoning to score socially appropriate directions and speeds for a downstream navigation system. Social-LLaVA~\citep{payandeh2025social} develops navigation-oriented visual reasoning through supervised social-scene understanding, while AutoSpatial~\citep{kong2025autospatial} improves spatial reasoning through structured grounding and hierarchical visual question answering. These approaches broaden the range of social situations that navigation systems can handle. However, understanding an interaction from an individual image remains limited when its meaning depends on dynamic information. AutoSpatial explicitly identifies temporal understanding as an open challenge, including cases in which human intentions become apparent only across successive observations. This limitation motivates combining image sequences with measured spatial and motion information.

Several methods already connect contextual reasoning with geometric planning. BehAV~\citep{weerakoon2025behav} translates language instructions into navigation and behavioral requirements, grounds the latter in a behavioral cost map, and integrates this representation with LiDAR occupancy information. \citet{fang2026obstacles} combine human-motion prediction with geometrically feasible path sampling and use a fine-tuned VLM to select a socially appropriate path. Their reported failure cases include missing visual memory, ambiguous simultaneous activities, and limitations in the available candidate paths. These studies establish that contextual reasoning and geometric or dynamic information can be integrated through several interfaces.

MOSU2 investigates a specific interface within this broader direction: the reasoning process requests relevant geometric and dynamic information from a local map and interprets the returned evidence together with consecutive RGB observations. This supports questions whose answers depend on both context and measured motion, such as whether an approaching vehicle is slowing near the intended crossing or whether a pedestrian's gesture is consistent with an available passage. Motion measurements provide supporting evidence rather than a definitive interpretation of intent. The resulting assessment guides selection among trajectories that have already passed geometric and semantic traversability evaluation.

\subsection{Agentic Robotics and Adaptive Reasoning}

Robotic planning agents increasingly use external representations and tools to connect language reasoning with observations of the physical environment. NLMap~\citep{chen2023open} constructs an open-vocabulary scene representation that allows a language-model planner to query object availability and location before generating a plan. SayPlan~\citep{rana2023sayplan} uses semantic search over hierarchical 3D scene graphs, together with conventional path planning and feedback-based replanning, to support large-scale mobile-manipulation tasks. More recently, the SuperNav preprint~\citep{zhang2026supernav} organizes navigation around a pretrained multimodal model with tool access, task-progress tracking, and context management. These approaches motivate separating information acquisition from planning while maintaining an explicit connection between the two. 


The computational cost of foundation-model reasoning has also motivated architectures that separate deliberation from rapid execution. DualVLN~\citep{wei2026dualvln} combines a VLM that predicts intermediate goals with a lightweight diffusion policy for local motion. Hydra-Nav~\citep{wang2026hydranav} adaptively switches between deliberative reasoning and reactive execution for object-goal navigation. Particularly relevant to outdoor deployment, the Walk With Me preprint~\citep{zhang2026walk} combines public-map routing, high-level VLM scene assessment, and low-level VLA trajectory generation. Its high-level model assesses whether the robot should proceed or stop and wait in situations such as crossings and crowded scenes. These methods demonstrate the relevance of hierarchical and adaptive reasoning, so the use of two stages alone does not distinguish MOSU2.

Our two-stage design instead focuses on deciding when additional evidence is required and what information should be retrieved. The first stage screens the current scene while retaining a subgoal-directed trajectory as the default choice. When further investigation is needed, the second stage examines recent images and retrieves relevant map information before selecting a trajectory and velocity mode. This design reserves the additional image-sequence analysis and map-query process for situations identified as requiring it, while allowing contextual observations to determine the investigation rather than relying exclusively on a predefined list of triggering conditions.

\section{Our Approach}

\begin{figure*}
    \includegraphics[width=\linewidth]{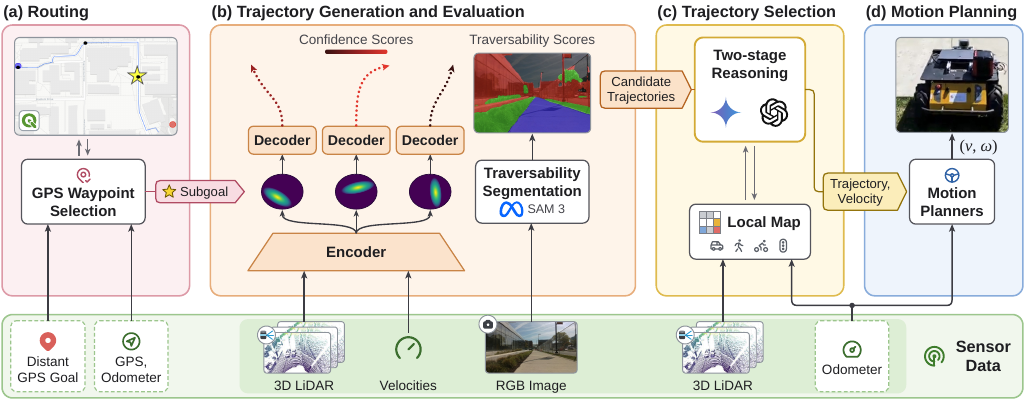}
    \caption{Overview of the MOSU2 architecture. Building on MOSU's routing and traversability framework, GPS-based routing provides subgoals, while LiDAR and robot-velocity observations support candidate trajectory generation. RGB-based semantic segmentation provides complementary traversability assessments. MOSU2 introduces a VLM-retrievable local map and two-stage trajectory selection: the first stage screens the current scene and determines if more analysis is required, and the second stage, when invoked, combines recent images with retrieved geometric and dynamic information to select a trajectory and velocity mode. The motion planner converts the selected motion into linear and angular velocity commands $(v,\omega)$.}
    \label{fig:mosu_arch}
\end{figure*}

Our approach consists of two parts: the navigation framework inherited from MOSU~\citep{liang2025mosu} and the extensions introduced in MOSU2. MOSU decomposes long-range navigation into routing and local trajectory generation and evaluation. The routing module uses OpenRouteService and GPS data to provide global guidance for long-range navigation, while trajectory generation and evaluation combine multi-modal sensor observations to generate candidate trajectories, assess traversability costs, and evaluate compliance with social and traffic rules from RGB perception. However, because MOSU does not use geometric and dynamic information about surrounding road users, it cannot determine from a single image whether a nearby pedestrian, cyclist, or vehicle is approaching, yielding, or moving away, so it may stop unnecessarily or proceed prematurely. Building on this framework, MOSU2 introduces a VLM-retrievable local map that provides geometric and dynamic information relevant to the observed scene context. Real-world environments are complicated and contain potentially hazardous conditions that cannot be easily detected by traversability analysis, such as dangerous construction sites or road closures marked only by subtle signs. To handle these challenges, we design a two-stage reasoning architecture. In the first stage, a VLM handles navigation decisions using the available observations together with MOSU's route guidance and trajectory assessments, and determines whether further investigation is needed. When additional information is required, the second stage retrieves relevant map information, coordinates visual-context interpretation with geometric and dynamic analysis, and combines the resulting evidence to select an appropriate trajectory.

\subsection{Global Direction From Routing}
We retain the routing strategy of MOSU~\citep{liang2025mosu} to provide coarse global guidance for long-range navigation. Given the robot's current GPS position and a target location, the routing module obtains a high-level route from a public routing service, such as Google Maps or OpenRouteService, and represents it as a sequence of intermediate GPS subgoals. These subgoals indicate the intended progression toward the destination, while local trajectory generation and contextual reasoning determine how the robot should move through the observed environment.

\begin{figure}
    \centering
    \includegraphics[width=\linewidth]{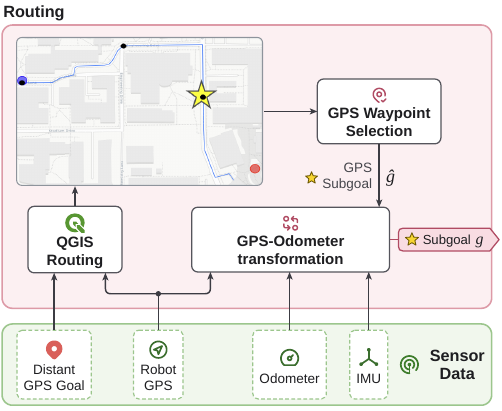}
    \caption{Routing and subgoal transformation. OpenRouteService routing uses the robot's GPS location and the destination to obtain a global route, from which an intermediate GPS waypoint $\hat{g}$ is selected. The waypoint is converted to the odometry frame with metric coordinates by aligning the GPS and robot motion, yielding the subgoal $g$. This subgoal provides directional guidance for trajectory selection.}
    \label{fig:routing_arch}
\end{figure}
Because GPS localization is subject to meter-scale errors~\citep{van2015world}, subgoal coordinates may not coincide with locally traversable terrain. We therefore use them as directional guidance rather than precise positions to be reached. Following the MOSU implementation, as shown in the Routing block of Fig.~\ref{fig:routing_arch}, consecutive subgoals are spaced approximately $35\,\mathrm{m}$ apart, a distance chosen in relation to the perceptual range of the onboard 3D LiDAR and RGB sensors. This spacing limits reliance on closely spaced GPS coordinates, while local geometric and semantic context assessments guide the robot along traversable terrain. We then transform the GPS subgoal into the robot's odometry frame using the Universal Transverse Mercator projection, and an extended Kalman filter updates this transformation. We have $g=\T \hat{g}$, where $\hat{g}$ contains the GPS latitude and longitude, $g$ is the subgoal in the odometry frame, and $\T$ is the calculated transformation matrix. 

During navigation, the active subgoal is transformed into the robot's coordinate frame using the estimated robot pose. The system monitors the approximate distance to this subgoal and advances to the next one when the distance falls below a predefined threshold of $15\,\mathrm{m}$. Route guidance specifies the desired direction of progress, while local traversability, geometric and dynamic information, and social and traffic context determine the appropriate motion toward it.


\subsection{Trajectory Generation and Evaluation}

Given the noisy subgoal from the GPS due to inherent sensor noise and the complex and unstructured outdoor environment, it is very challenging to generate an optimal trajectory to solve the navigation problem at once. Relying solely on geometric or color information is insufficient to fully understand the environment’s traversability (i.e., which areas are safe for the robot to traverse), as shown in Fig.~\ref{fig:challenge}. Therefore, our approach integrates both geometric and color information for trajectory generation and evaluation. Classical approaches perform traversability analysis of the environment to provide the constraints for trajectory optimization, but these approaches often require complete perception and are computationally slow and heavy. Learning-based approaches use end-to-end models to directly convert high-dimensional perceptual data into low-dimensional trajectories. They are faster but require large amounts of data to cover complex outdoor conditions. Our approach separates this problem into two steps: we first generate trajectories that cover the traversable area in front of the robot, and then use various models to provide constraints, evaluate the trajectories, and select the best trajectory to follow. In this way, we do not need to build an integrated map for trajectory optimization, and we can also leverage the contextual understanding of large VLMs to handle traffic and social conditions. In this work, we inherit two components from MOSU: (1) CVAE-based multiple trajectory generation, which models the distribution of feasible trajectories based on LiDAR and velocity inputs, and (2) semantic segmentation for traversability constraints, which leverages RGB images to identify navigable surfaces.

\textbf{CVAE-based Multiple Trajectory Generation:} This trajectory generation model understands the geometric information of the environment for traversability analysis through 3D LiDAR point clouds. However, directly modeling the environment using 3D point clouds is memory-intensive, and segmenting large point clouds is computationally expensive~\citep{guo2020deep, sohail2024advancing}. To address this issue, we adopt MTG~\citep{liang2024mtg}, a lightweight learning-based approach that generates multiple candidate trajectories to cover traversable regions in front of the robot. MTG employs a Conditional Variational Autoencoder (CVAE)~\citep{cvae} to model the distribution of feasible trajectories. 
Given sensor input $\x=\set{\l, \v}$, where $\l\in \mathcal{L}$ represents a sequence of LiDAR observations and $\v\in\mathcal{V}$ denotes the robot's historical velocities, the model generates a set of $N$ candidate trajectories $\mathcal{T}=\{\tau_1,...,\tau_N\}$. For each trajectory $\tau_n\in\ct$, we have the following formulation: 
\begin{align}
     &p_\theta(\tau_n|\x) = p_\theta(\tau_n | \z_n, \c), \;\;\;\z_n \sim \cn(\mu_n, \nu_n), \\
     &\c = f_\theta(\x), \;\;\; \mu, \nu = e_\theta(\x)
     \vspace{-3em}, \;\;\; \\
     &\cn(\mu_n, \nu_n) = h_\theta(\cn(\mu, \nu), \c)
     \label{eq:trajectory_generation}
\end{align}
We slightly abuse notation by using $\theta$ to represent the parameters of all neural networks. $\c$ is the conditional vector obtained from an encoder network $f_\theta(\cdot)$. $e_\theta(\cdot)$ is the latent encoder network that takes the sensor inputs $\x$ and predicts the parameters, mean $\mu$ and variance $\nu$, of a Gaussian distribution $\cn(\mu, \nu)$. In our approach, the LiDAR data is processed by PointCNN~\citep{li2018pointcnn}, and the velocities are processed by a sequence of linear layers, as shown in~\citep{liang2024mtg}. Then we linearly transform the distribution to $N$ Gaussian distributions $\cn(\mu_n, \nu_n)$ by the learnable linear transformation (neural network) $h_\theta(\cdot)$. We then sample the latent vector $\z_n$ from the distribution $\cn(\mu_n, \nu_n)$ and use it as the input to the trajectory decoder $p_\theta(\cdot)$ to generate trajectory $\tau_n$. The variance $\nu_n$ of each latent distribution represents its uncertainty and is used to compute the confidence score $c_\tau$ of trajectory $\tau_n$. This formulation takes the LiDAR geometric information as input and generates trajectories to cover geometrically traversable areas in front of the robot.

\textbf{Open-vocabulary Semantic Segmentation for Traversability Analysis:} 
However, the trajectory generation model struggles to accurately detect the boundaries between adjacent traversable and non-traversable areas when they have similar geometric structures (e.g., off-road terrain vs.\ sidewalks), as shown in Fig.~\ref{fig:challenge}(a). To address this limitation, we incorporate semantic information by using SAM3~\citep{carion2026sam}, an RGB-based semantic segmentation model with open-vocabulary queries. The model segments the image into multiple regions with different queried semantic categories by predicting the class label of each pixel. We define five traversability categories~\citep{gnd}: road, sidewalk, vegetation, building, and stairs. Different types of robots have different traversable areas: wheeled robots are restricted to the sidewalk and road categories, whereas legged robots can also traverse stairs.

Given the segmented image, we overlay the set of candidate trajectories $\mathcal{T}$, generated by the CVAE-based trajectory generator, onto the image and evaluate their traversability scores. First, we use the Bresenham algorithm~\citep{bresenham1998algorithm} to convert trajectory waypoints into connected pixels in the image. Then, the semantic traversability score $s_\tau$ is computed for each trajectory as the ratio of pixels falling in traversable areas to the total number of pixels along the trajectory.

In this work, we decouple the traversability analysis from social and traffic decision-making. This decoupling is reasonable because physical traversability determines which motions are feasible, whereas social and traffic reasoning determines which feasible motion is appropriate. Therefore, unlike the original MOSU, we use these two complementary components only to filter the candidate trajectories before contextual reasoning. For the generated $N$ trajectories, let $c_\tau^i$ denote the confidence score from geometric information and $s_\tau^i$ denote the semantic traversability score of trajectory $i\in[1,N]$. We calculate the traversability of each trajectory as

\begin{align}
    \kappa_\tau^i = \beta_1 c_\tau^i + \beta_2 s_\tau^i; \;\;\;\;\;
    \mathcal{T}_s = \{ \tau_i \in \mathcal{T} \mid \kappa_\tau^i > \kappa_{\min} \},
    \label{eq:traversability}
\end{align}
where $\beta_{1,2} > 0$ are the weights of the components, $\kappa_{\min}$ is a predefined threshold, and $\mathcal{T}_s$ represents the set of trajectories retained by traversability analysis.

\subsection{VLM-retrievable Local Map}
\label{sec:localmap}

\begin{figure}
    \centering
    \includegraphics[width=\linewidth]{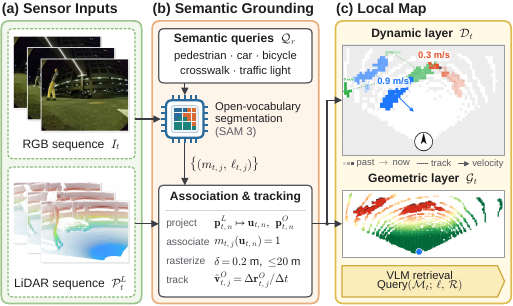}
    \caption{Construction of the VLM-retrievable local map. A VLM generates robot-specific semantic queries for the segmentation of consecutive RGB images. The corresponding LiDAR points are then used to construct a 2D local map that provides the geometric information of the semantic objects and to calculate the motion information of the objects. The constructed geometric and dynamic layers allow VLMs to retrieve relevant information by semantic category and spatial location for social context understanding.}
    \label{fig:localmap}
\end{figure}

We construct a local map that associates open-vocabulary semantic information with the geometry and motion of the surrounding environment, as illustrated in Fig.~\ref{fig:localmap}. The map represents traffic and social components in the environment, including crosswalks and traffic lights, together with the positions and velocities of nearby cars, pedestrians, and bicycles. These attributes provide evidence for social and traffic decisions beyond the physical traversability assessments described above. By exposing the traffic and social information through queryable function interfaces, the map allows VLMs to retrieve relevant environmental measurements and interpret them together with visual context.

At the beginning of each experiment, a VLM generates a set of semantic queries, $\mathcal Q_r$, based on the robot's type and mobility capabilities. The query set is not fixed during a run; during navigation, the VLM can add query terms according to the first-stage request. Given an RGB image $I_t$, the open-vocabulary segmentation model produces
\begin{equation}
\begin{aligned}
    \mathcal S_t
    &= F_{\mathrm{seg}}(I_t;\mathcal Q_r)
     = \{(m_{t,j},\ell_{t,j})\}_{j=1}^{J_t},
\end{aligned}
\label{eq:map_semantic_queries}
\end{equation}
where $m_{t,j}$ is the binary mask of segment $j$, $\ell_{t,j}$ is its semantic label, and $J_t$ is the number of detected segments. We use SAM3~\citep{carion2026sam}, consistent with the preceding traversability analysis. The queries include traffic-related concepts such as  ``traffic light,'' ``car,'' ``pedestrian,'' and ``bicycle,'' together with additional concepts relevant to the robot platform. 

To recover metric geometry, we associate each segmented image with the synchronized LiDAR point cloud $\mathcal P_t^L=\{\mathbf p_{t,n}^L\}$, expressed in the LiDAR frame $L$. We project the points onto the image plane to get the distances of the segments. Let $\mathbf K$ denote the camera intrinsic matrix, and let $(\mathbf R_{CL},\mathbf b_{CL})$ denote the calibrated transformation from the LiDAR frame to the camera frame. A LiDAR point is projected into the rectified image as
\begin{equation}
    \mathbf u_{t,n}
    =
    \pi\!\left(
    \mathbf K
    \left(
    \mathbf R_{CL}\mathbf p_{t,n}^L+\mathbf b_{CL}
    \right)
    \right),
\label{eq:lidar_image_projection}
\end{equation}
where $\pi([x,y,z]^\top)=[x/z,y/z]^\top$. Points with valid positive depth and projections within the image are associated with the masks containing their projected pixels.

The associated points are then transformed into the odometry frame $O$:
\begin{equation}
    \mathbf p_{t,n}^O
    =
    \mathbf R_{OL,t}\mathbf p_{t,n}^L+\mathbf b_{OL,t},
\label{eq:lidar_odometry_projection}
\end{equation}
where $(\mathbf R_{OL,t},\mathbf b_{OL,t})$ is obtained from the robot's odometry pose and the calibrated LiDAR mounting transformation. Expressing consecutive observations in the same odometry frame accounts for the robot's motion when comparing segment positions. For each segment, its associated point set and representative position are
\begin{equation}
\begin{aligned}
    \mathcal P_{t,j}^O
    &=
    \left\{
    \mathbf p_{t,n}^O
    \mid m_{t,j}(\mathbf u_{t,n})=1
    \right\},\\
    \mathbf r_{t,j}^O
    &=
    \frac{1}{|\mathcal P_{t,j}^O|}
    \sum_{\mathbf p\in\mathcal P_{t,j}^O}\mathbf p.
\end{aligned}
\label{eq:segment_geometry}
\end{equation}
Here, $\mathbf r_{t,j}^O$ is the centroid of the observed segment points. The full point set represents its observed spatial extent. Segments without valid LiDAR support do not receive a metric position estimate.

We maintain observations within a $20\,\mathrm{m}$ radius around the robot, and discretize them into a local grid with resolution $\delta=0.2\,\mathrm{m}$. Let $\mathbf o_t^O\in\mathbb R^2$ denote the grid origin in the odometry frame. The cells associated with segment $j$ are
\begin{equation}
    \Omega_{t,j}
    =
    \left\{
    \left\lfloor
    \frac{\mathbf p_{xy}^O-\mathbf o_t^O}{\delta}
    \right\rfloor
    \;\middle|\;
    \mathbf p^O\in\mathcal P_{t,j}^O
    \right\},
\label{eq:segment_grid_projection}
\end{equation}
where $\mathbf p_{xy}^O$ contains the horizontal ($xy$) coordinates. This projection preserves the observed layout of an extended region, instead of representing it only by its centroid.

Motion information is estimated from the displacement of corresponding segments across consecutive observations. Let $a_t(j)$ identify the segment in the previous frame associated with segment $j$ at time $t$. We associate each segment with the nearest segment of the same semantic class. Given an observation interval $\Delta t$, the object velocity estimate is
\begin{equation}
    \hat{\mathbf v}_{t,j}^O
    =
    \frac{
    (\mathbf r_{t,j}^O)_{xy}
    -
    (\mathbf r_{t-1,a_t(j)}^O)_{xy}
    }{\Delta t}.
\label{eq:segment_velocity}
\end{equation}
A velocity estimate is provided only when a valid temporal correspondence and sufficient geometric observations are available; otherwise, velocity is marked as unknown.

The semantic and motion attributes are organized into geometric and dynamic layers,
\begin{equation}
\begin{aligned}
    \mathcal G_t(\mathbf a)
    &=
    \left\{
    (j,\ell_{t,j},\mathbf r_{t,j}^O)
    \mid \mathbf a\in\Omega_{t,j}
    \right\},\\
    \mathcal D_t(\mathbf a)
    &=
    \left\{
    (j,\hat{\mathbf v}_{t,j}^O)
    \mid \mathbf a\in\Omega_{t,j}
    \right\},
\end{aligned}
\label{eq:semantic_dynamic_layers}
\end{equation}
where $\mathbf a$ denotes a grid cell. These layers associate semantic regions with their measured geometry and estimated motion.

Finally, we expose the map through function interfaces that support retrieval by semantic type, spatial location, or their combination. For a requested semantic label $\ell$ and a spatial region $\mathcal R$ expressed as a set of grid cells, a combined query is written as
\begin{equation}
\begin{aligned}
    &\operatorname{Query}(\mathcal M_t;\ell,\mathcal R)
    =
    \left\{
    (\mathbf r_{t,j}^O,\hat{\mathbf v}_{t,j}^O)
    \;\middle|\;
    \ell_{t,j}=\ell,\;
    \Omega_{t,j}\cap\mathcal R\neq\emptyset
    \right\},
\end{aligned}
\label{eq:map_query}
\end{equation}
where $\mathcal M_t$ denotes the local map with its position, semantic, and dynamic information. We then convert the query results to the robot's frame. The returned measurements provide evidence for the subsequent two-stage reasoning process, which combines them with visual context and route guidance to select a trajectory from $\mathcal T_s$. 

\subsection{Two-stage Trajectory Selection}

Given the traversable candidate set $\mathcal T_s$, we design a two-stage trajectory selection process to balance the computational cost and latency of VLM-based reasoning with the need to recognize context-dependent navigation requirements that call for additional geometric and dynamic features. VLMs can interpret social and traffic cues using contextual knowledge, but detailed analysis of image sequences and geometric/dynamic information retrieval can introduce substantial computation and latency. Meanwhile, manually specified detection rules may not cover the diverse interactions and unfamiliar conditions encountered during outdoor navigation. We therefore use a VLM in the first stage with a single image input to examine the current observation and identify whether additional geometric and dynamic information is needed, allowing the observed context to guide further investigation. When the available information is sufficient, the system selects a trajectory from $\mathcal T_s$ according to the routed subgoal. Otherwise, the second stage analyzes consecutive RGB observations and retrieves relevant geometric and dynamic information from the local map to support trajectory selection and provide a velocity mode. This architecture reserves detailed reasoning for situations identified as requiring further investigation, without requiring every triggering condition to be explicitly specified in advance.

In the first stage, candidate trajectories are projected onto the image plane and labeled with numbers from right to left~\citep{liang2025mosu}. Then the projected trajectories are evaluated according to their progress toward the subgoal. Let $\mathbf g_t^B$ denote the routed subgoal expressed in the robot's current coordinate frame, and let $\mathbf x_{i,\mathrm{end}}^B$ denote the endpoint of candidate trajectory $\tau_i$ in the same frame. For a nonempty candidate set, the default trajectory is selected as
\begin{equation}
\begin{aligned}
    \tau_t^{(1)}
    &=\underset{\tau_i\in\mathcal T_s}{\arg\min} \left\|
    \mathbf x_{i,\mathrm{end}}^B-\mathbf g_t^B
    \right\|_2^2.
\end{aligned}
\label{eq:first_stage_goal_selection}
\end{equation}
This selection uses the geometric and semantic traversability assessments already incorporated into $\mathcal T_s$. Besides the selected $\tau_t^{(1)}$, the model also outputs a suggested velocity mode, [normal speed, slow speed, stop]~\citep{liang2025mosu}.

Before executing the selected trajectory, a VLM examines the current RGB observation together with the subgoal and candidate trajectories to determine whether additional scene information is needed:
\begin{equation}
    (b_t,\mathcal U_t)
    =
    F_{\mathrm{screen}}
    \left(I_t,\mathbf g_t^B,\mathcal T_s\right),
\label{eq:reasoning_trigger}
\end{equation}
where $b_t\in\{0,1\}$ indicates whether to invoke the second stage, and $\mathcal U_t$ describes the information to be investigated. For example, the first stage may identify a pedestrian interaction, a traffic signal relevant to the route ahead, or a temporary restriction that is not captured by physical traversability alone. The resulting request specifies the relevant scene elements and the unresolved question, such as whether the pedestrian in front is blocking the road or whether the bicycles or cars in traffic on the left are yielding.  When no further investigation is requested, the system selects $\tau_t^{(1)}$.

When $b_t=1$, the second stage receives the information requests and a sequence of the $H$ most recent RGB observations, $\mathcal I_t^H$.
The image sequence provides temporal context for interpreting actions and changes in the scene. Based on $\mathcal U_t$ and these observations, the second stage selects semantic and spatial queries for the local map. Let $\mathcal Q_t^{\mathrm{map}}$ denote the requested combinations of semantic labels and spatial regions. The retrieved evidence is
\begin{equation}
    \mathcal Z_t
    =
    \bigcup_{(\ell,\mathcal R)\in\mathcal Q_t^{\mathrm{map}}}
    \operatorname{Query}
    \left(\mathcal M_t;\ell,\mathcal R\right).
\label{eq:selection_map_retrieval}
\end{equation}

For social interactions, the second stage interprets the consecutive images together with the retrieved positions and velocities of relevant pedestrians, cyclists, or vehicles. The images provide evidence about gestures, activities, and interaction context, while the map provides metric information about surrounding geometry and motion. Their combination supports distinctions that affect trajectory and velocity selection, such as whether a nearby road user is continuing toward the robot's path, moving away, or slowing near a crossing. These observations inform the interpretation of an interaction without treating motion alone as definitive evidence of intent.
For traffic conditions, the second stage additionally interprets the state of the relevant traffic signal and its relationship to the intended crossing. Signal indications and any visible countdown information are considered together with the crossing geometry and the motion of nearby road users. The map supplies the spatial relationship of surrounding traffic, while the RGB observations provide the visual information needed to interpret the signal state.
For irregular conditions, the second stage examines scene details that may change whether an otherwise traversable trajectory is appropriate. Examples include temporary barriers, construction activities, warning signs, and access restrictions. It uses the image sequence to interpret these cues and the retrieved geometry to locate the affected region relative to the candidate trajectories. A scene may involve several of these conditions simultaneously, in which case their evidence is considered jointly.

The second stage combines the retrieved evidence, visual context, and subgoal information to select the final motion:
\begin{equation}
    \tau_t, \gamma_t
    =
    F_{\mathrm{select}}
    \left(
    \mathcal T_s,
    \mathbf g_t^B,
    \mathcal I_t^H,
    \mathcal Z_t,
    \mathcal U_t
    \right),
\label{eq:second_stage_selection}
\end{equation}
where $\gamma_t$ indicates the velocity mode: stop, slow, or normal speed. The selection considers progress toward the subgoal together with the social, traffic, and contextual implications of each candidate. The selected trajectory and velocity mode are passed to the motion controller, as shown in Fig.~\ref{fig:mosu_arch}, and the selection process is repeated as new observations become available.

\section{Experiments}
We evaluate the navigation capabilities inherited from MOSU and the improvements introduced in MOSU2 through two groups of experiments. The first evaluates MOSU's traversability assessment and the ability of a single VLM to interpret social and traffic cues from an RGB image when ranking candidate trajectories. We retain selected results from the conference paper~\citep{liang2025mosu} to establish the performance of this underlying navigation framework. The second compares MOSU2 with MOSU in real-world social, traffic, and irregular scenarios to examine how MOSU2 affects navigation decisions and the recognition of situations that require a specific response. Together, these experiments assess navigation performance, contextual understanding, and time cost of reasoning.

\subsection{Experimental Setup}
We conduct experiments on three robot platforms, as shown in Fig.~\ref{fig:front}: Husky, Unitree Go2, and Diablo robots, representing wheeled, legged, and wheeled-legged robots, respectively. Each platform is equipped with calibrated RGB cameras and 3D LiDAR for perceiving the surrounding environment, as well as GPS for global guidance. 

Experiments are performed on a computer with an Intel i9 CPU (31 GB RAM) and an Nvidia RTX 3060 GPU (6.4 GB VRAM). We use ChatGPT-realtime~\citep{wu2023brief} or Gemini~\citep{GeminiTeam2024} for visual-context interpretation and trajectory evaluation, with inference performed in the cloud through API access. The trajectory generator and motion planner operate at $20\,\mathrm{Hz}$ and $50\,\mathrm{Hz}$, respectively. For MOSU2, the local map is updated at $2\,\mathrm{Hz}$ within a $20\,\mathrm{m}$ radius around the robot.

MOSU2 uses the conditional two-stage procedure described above. In the first stage, as in MOSU, the input includes the current RGB observation, the candidate trajectories, and the subgoal in the robot frame. If further investigation is required ($b_t=1$), the second stage receives from the first stage the names of the semantic items and the type of information needed. According to the request $\mathcal U_t$, it retrieves $H=5$ consecutive RGB observations sampled at $0.2\,\mathrm{s}$ intervals and spanning $1.0\,\mathrm{s}$, together with the requested geometric and dynamic information about these items. The second stage then outputs a trajectory number and a velocity mode. If no further investigation is required ($b_t=0$), the first stage directly outputs the trajectory and velocity mode without invoking the second stage.



\subsection{Evaluation of MOSU}
The experiment is designed to evaluate the efficacy of the system in two aspects: (1) the performance of each component, including point-cloud-based trajectory generation, semantic traversability analysis, and VLM's social understanding capability; (2) Overall MOSU system performance in real-world long-range navigation compared with other approaches. 

We first summarize the trajectory selection mechanism used in MOSU~\citep{liang2025mosu}, which provides the baseline for evaluating the extensions introduced in MOSU2. MOSU combines geometric trajectory generation, semantic traversability assessment, and contextual evaluation by a single VLM. At each decision step, candidate trajectories are projected onto the current RGB image and assigned numerical labels according to their endpoint positions in the image. The VLM receives this annotated image together with the relative position of the current subgoal and navigation instructions concerning pedestrian interactions, traffic rules, and robot-specific traversability. It returns a ranked list of trajectories and a velocity mode based on the observed scene.

The VLM ranking is converted into a score $r_\tau^i$, with higher-ranked trajectories receiving larger scores. MOSU then selects a trajectory by combining this score with geometric confidence ($c_\tau^i$), semantic traversability ($s_\tau^i$), and progress toward the subgoal ($g_\tau^i$):
\begin{equation}
\tau^{*}
=
\underset{\tau_i\in\mathcal{T}}{\arg\max}
\left(
\alpha_c c_\tau^i
+ \alpha_s s_\tau^i
+ \alpha_r r_\tau^i
+ \alpha_g g_\tau^i
\right),
\label{eq:mosu_selection}
\end{equation}
where $\mathcal{T}$ denotes the candidate trajectory set, $c_\tau^i$ is the geometric confidence score, $s_\tau^i$ is the semantic traversability score, and $g_\tau^i$ favors trajectories approaching the current subgoal. The coefficients $\alpha_c$, $\alpha_s$, $\alpha_r$, and $\alpha_g$ weight the respective components. The selected trajectory and velocity mode are passed to the motion controller. This mechanism uses a single RGB image for VLM-based contextual evaluation, without the map retrieval or conditional second-stage reasoning introduced in MOSU2.

\begin{figure*}
  \centering
  \begin{tabular}{cccc}
  \begin{minipage}{.23\linewidth}
   \centering
   \includegraphics[width=\linewidth,height=0.8\linewidth]{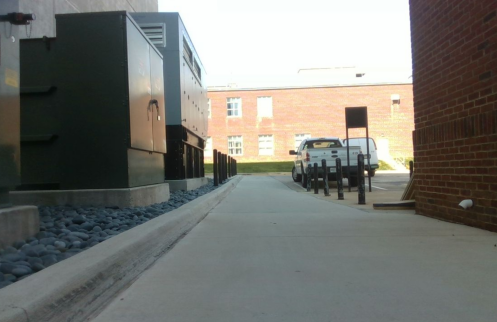} (a) Narrow Space
   \end{minipage}
   &
   \begin{minipage}{.23\linewidth}    \centering
\includegraphics[width=\linewidth,height=0.8\linewidth]{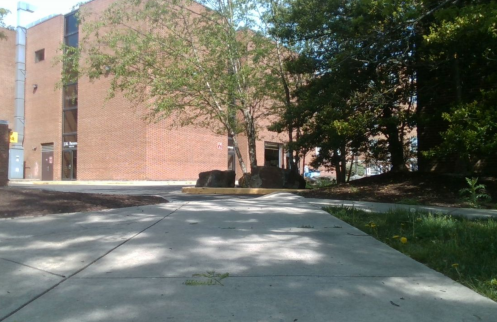}  (b) Off-road Terrain
   \end{minipage} 
   & 
   \begin{minipage}{.23\linewidth}   \centering

   \includegraphics[width=\linewidth,height=0.8\linewidth]{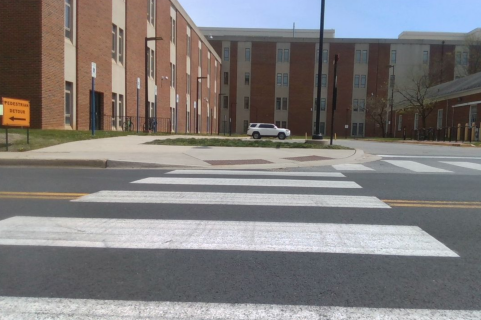} (c) Crosswalks
   \end{minipage}
   & 
   \begin{minipage}{.23\linewidth}    \centering
\includegraphics[width=\linewidth,height=0.8\linewidth]{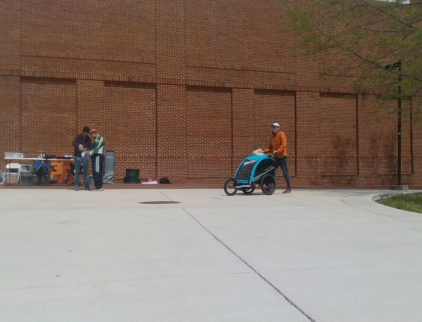} (d) Social Scenarios
\end{minipage}
  \end{tabular}
  \caption{\textbf{Complex Outdoor Scenarios:} We evaluate the approach in large-scale, complex outdoor environments with various challenging scenarios, such as (a) narrow spaces, (b) areas with dense off-road vegetation, (c) crosswalks, and (d) social situations involving pedestrians.}
    \label{fig:scenarios}
\end{figure*}

We first qualitatively evaluate each component, including traversability analysis using geometric trajectory generation, semantic information, and VLM-based social and traversability understanding of the environment, and then we quantitatively evaluate whether this combination produces traversable trajectories that approach the destination. The quantitative evaluation uses the GND dataset~\citep{gnd}, which contains observations from ten campuses spanning urban and rural environments. The evaluated scenarios include narrow passages, off-road terrain, crosswalks, and pedestrian interactions, as illustrated in Fig.~\ref{fig:scenarios}. On this dataset we compare MOSU against MTG~\citep{liang2024mtg}, DTG~\citep{liang2024dtg}, and VL-TGS~\citep{song2024tgs}, NoMaD~\citep{sridhar2024nomad}, ViNT~\citep{shah2023vint}, and PIVOT~\citep{nasiriany2024pivot}. The results retained here use the original MOSU configuration, including Gemini as the VLM backend.

The metrics in Table~\ref{tab:gnd_quantitative} are calculated as follows:
\begin{itemize}
    \item \textbf{Traversability:} We overlay the generated trajectory onto the traversability map from the GND dataset~\citep{gnd}. The traversability score is then calculated as the percentage of fully traversable waypoints over the entire trajectory length.
    \item \textbf{Distance to Target: } $1 - \frac{d_\tau - d_o}{|\tau|}$, where $d_\tau$ and $d_o$ represent the distances between the subgoal and the current trajectory and between the subgoal and the optimal ground-truth trajectory, respectively. $|\tau|$ denotes the length of the generated trajectory.
\end{itemize}



\begin{figure*}
  \centering
  \begin{tabular}{cccc}
  \begin{minipage}{.23\linewidth}
   \centering
   \includegraphics[width=\linewidth,height=0.8\linewidth]{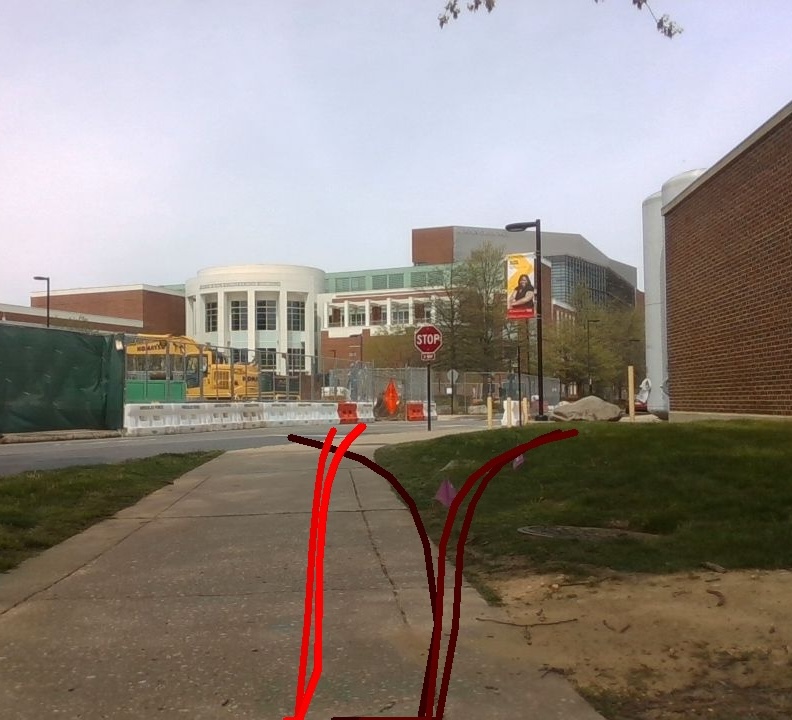} (a)
   \end{minipage}
   &
   \begin{minipage}{.23\linewidth}    \centering
\includegraphics[width=\linewidth,height=0.8\linewidth]{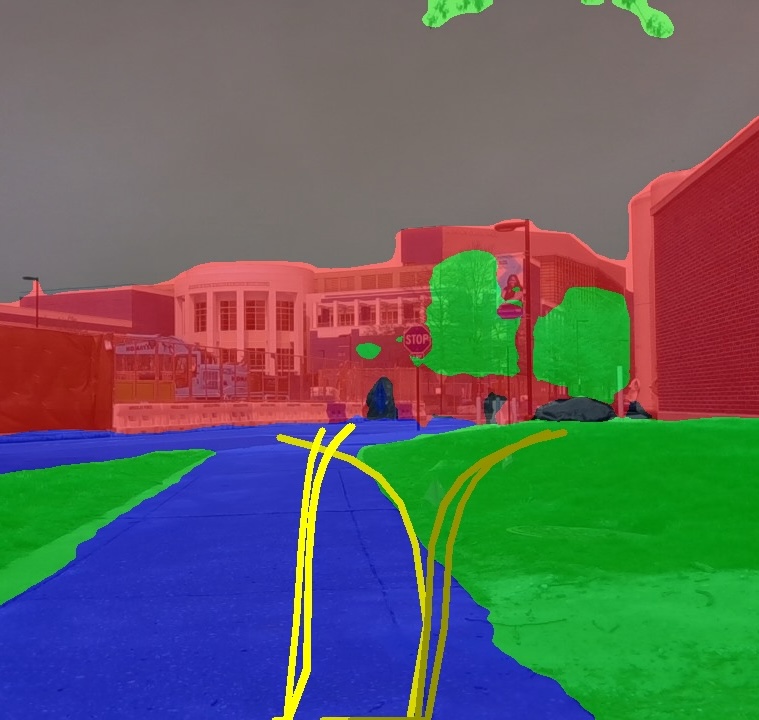}  (b)
   \end{minipage} 
   & 
   \begin{minipage}{.23\linewidth}   \centering

   \includegraphics[width=\linewidth,height=0.8\linewidth]{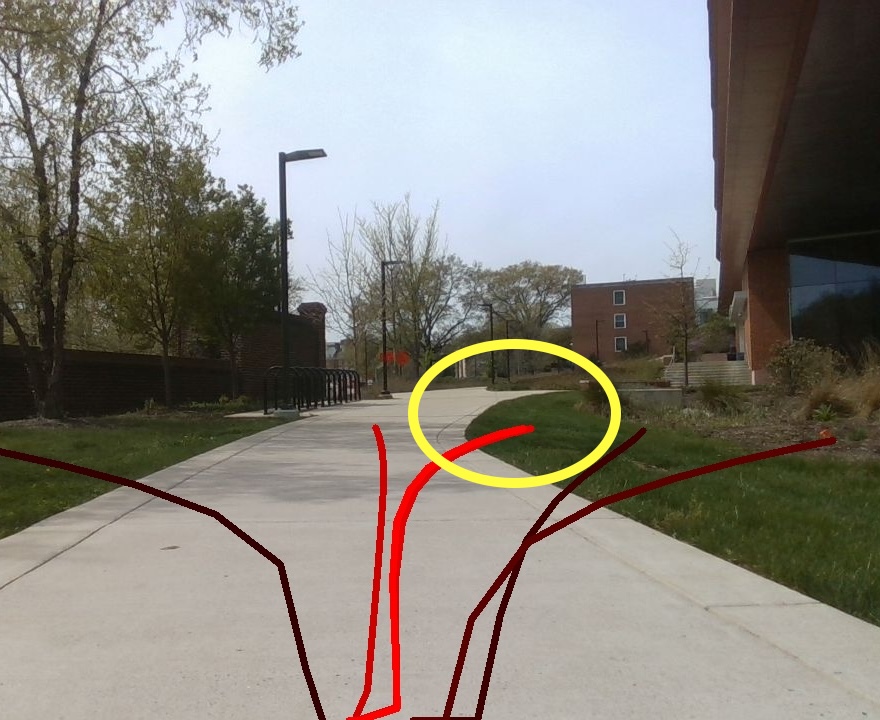} (c)
   \end{minipage}
   & 
   \begin{minipage}{.23\linewidth}    \centering
\includegraphics[width=\linewidth,height=0.8\linewidth]{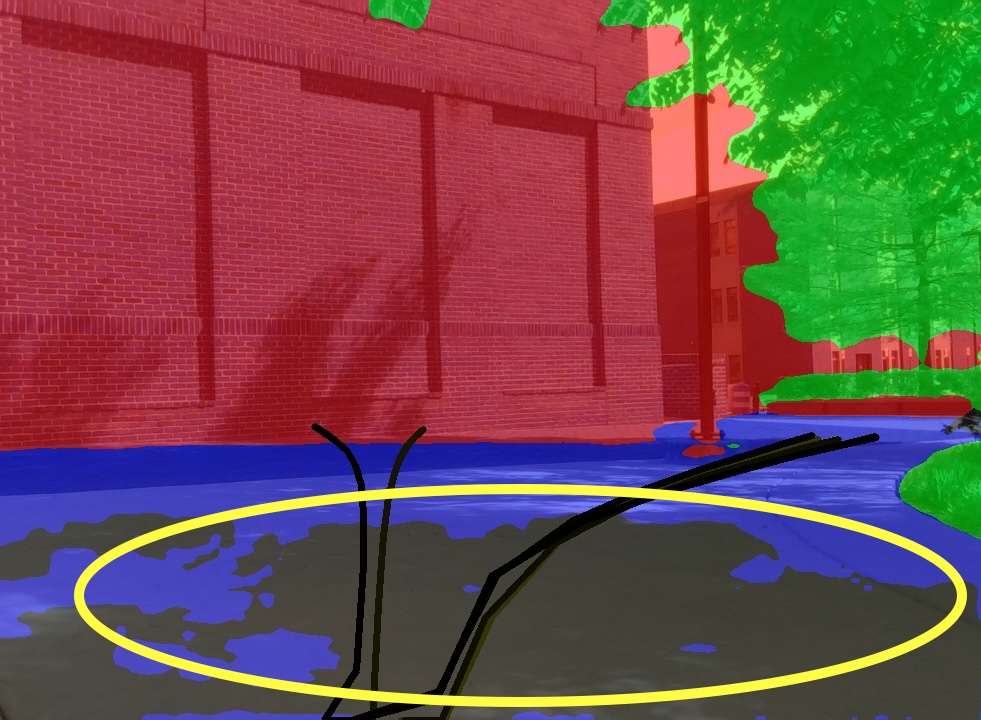} (d)\end{minipage}
  \end{tabular}
  \caption{\textbf{Geometric and Semantic Analysis:} Both geometric trajectory generation and semantic segmentation perform well in most cases, but there are also failure cases, as marked in yellow circles in (c) and (d). When the elevations of the lawn and sidewalk are similar, the geometric model struggles to perform accurately. Additionally, segmentation fails in out-of-distribution scenarios.}
    \label{fig:challenge}
\end{figure*}

\begin{figure*}[!h]

  \centering
  \begin{tabular}{ccc}
   \begin{minipage}{.3\linewidth}
   \includegraphics[width=\linewidth]{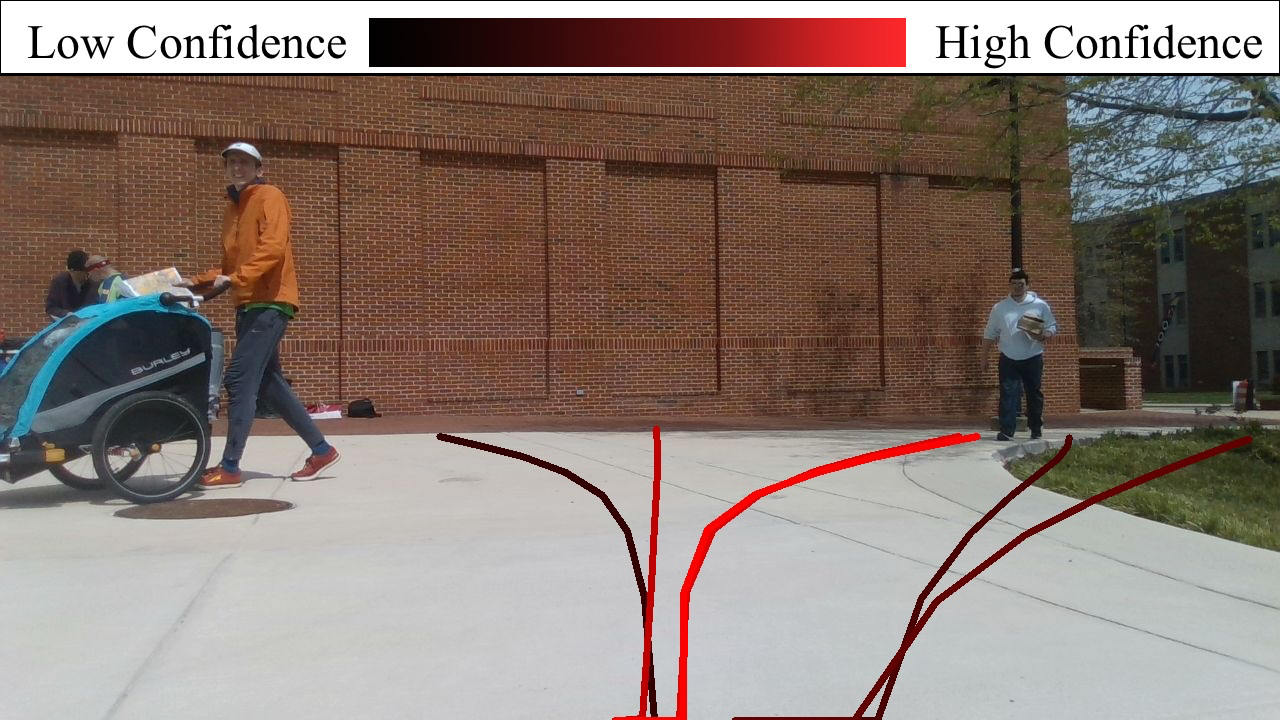}
   \end{minipage}
   & 
   \begin{minipage}{.3\linewidth} \includegraphics[width=\linewidth]{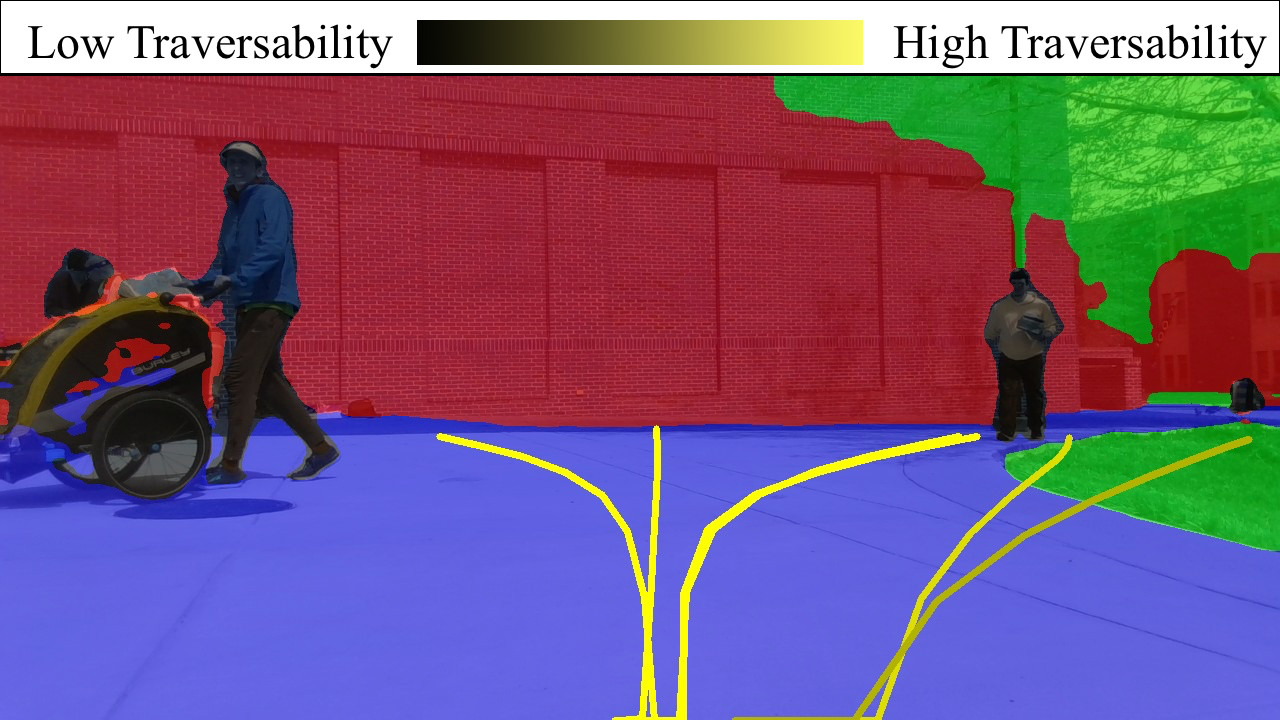} \end{minipage}
   & 
   \begin{minipage}{.3\linewidth} \includegraphics[width=\linewidth]{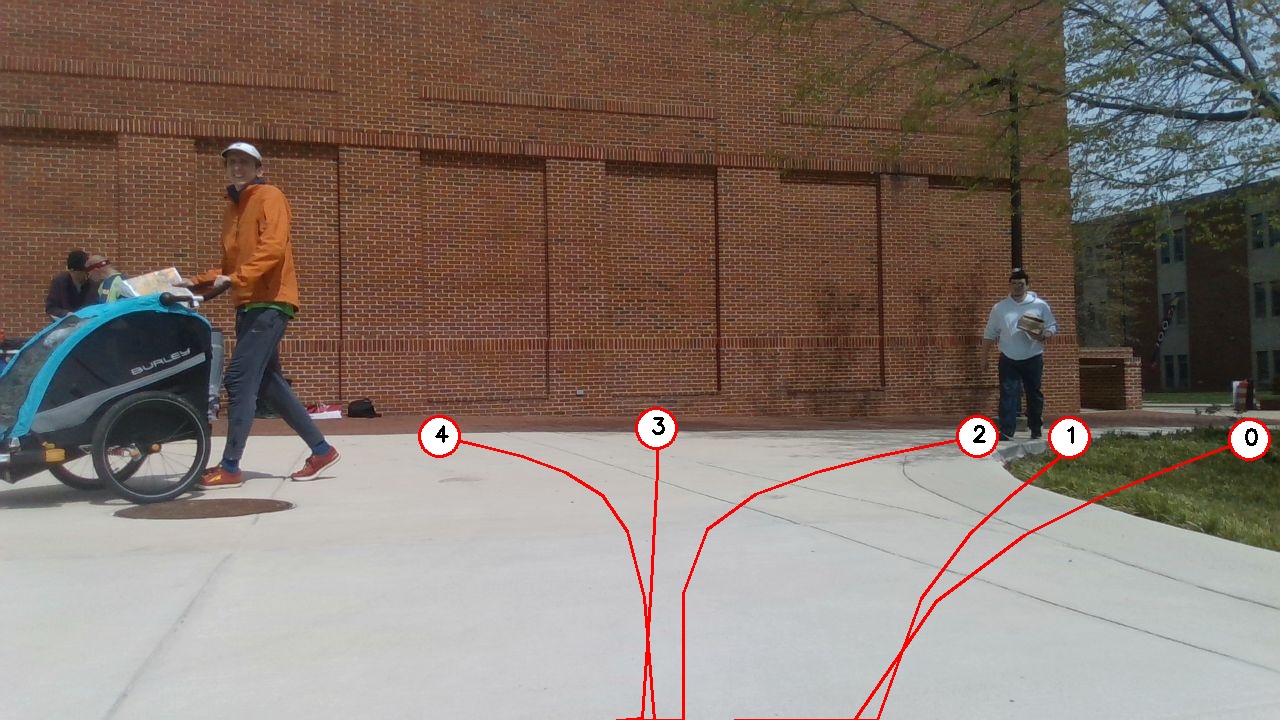} \end{minipage}
   \\ 
   \begin{minipage}{.3\linewidth}
   \includegraphics[width=\linewidth]{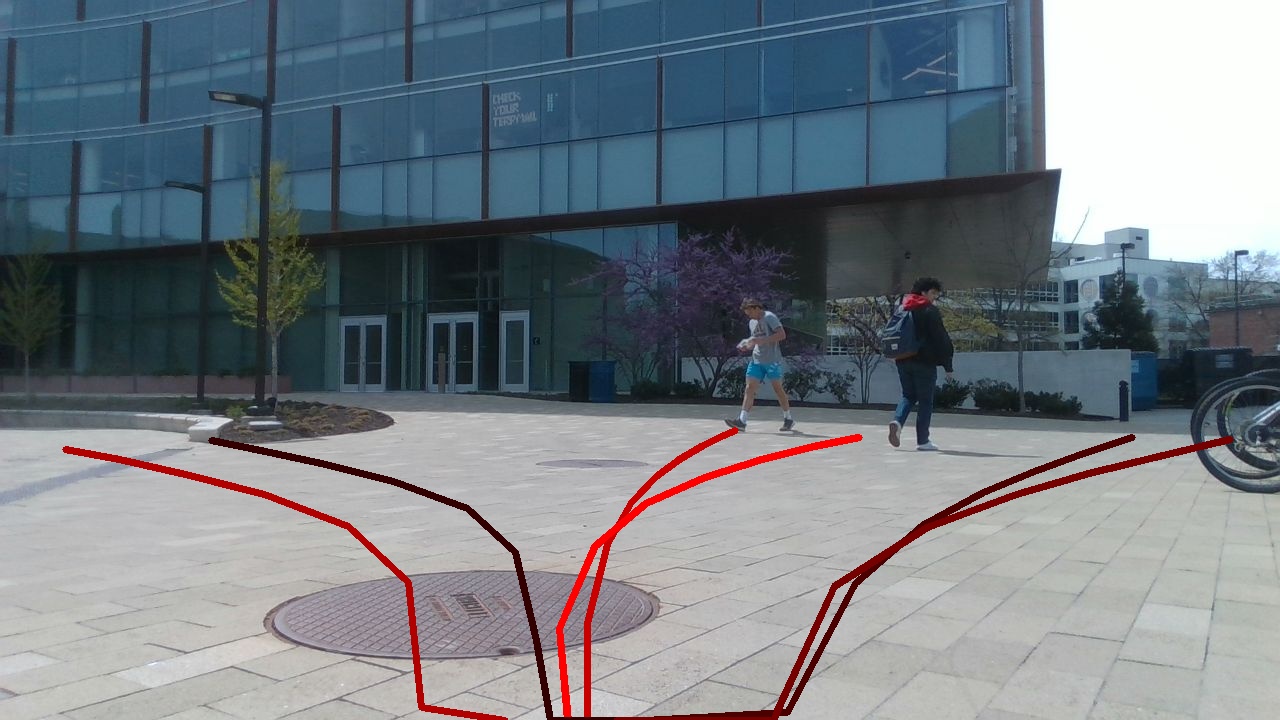}
   \end{minipage}
   & 
   \begin{minipage}{.3\linewidth} \includegraphics[width=\linewidth]{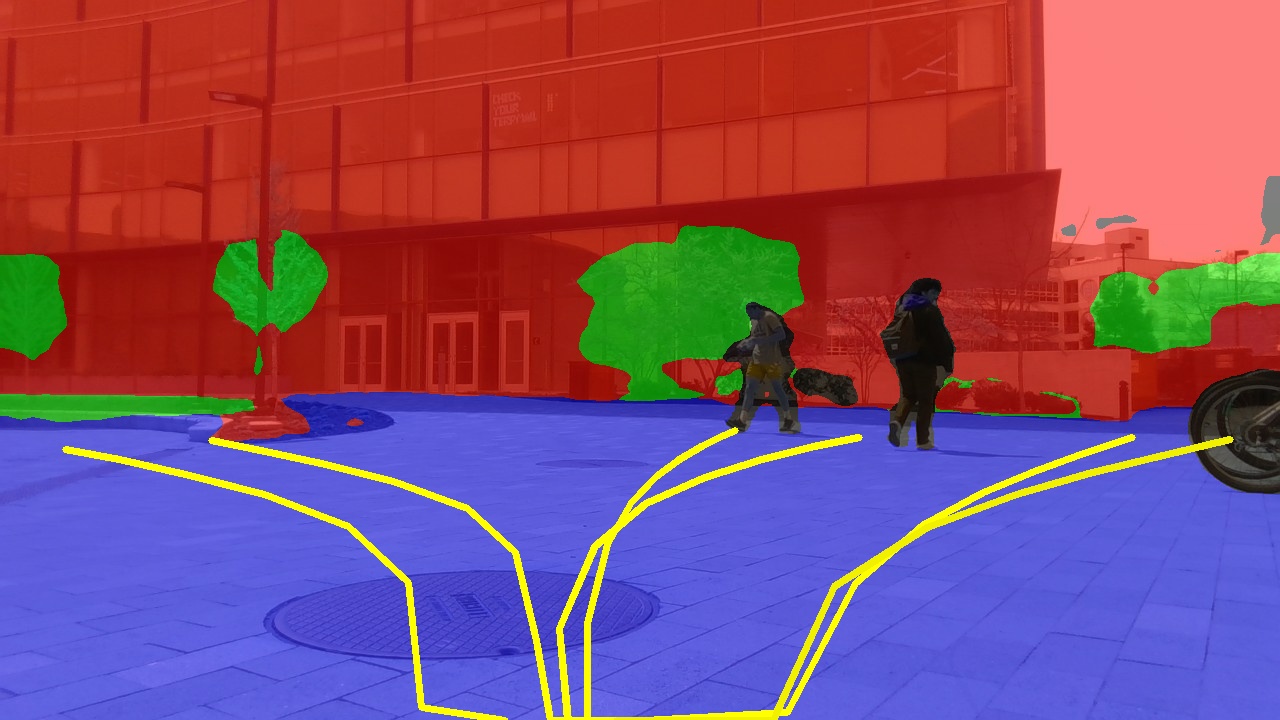} \end{minipage}
   & 
   \begin{minipage}{.3\linewidth} \includegraphics[width=\linewidth]{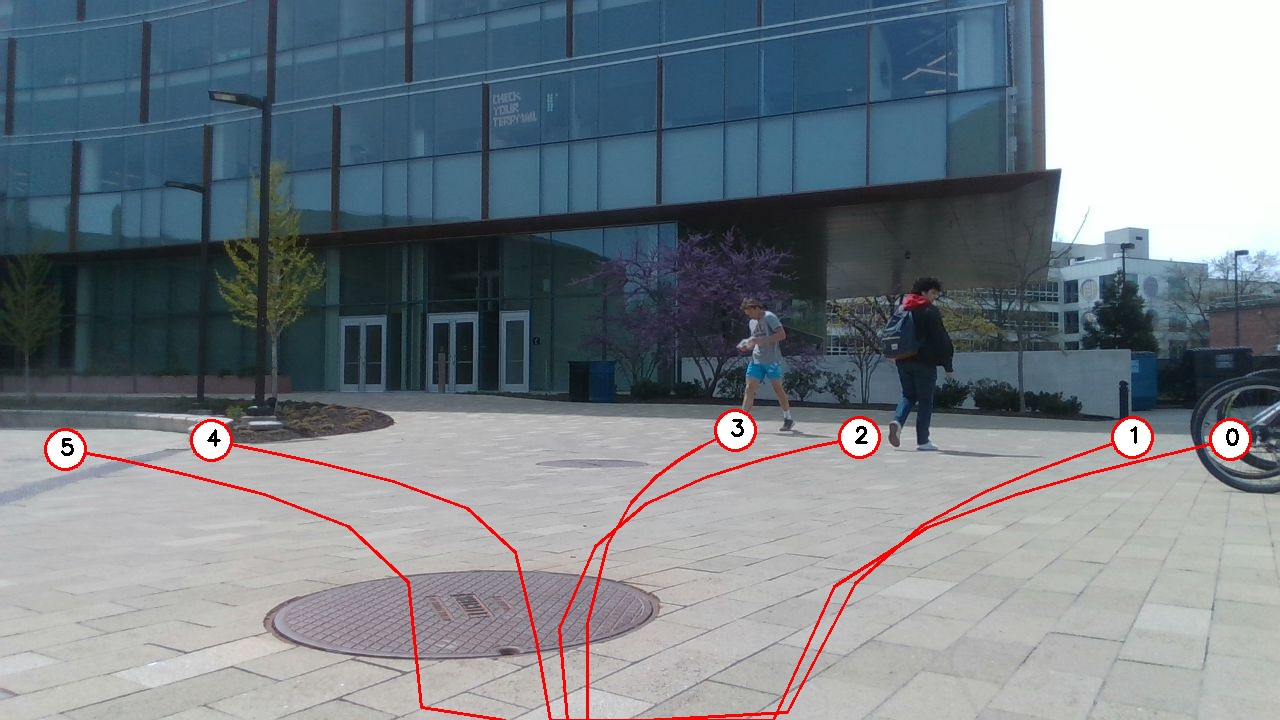} \end{minipage}
   \\ 
   (a) & (b) & (c) \\
  \end{tabular}
  \caption{
  \textbf{Qualitative Evaluation:} 
  Each column illustrates one of the three components in our trajectory generation pipeline, (a) CVAE-based multiple trajectory generation, (b) semantic segmentation for traversability analysis, and (c) VLM-based trajectory ranking in the outdoor scenarios with various social conditions. In (a) and (b), lighter colors indicate higher scores, confidence scores in (a) and traversability scores in (b). In (c), the VLM ranks trajectories based on social cues. The top example shows a ranking of [3, 0, 1, 2, 4]. The bottom example shows a ranking of [4, 5, 0, 1, 2, 3].}
  \label{fig:gnd_qualitative}
\end{figure*}

\setlength{\tabcolsep}{4pt}

\begin{table*}[h]
\centering
\begin{tabularx}{\textwidth}{c|c|c|c|c}
\multirow{2}{*}{\textbf{Method}} & \multirow{2}{*}{\textbf{Modality}} & \multirow{2}{*}{\textbf{Traversability (\%) $\uparrow$}} & \textbf{Distance to Target} & \textbf{Inference Time} \\  
 &  &  & \textbf{(\%) $\uparrow$} & \textbf{ (s) $\downarrow$} \\ \hline
PIVOT~\citep{nasiriany2024pivot} & RGB + Language & 70 & 69 & 2.30 \\ \hline
ViNT~\citep{shah2023vint} & RGB & 57 & 62 & 0.69 \\\hline
NoMaD~\citep{sridhar2024nomad} & RGB & 59 & 61 & 0.24 \\\hline
MTG~\citep{liang2024mtg} & Points & 61 & 64 & 0.01 \\ \hline
DTG~\citep{liang2024dtg} & Points & 67 & 66 & 0.13 \\ \hline
VL-TGS~\citep{song2024tgs} & Points, RGB, Language & 65 & 70 & 2.31 \\ \hline
\textbf{MOSU (ours)}~\citep{liang2025mosu} & Points, RGB, Language & \textbf{77} & \textbf{73} & 2.30 \\
\end{tabularx}
\vspace{0.5em}
\caption{\textbf{Quantitative Evaluation:} Our approach achieves the best Traversability and comparable Distance to Target.}
\label{tab:gnd_quantitative}
\end{table*}

      



\paragraph{Quantitative results.} Tab.~\ref{tab:gnd_quantitative} reports the retained comparison. MOSU achieves a traversability score of $77\%$, exceeding the strongest competing result, PIVOT's $70\%$, by seven percentage points, corresponding to a relative improvement of $10\%$. MOSU also achieves the highest distance-to-target score of $73\%$, compared with $70\%$ for VL-TGS and $69\%$ for PIVOT. These results indicate that the combined evaluation improves trajectory traversability while maintaining progress toward the destination under the reported benchmark. The reported inference time for MOSU is $2.30\,\mathrm{s}$, matching PIVOT and closely matching VL-TGS at $2.31\,\mathrm{s}$ with Gemini. The remaining methods report inference times between $0.01\,\mathrm{s}$ and $0.69\,\mathrm{s}$. The results show that the VLM is the bottleneck in inference time. Therefore, in MOSU2 we leverage the ChatGPT-realtime with shorter inference time for each analysis.

\paragraph{Qualitative results.} As shown in Fig.~\ref{fig:challenge} (a), geometry-based trajectory generation~\citep{liang2024mtg} performs well in scenarios where geometric structures are easily detected. However, when the elevations of off-road areas and sidewalks are similar, it often fails, as shown in (c). In Fig.~\ref{fig:challenge} (a) and (b), lighter colors indicate higher scores in both geometric confidence and color-based traversability. As shown in (d), semantic segmentation also encounters out-of-distribution scenarios, where large ground areas cannot be properly segmented. Therefore, geometric and semantic information must be combined for reliable traversability analysis, and MOSU2 inherits this design to filter candidate trajectories. As shown in Table~\ref{tab:gnd_quantitative}, our approach achieves the best traversability among all methods, and the generated trajectories lead to the target closely.

As shown in Fig.~\ref{fig:gnd_qualitative}, beyond traversability analysis, VLMs process labeled trajectories and analyze social cues from the images. In the first row, the VLM ranks the trajectories as [3, 0, 1, 2, 4], given the target at the right front. It suggests following trajectory 3 at normal speed, while other trajectories should be taken at a slower speed when encountering humans. The ranking for the second row is [4, 5, 0, 1, 2, 3] with normal speed. In the experiment, we observed that VLMs take a significant amount of time to process, as shown in Table~\ref{tab:gnd_quantitative}, but they demonstrate strong capability in understanding social cues.

These examples provide qualitative evidence of contextual trajectory ranking. They do not quantify social-cue recognition accuracy or establish the reliability of decisions that depend on accurately measured object motion. The following experiments therefore examine whether access to local geometric and dynamic information improves navigation under social, traffic, and irregular conditions, and at what time cost.

\subsection{Evaluation of MOSU2}

\begin{figure*}[t]
  \centering
  \begin{tabular}{ccc}
  
   \begin{minipage}{.31\linewidth}
    \centering
    \includegraphics[width=\linewidth,height=0.6\linewidth]{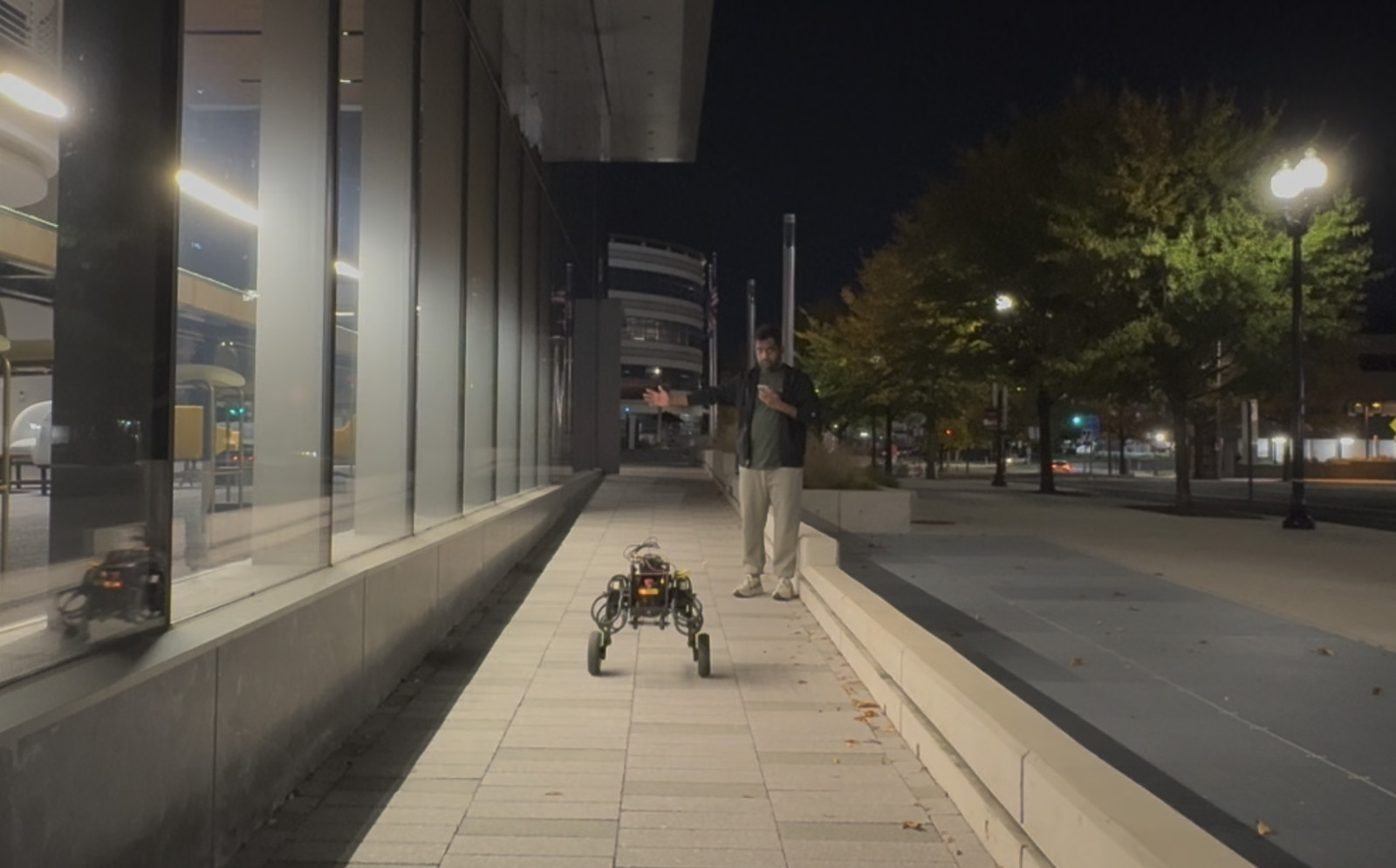}
    (a)
   \end{minipage}
   &
   \begin{minipage}{.31\linewidth}
    \centering
    \includegraphics[width=\linewidth,height=0.6\linewidth]{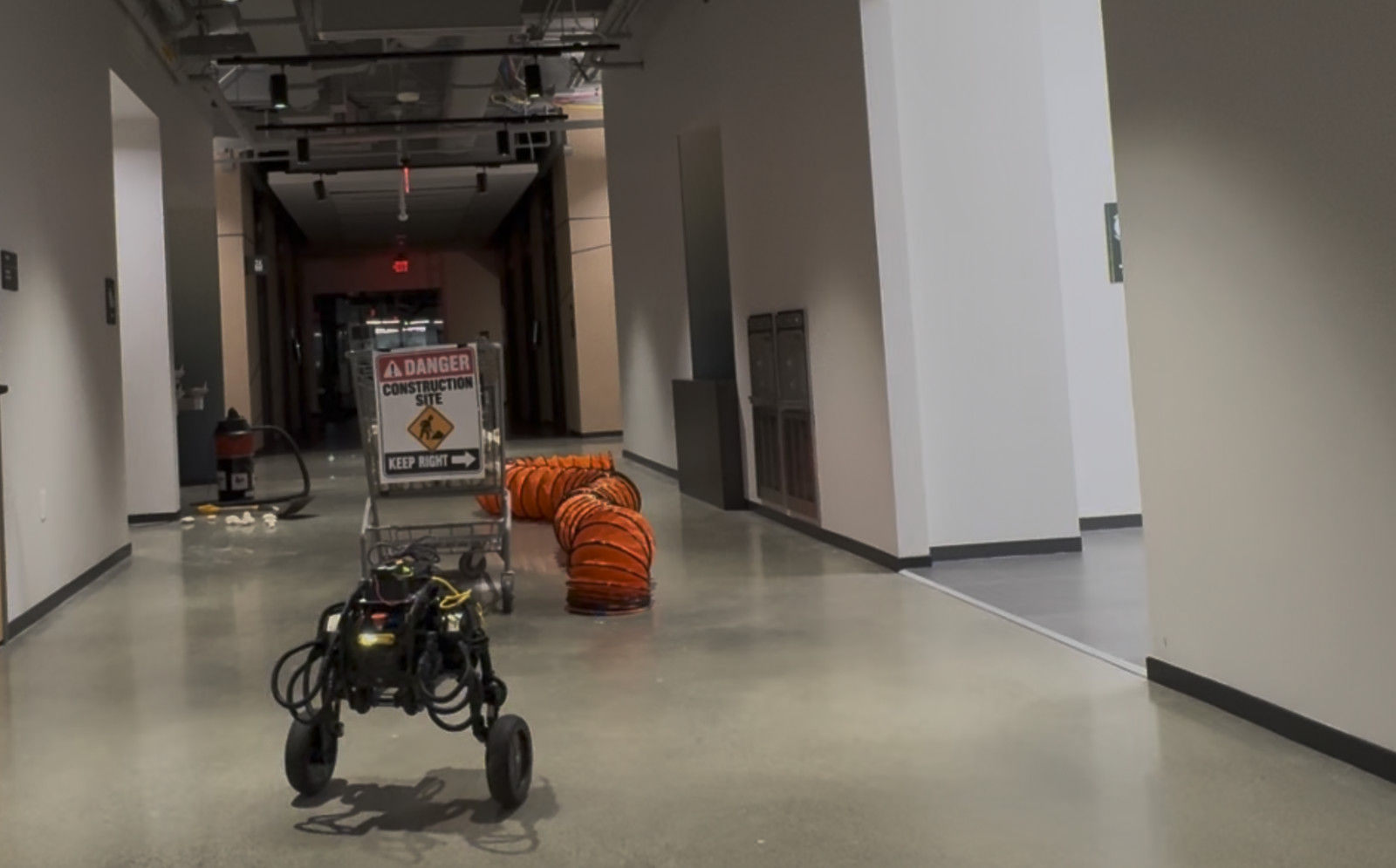}
    (b)
   \end{minipage} 
   & 
   \begin{minipage}{.31\linewidth}
    \centering
    \includegraphics[width=\linewidth,height=0.6\linewidth]{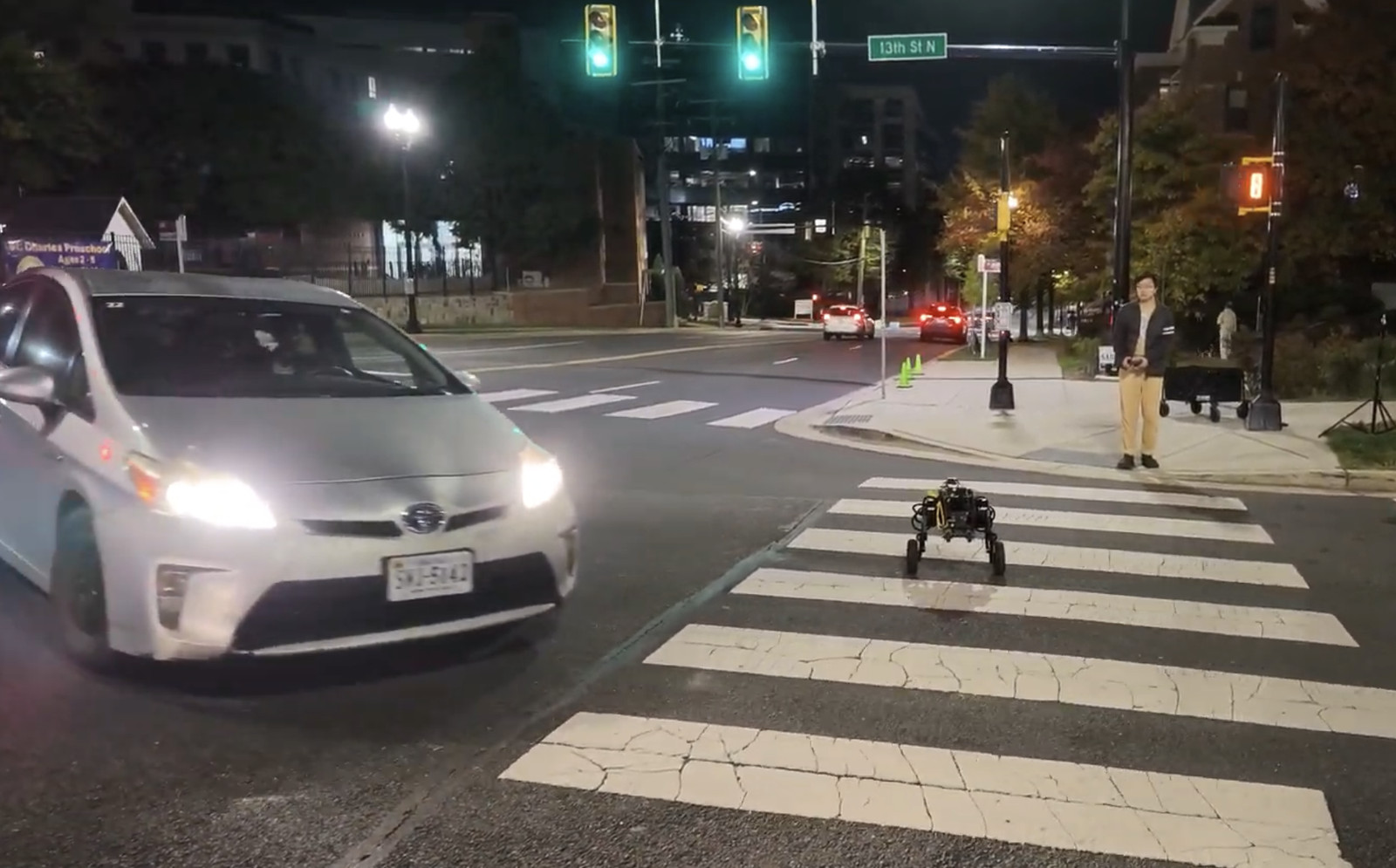}
    (c)
   \end{minipage}
   \\
   \rule{0pt}{3ex} 
   \begin{minipage}{.31\linewidth}
    \centering
    \includegraphics[width=\linewidth,height=0.6\linewidth]{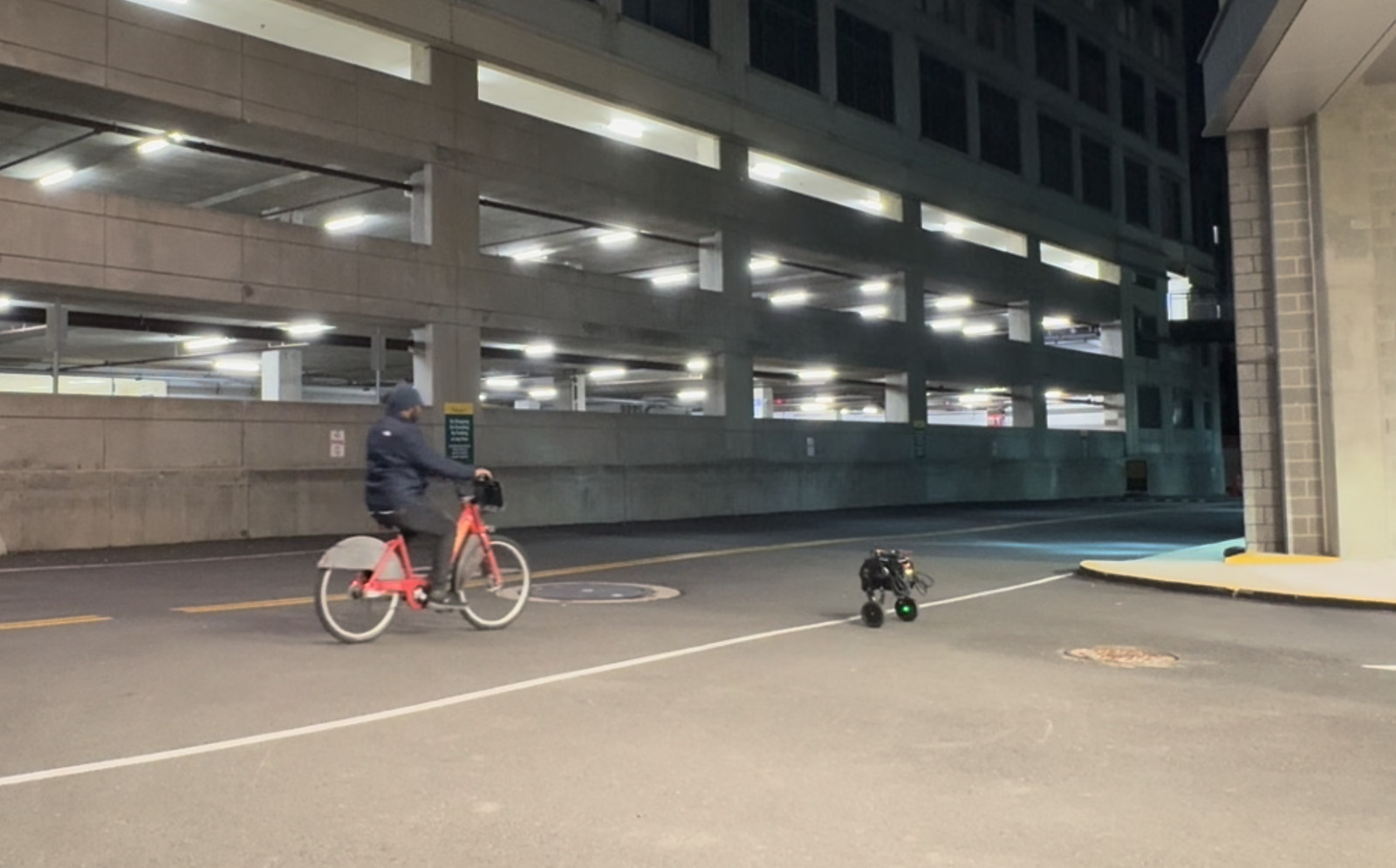}
    (d)
   \end{minipage}
   & 
   \begin{minipage}{.31\linewidth}
    \centering
    \includegraphics[width=\linewidth,height=0.6\linewidth]{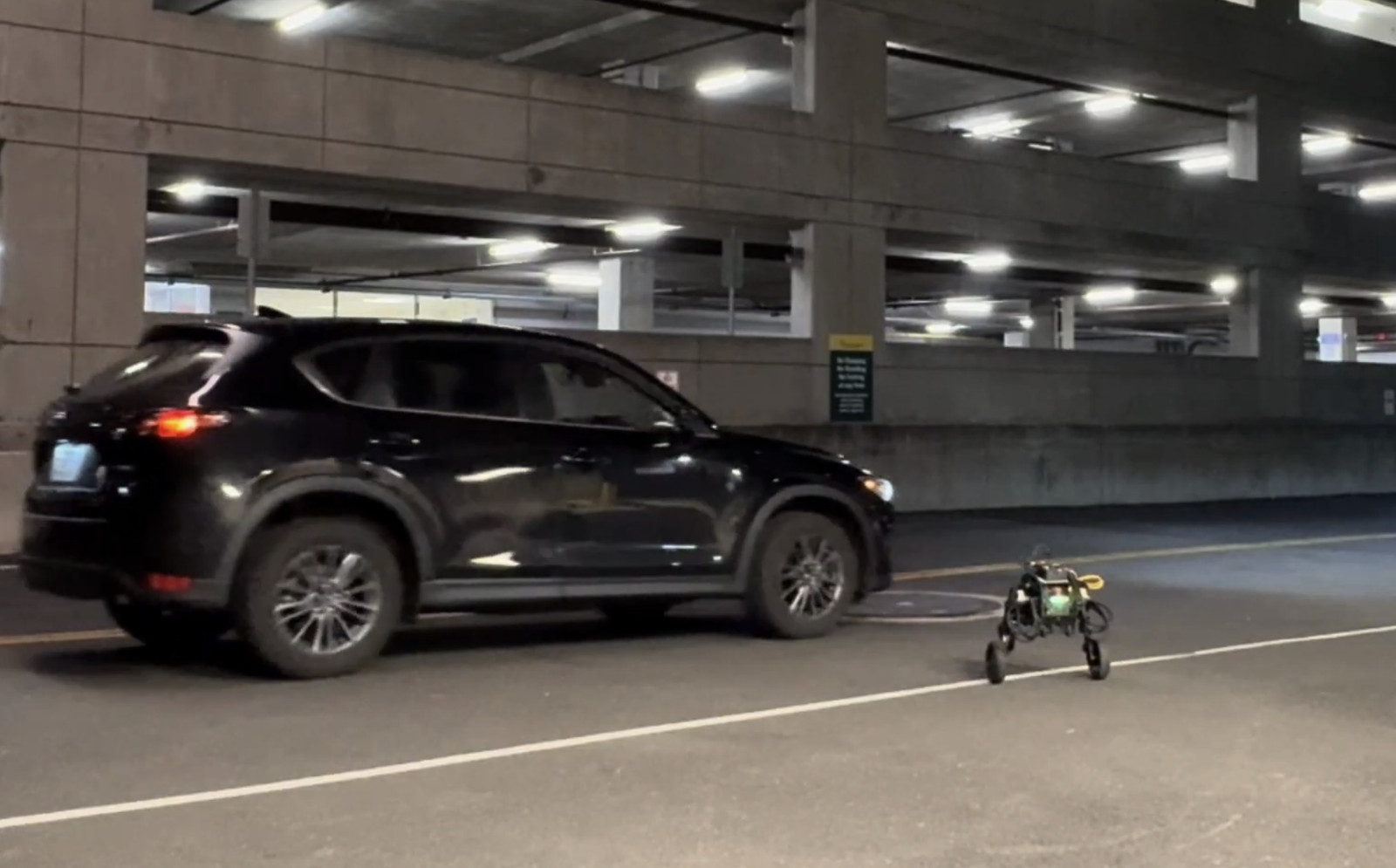}
    (e)
   \end{minipage}
   & 
   \begin{minipage}{.31\linewidth}
    \centering
    \includegraphics[width=\linewidth,height=0.6\linewidth]{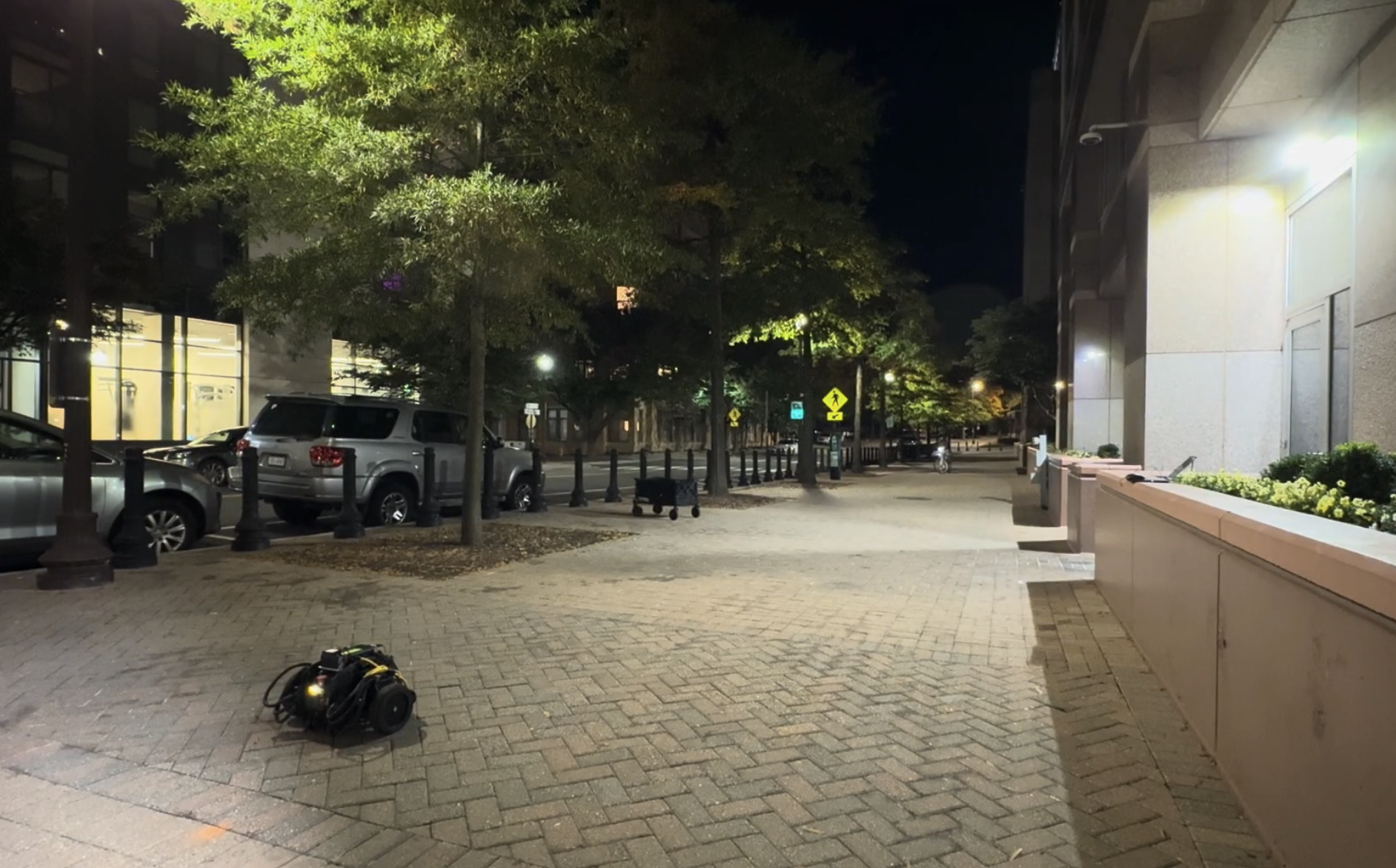}
    (f)
   \end{minipage}

  \end{tabular}
  \caption{\textbf{Representative social, traffic, and irregular navigation scenarios used to evaluate MOSU2}: (a) responding to a pedestrian's stop gesture, (b) navigating a construction area with warning and directional signs, (c) crossing at a signalized intersection, (d) sharing the road with moving agents, illustrated by a cyclist, (e) intersecting with a vehicle, requiring the robot to analyze the car's velocity to determine if it is yielding, and (f) navigating in sidewalk baseline without any social or traffic situation. Panels (d) and (e) both belong to the shared-road group (Road + Vehicles/Bikes/Peds) in Table~\ref{tab:mosu_mosu2_comparison}. These scenarios require interpreting contextual cues and, where relevant, considering the geometry and motion of surrounding road users to determine appropriate robot behavior.}
  \label{fig:social_traffic_challenge}
\end{figure*}

\begin{table*}[t]
    \centering
    \caption{\textbf{Performance Comparison:} MOSU vs. MOSU2 across diverse dynamic scenarios. Nav. Success Rate is the fraction of trials in which the robot successfully reached the subgoal, and Det. Accuracy is the accuracy of detecting the social and traffic conditions. Our approach substantially improves the Nav. Success Rate and Det. Accuracy in most of the social and traffic scenarios. Tokens are given as input/output per decision step, and those of MOSU2 are reconstructed from its prompts and outputs.}
    \label{tab:mosu_mosu2_comparison}
    \small
    \newcolumntype{C}{>{\centering\arraybackslash}X}
    \begin{tabularx}{\textwidth}{@{} l C C C C C C C C @{}}
        \toprule
        & \multicolumn{4}{c}{\textbf{MOSU}} & \multicolumn{4}{c}{\textbf{MOSU2}} \\
        \cmidrule(lr){2-5} \cmidrule(lr){6-9}
        \textbf{Scenario} & \textbf{Nav. Success Rate (\%)$\uparrow$} & \textbf{Latency (s)$\downarrow$} & \textbf{Det. Accuracy (\%)$\uparrow$} & \textbf{Tokens (in/out)$\downarrow$} & \textbf{Nav. Success Rate (\%)$\uparrow$} & \textbf{Latency (s)$\downarrow$} & \textbf{Det. Accuracy (\%)$\uparrow$} & \textbf{Tokens (in/out)$\downarrow$} \\ \midrule
        Stop gesture & 61 & \textbf{1.51} & 70 & \textbf{821/53} & \textbf{71} & 2.52 & \textbf{89} & 1,507/135 \\
        Construction site & 50 & \textbf{1.68} & \textbf{100} & \textbf{821/56} & \textbf{100} & 5.16 & \textbf{100} & 5,386/446 \\
        Signalized crosswalk & 42 & \textbf{1.90} & 41 & \textbf{821/65} & \textbf{100} & 4.72 & \textbf{93} & 4,948/478 \\
        Road + Vehicles/Bikes/Peds & 56 & \textbf{1.79} & 50 & \textbf{822/62} & \textbf{100} & 4.95 & \textbf{89} & 5,142/454 \\
        Sidewalk baseline & \textbf{100} & \textbf{1.74} & \textbf{100} & \textbf{821/61} & \textbf{100} & 1.80 & 86 & 821/53 \\ \bottomrule
    \end{tabularx}
\end{table*}


\begin{figure*}
  \centering
  \setlength{\tabcolsep}{2pt}
  \begin{tabular}{ccc}
    (a) Points and RGB Inputs & (b) Geometric map & (c) Dynamic map \\
    \includegraphics[width=0.40\textwidth,height=0.27\textwidth,keepaspectratio]{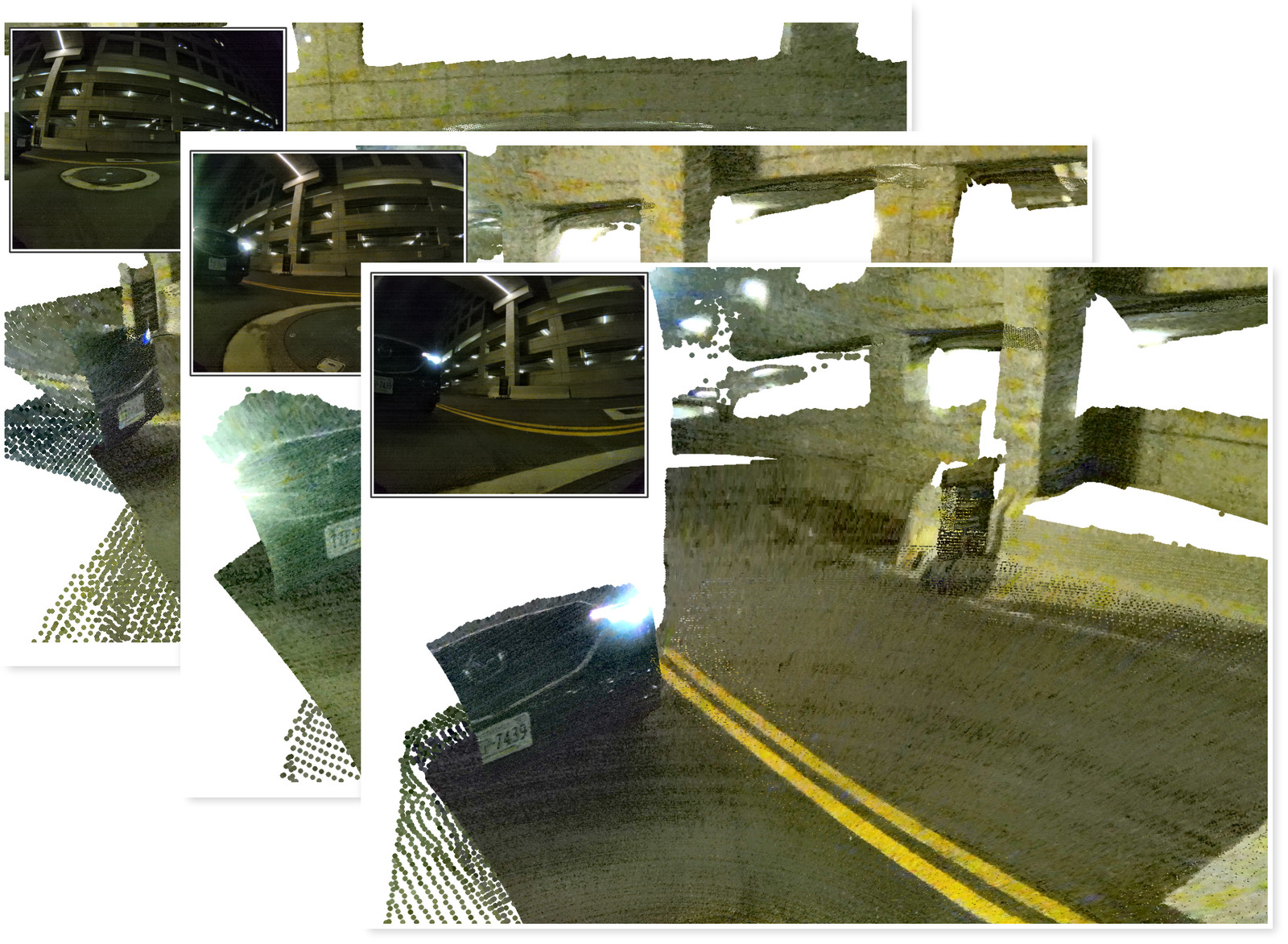} & \includegraphics[width=0.27\textwidth,height=0.27\textwidth,keepaspectratio]{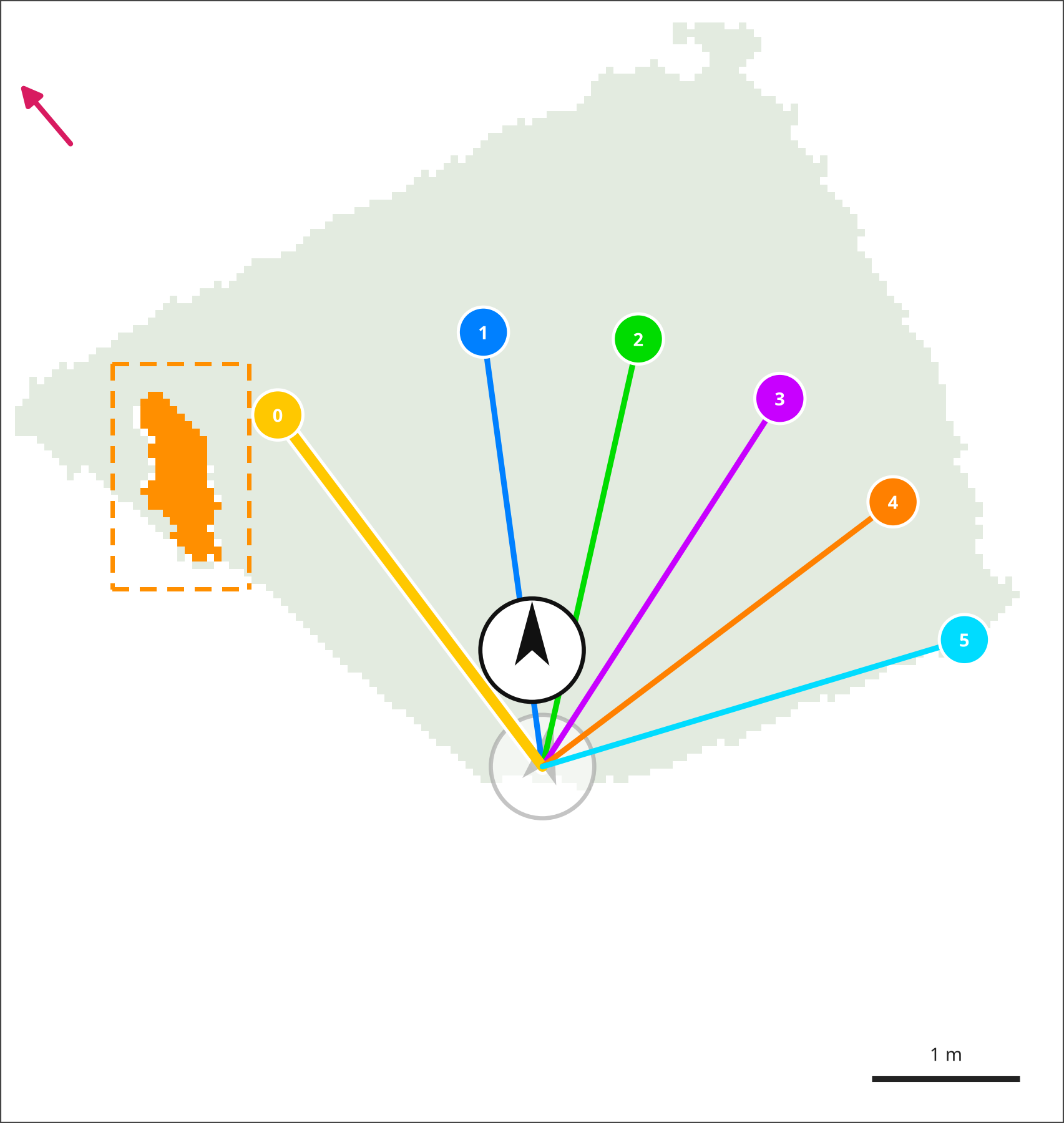} & \includegraphics[width=0.27\textwidth,height=0.27\textwidth,keepaspectratio]{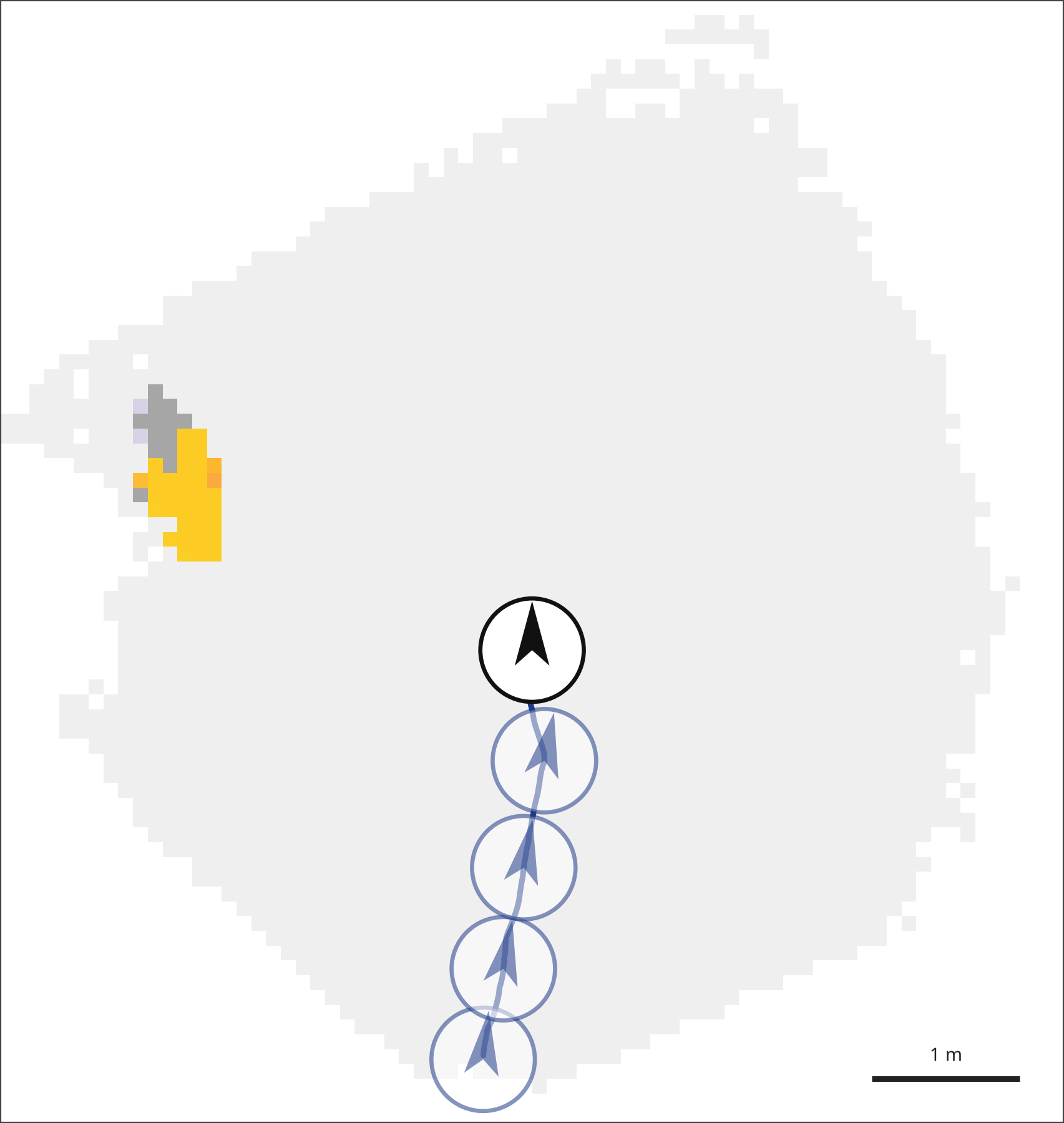} \\
    \includegraphics[width=0.40\textwidth,height=0.27\textwidth,keepaspectratio]{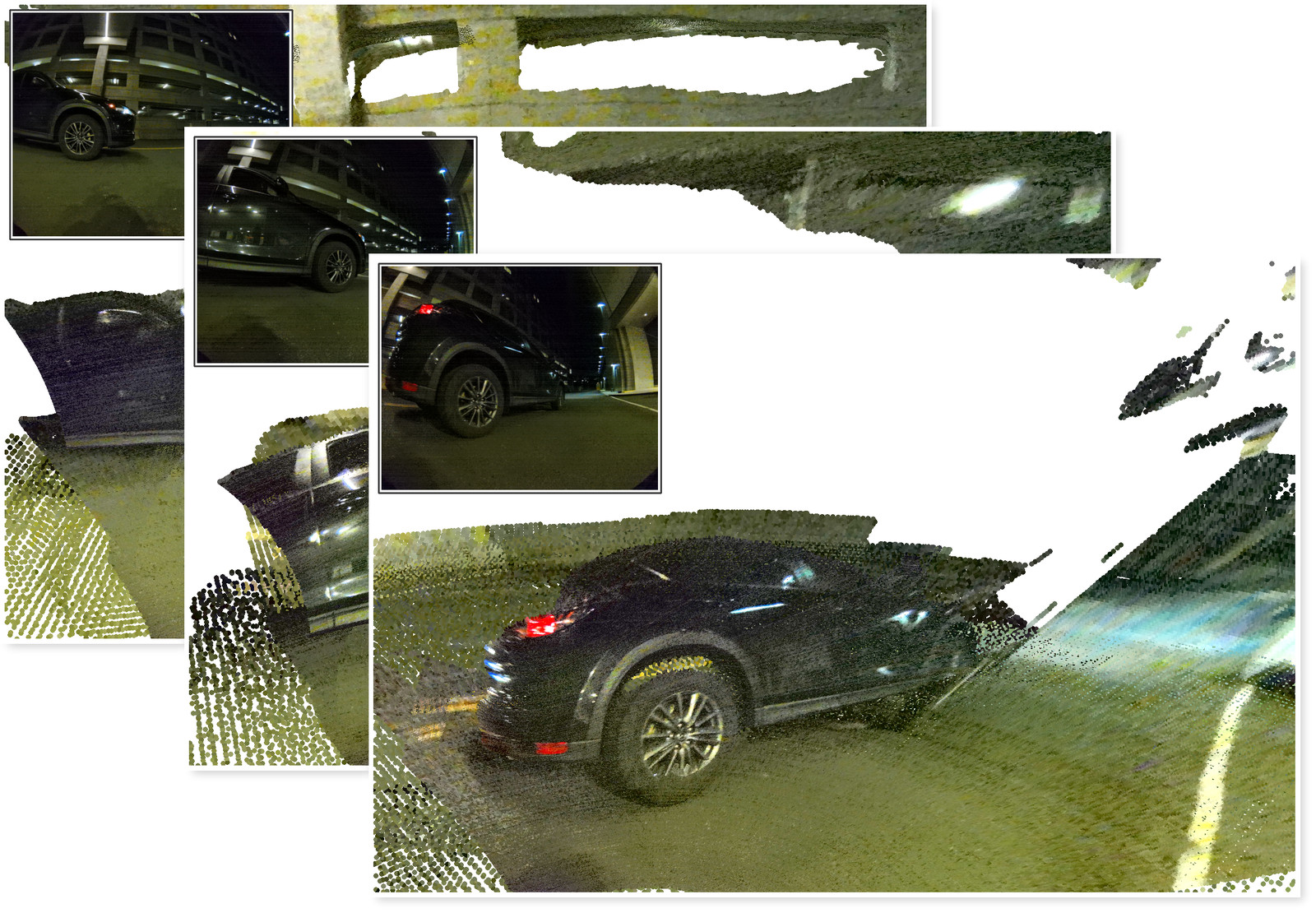} & \includegraphics[width=0.27\textwidth,height=0.27\textwidth,keepaspectratio]{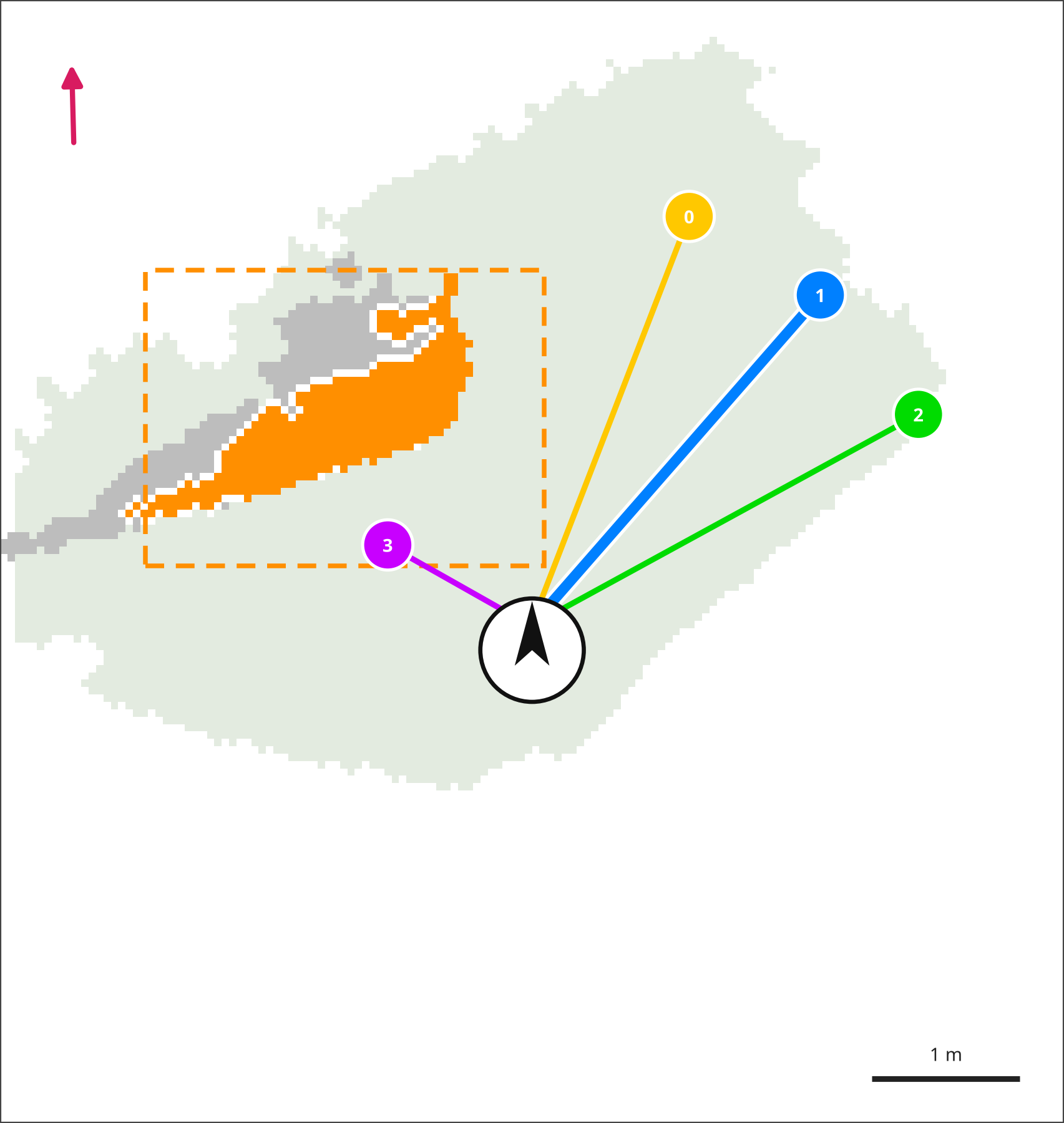} & \includegraphics[width=0.27\textwidth,height=0.27\textwidth,keepaspectratio]{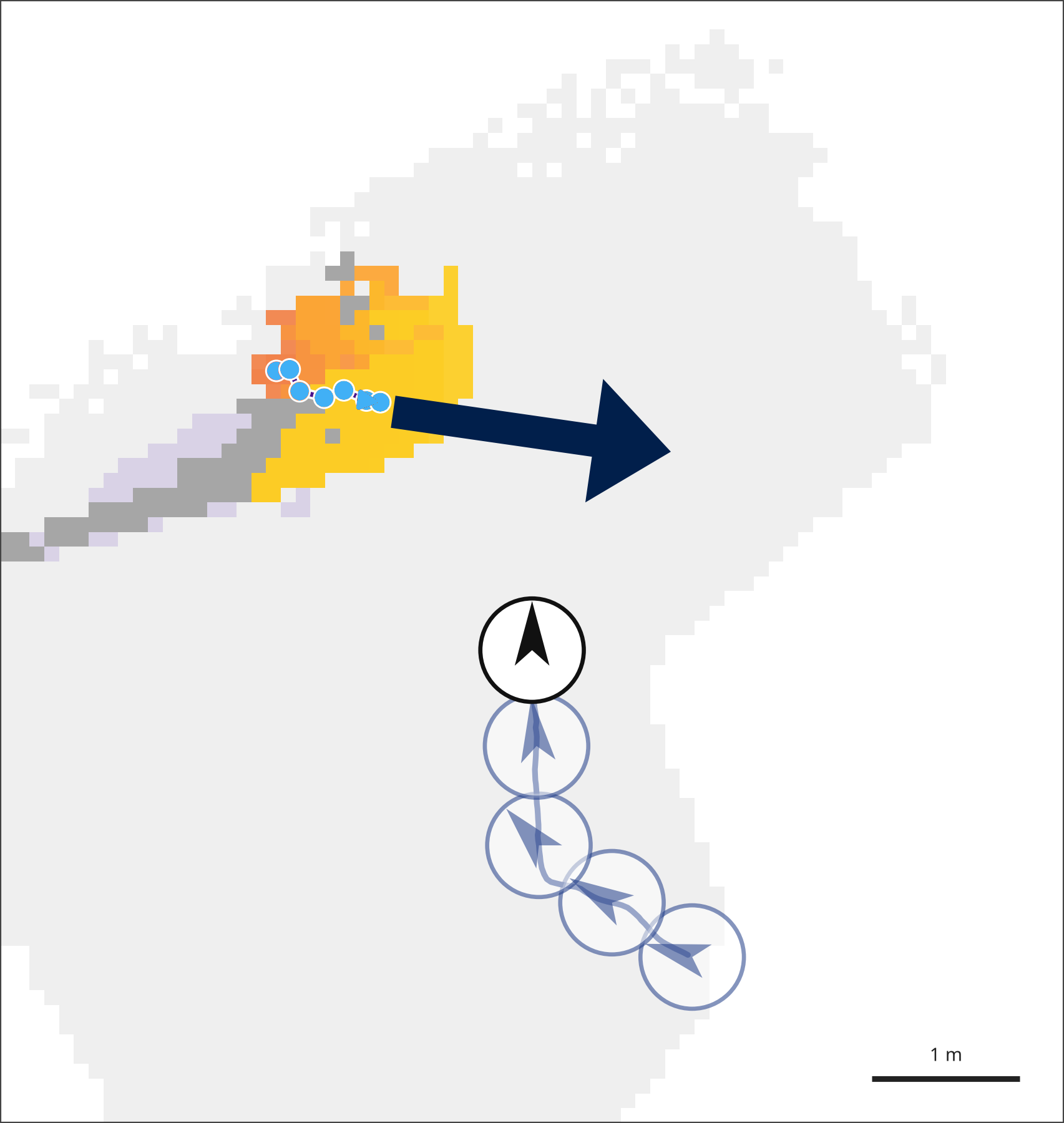} \\
    \includegraphics[width=0.40\textwidth,height=0.27\textwidth,keepaspectratio]{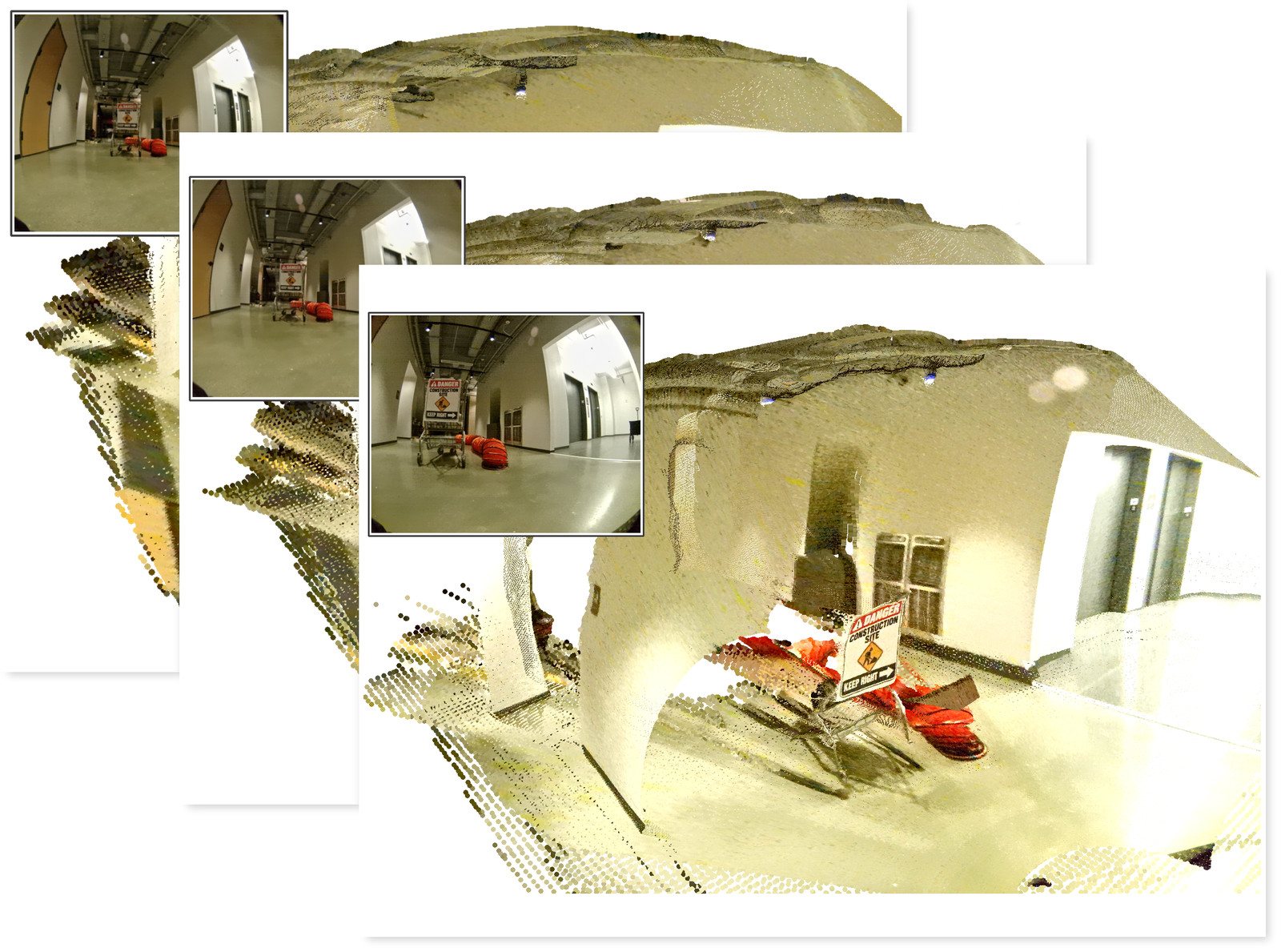} & \includegraphics[width=0.27\textwidth,height=0.27\textwidth,keepaspectratio]{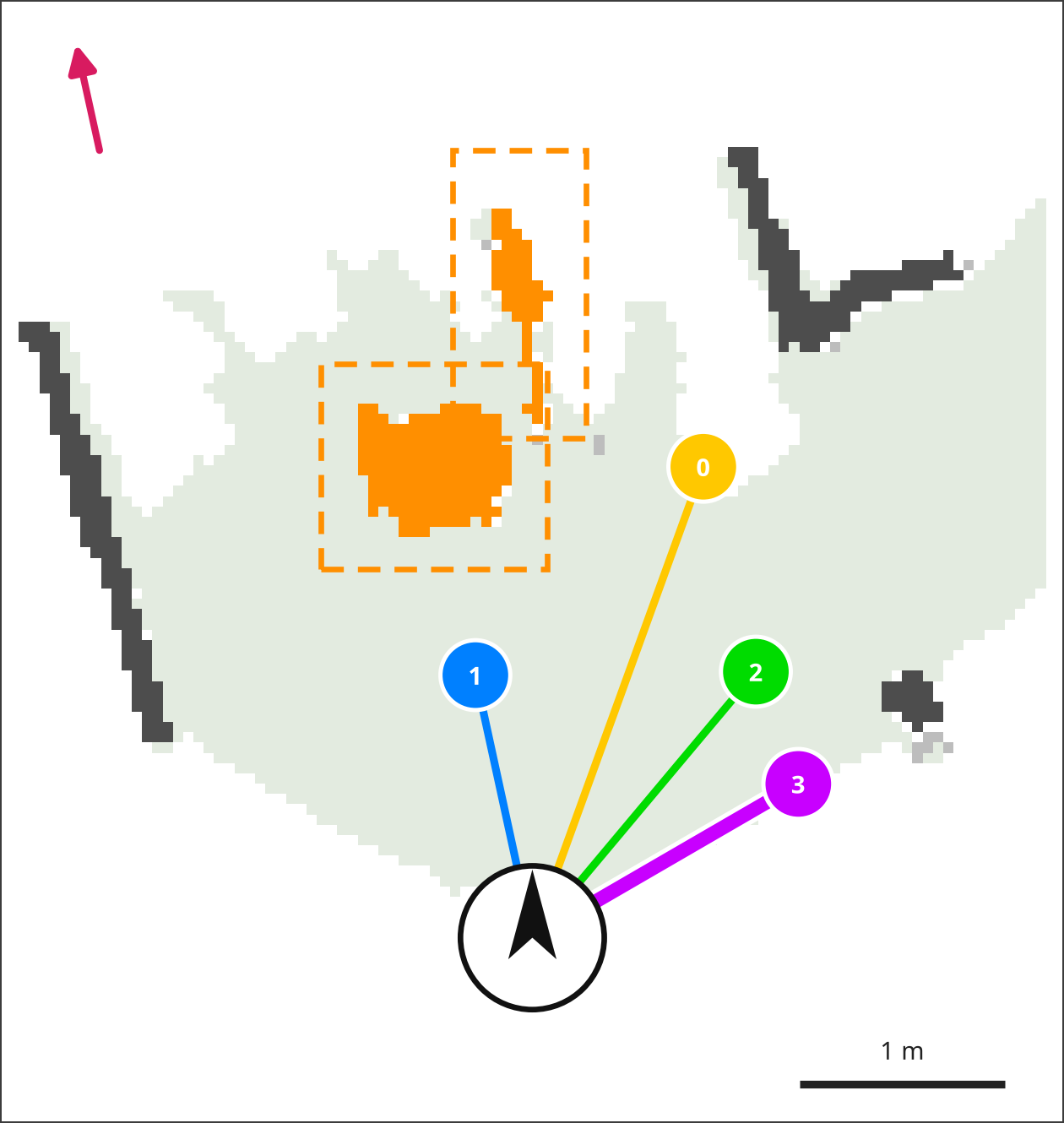} & \includegraphics[width=0.27\textwidth,height=0.27\textwidth,keepaspectratio]{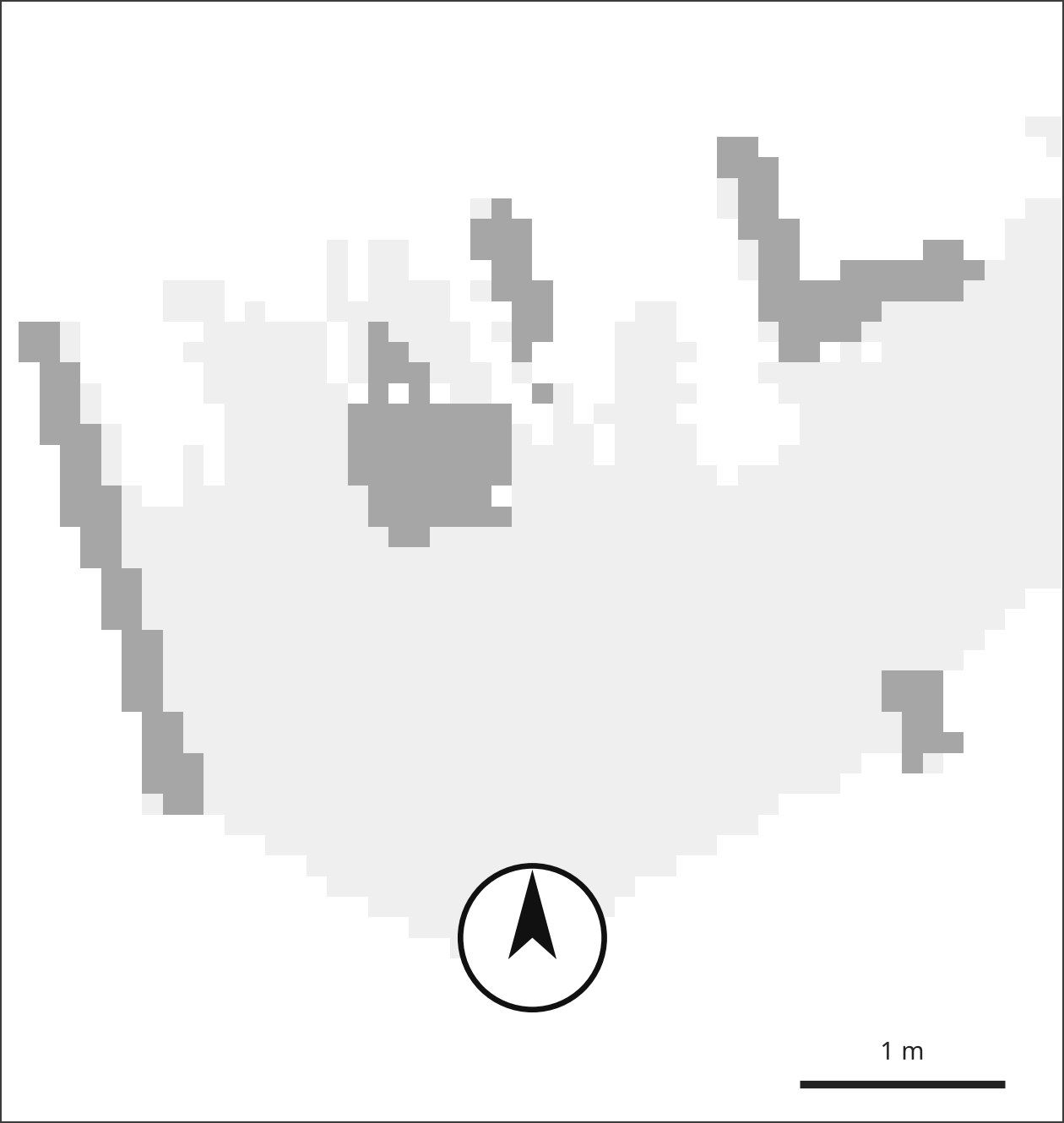} \\
  \end{tabular}
  \caption{\textbf{VLM-retrievable local maps in three scenarios} (one per row): a stopped car beside the robot's path (top), a car crossing in front of the robot (middle), and a construction site in an indoor corridor (bottom). (a) Consecutive RGB images (insets) and the corresponding LiDAR point clouds colored by the camera images. (b) Geometric layer with the observed free space (light green), segmented objects (orange, dashed boxes), other occupied cells (gray), numbered candidate trajectories, the robot pose (black marker), and the subgoal direction (red arrow). (c) Dynamic layer: the crossing car leaves a track of positions (blue points in the middle figure) with a black arrow to indicate the velocity direction, whereas the stopped car (top) and the construction site (bottom) show no displacement. Blue markers show the robot's recent poses, and the scale bars indicate 1\,m.}
  \label{fig:local_map_scenarios}
\end{figure*}

\newlength{\qualH}  \setlength{\qualH}{26mm}       
\newlength{\fpvW}   \setlength{\fpvW}{1.293\qualH}  
\newlength{\tpvW}   \setlength{\tpvW}{1.50\qualH}   
\begin{figure*}[p]
  \centering
  \footnotesize
  \renewcommand{\tabularxcolumn}[1]{m{#1}}  
  \setlength{\tabcolsep}{4pt}

  \begin{tabularx}{\textwidth}{@{} >{\centering\arraybackslash}m{\fpvW}
                                   >{\raggedright\arraybackslash}X
                                   >{\centering\arraybackslash}m{\tpvW} @{}}
    \toprule
    \textbf{First-Person View} & \textbf{VLM Reasoning \& Decision Output} & \textbf{Third-Person View (Execution)} \\
    \midrule

    \includegraphics[height=\qualH]{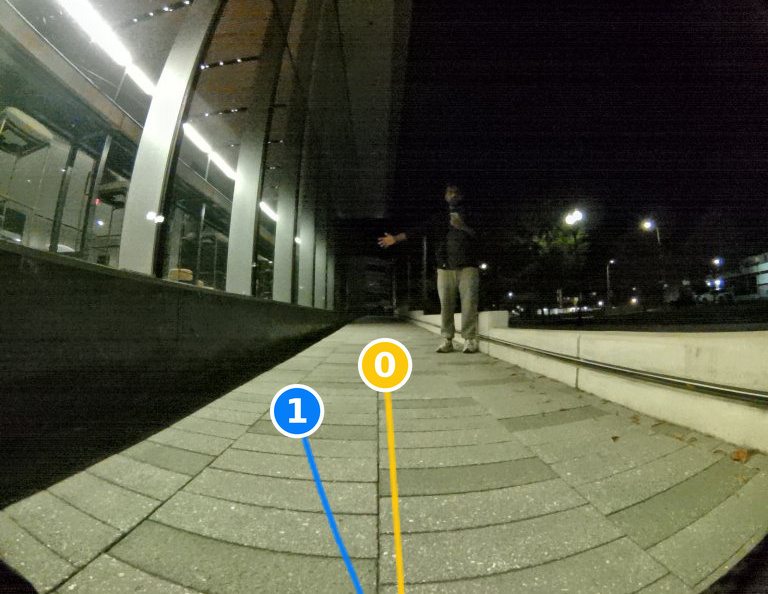} &
    \textbf{MOSU:} mode: Slow, ranking: [1, 0], reason: "A pedestrian is ahead. Path 1 provides more clearance while remaining on the paved walkway...."\par\vspace{3pt}
    \textbf{MOSU2:} 
    \textbf{Stage 1:} False, "Stop", 0, "no further information needed", "A nearby pedestrian is holding out an open palm in a stop gesture." &
    \includegraphics[height=\qualH]{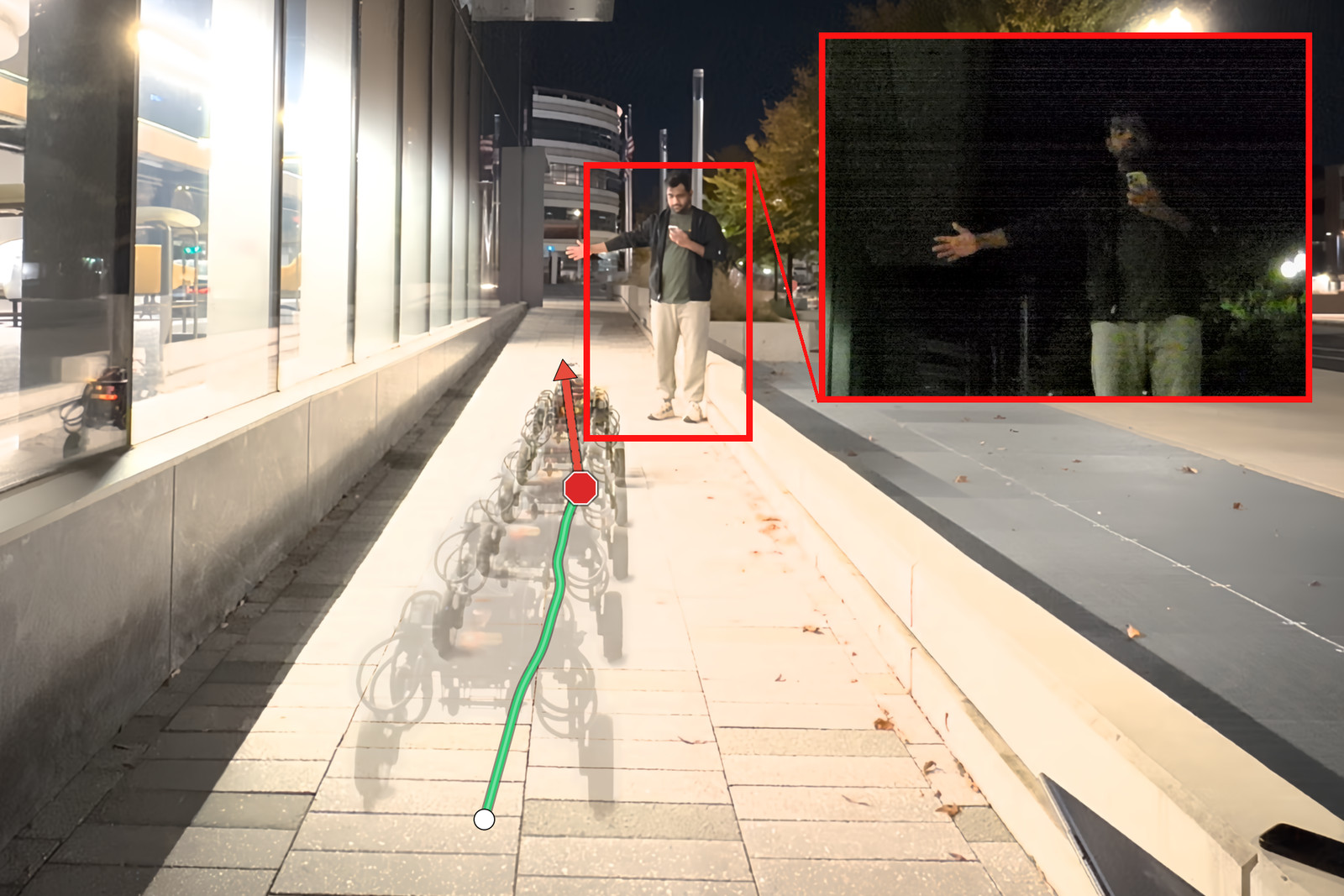} \\
    \addlinespace[3pt]   
    \midrule

    \includegraphics[height=\qualH]{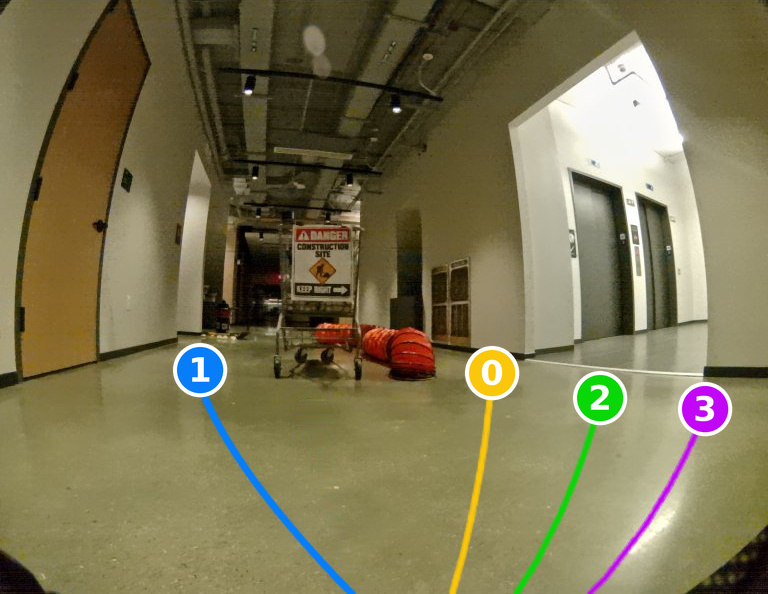} &
    \textbf{MOSU:} mode: Slow, ranking: [0, 2, 3, 1], reason: "The construction sign directs traffic to keep right. Path 0 stays right of the equipment and is closest to the goal; paths 2 and 3 are alternatives. Path 1 approaches the construction equipment."\par\vspace{3pt}
    \textbf{MOSU2:} \textbf{Stage 1:} True, "Items: construction cart, ducts and sign (static obstacles). Needed: position, and whether the right side is passable. ..." \par\vspace{3pt}
    \textbf{Stage 2:} Slow, 3, "the site is in the middle of the corridor 5 m ahead, so any path through the center (1 and 0) collides with it or passes too close. The input confirms the sign that the right side is open." &
    \includegraphics[height=\qualH]{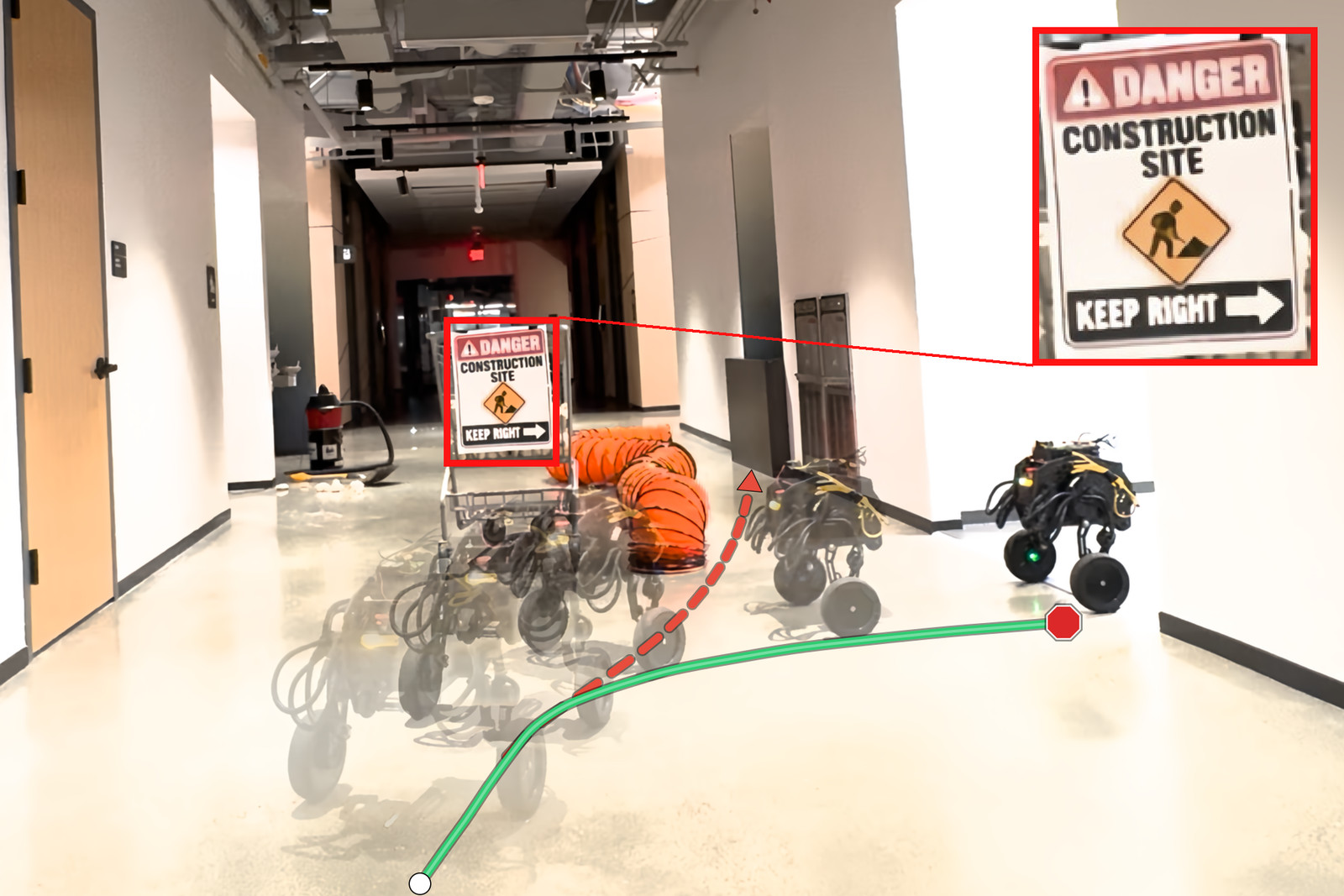} \\
    \addlinespace[3pt] 
        \midrule

    \includegraphics[height=\qualH]{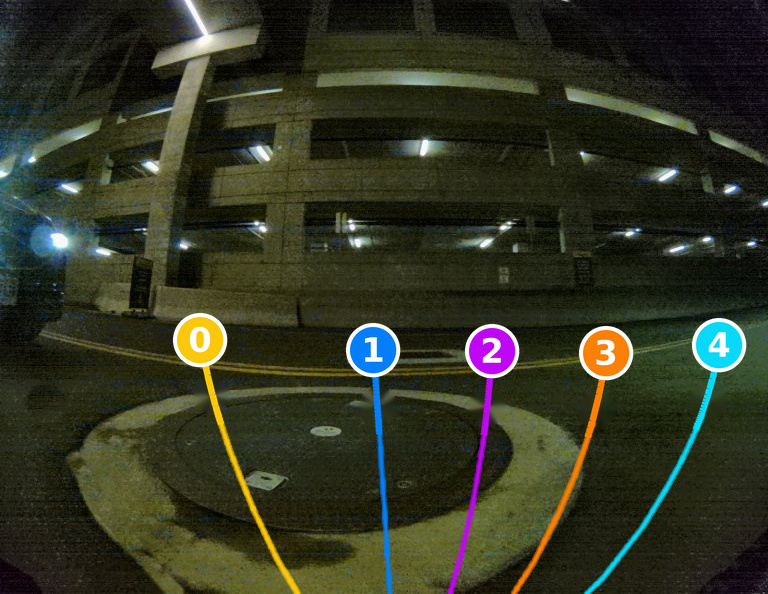} &
    \textbf{MOSU:} [4, 3, 2, 1, 0], Stop, "A dark vehicle is close on the left, and a round manhole cover and its raised rim sit in front of the robot...." \par\vspace{3pt}
    \textbf{MOSU2:} \textbf{Stage 1:} True, "Item: black vehicle at the left edge. Needed: position relative to the robot, to judge clearance from path 0, and velocity, to know whether it is stopped or about to pull out." \par\vspace{3pt}
    \textbf{Stage 2:} Slow, 1, "the goal is 5.7 m away at 29° left, ... The car is stopped, so there is no collision risk now, ..."
    &
    \includegraphics[height=\qualH]{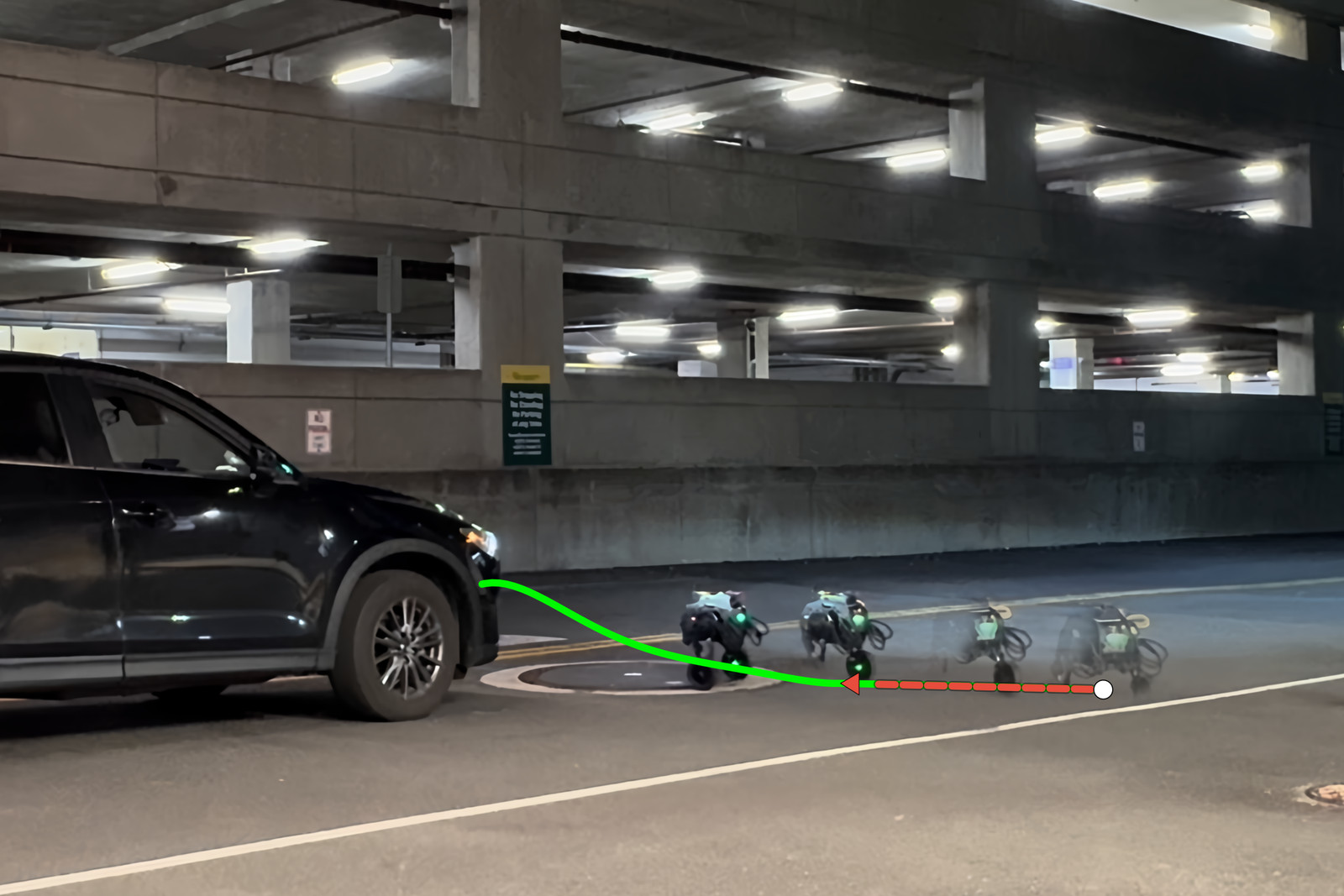} \\
    \addlinespace[3pt] 
        \midrule

    \includegraphics[height=\qualH]{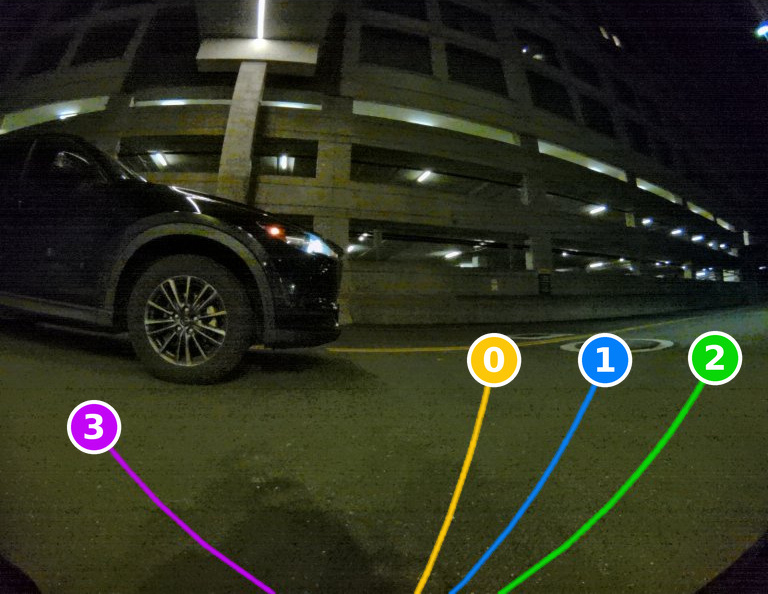} &
    \textbf{MOSU:} [2, 1, 0, 3], "Stop", "A dark SUV is close on the left with its headlight on, and it is moving across or into the robot's path...."\par\vspace{3pt}
    \textbf{MOSU2:}\textbf{Stage 1:} True, "Item: a black SUV. Needed: position and velocity, to know if it crosses the robot's path. ..." \par\vspace{3pt}
    \textbf{Stage 2:} Stop, "After the car has passed, take path 0....", "at 2 m/s the car crosses the robot's forward line in about 0.5 s and sweeps across all four candidate paths, so none is safe now. It is only 1 m away, so the robot must not move at all. Once it has crossed to the right, the area in front is clear and path 0 goes toward the goal"
    &
    \includegraphics[height=\qualH]{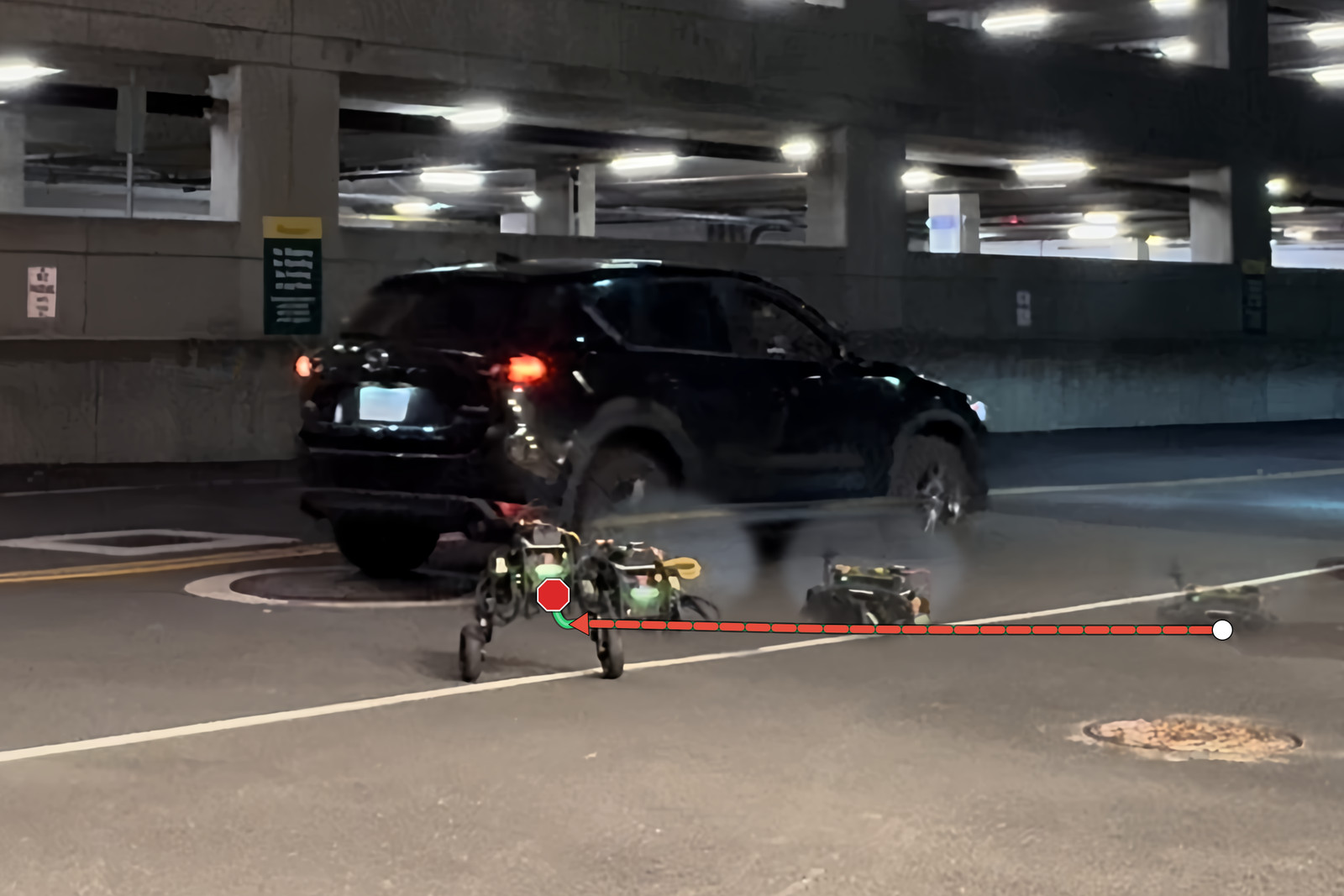} \\
    \addlinespace[3pt]
    \midrule

    \includegraphics[height=\qualH]{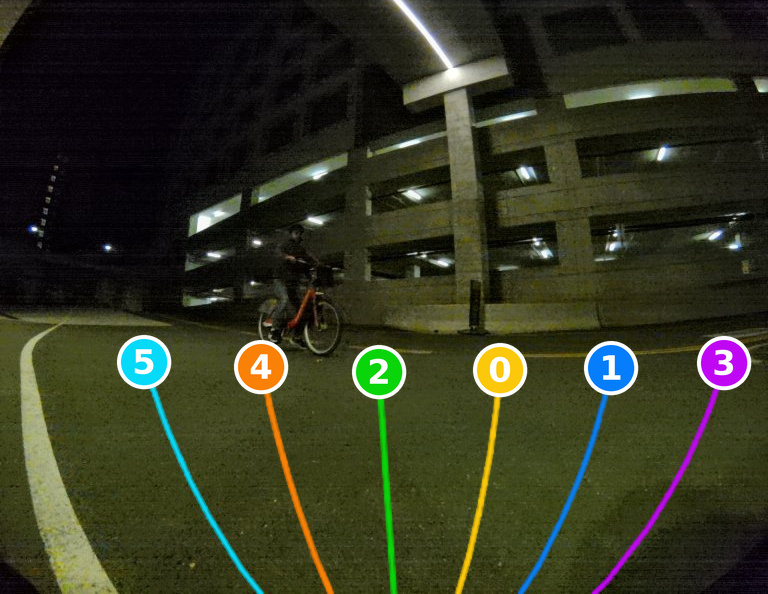} &
    \textbf{MOSU:} [1, 0, 3, 2, 4, 5], "Stop", "A cyclist is crossing in front of the robot from the left, about 6–8 m away. The goal is 21° right...."\par\vspace{3pt}
    \textbf{MOSU2: Stage 1:} True, "Item: cyclist (a fast, hard-to-predict road user). Needed: position and velocity, to see where the cyclist will be when the robot would cross."\par\vspace{3pt}
    \textbf{Stage 2:} Stop, "After the cyclist has passed, take path 0.", "at 1 m/s the cyclist reaches the robot's forward .... robot moving forward would meet the cyclist about 3 m ahead. The robot stops and lets the cyclist cross, which takes about 4–5 s. Path 0 is the direct route to the goal and is clear once the cyclist is past."&
    \includegraphics[height=\qualH]{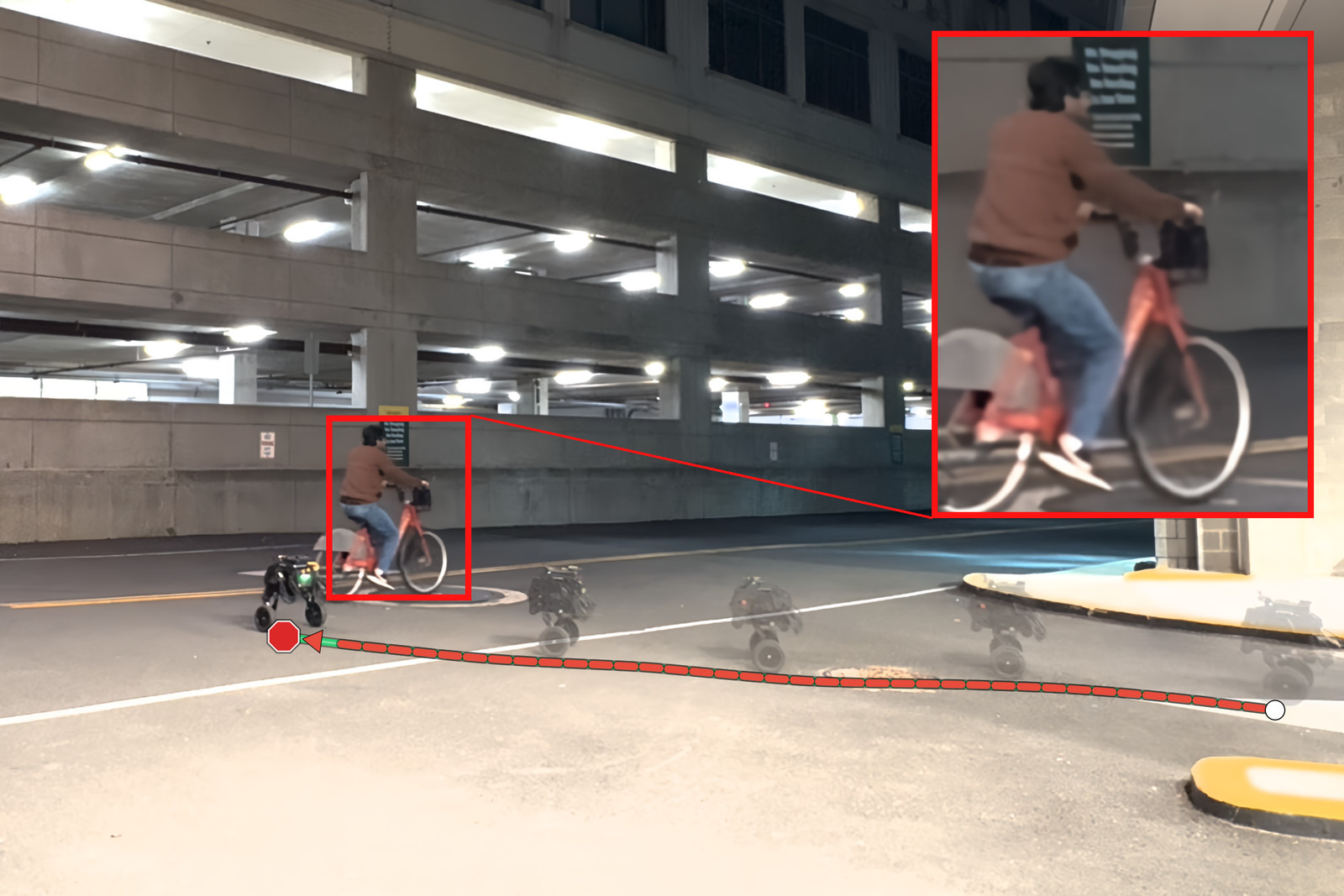} \\
    \addlinespace[3pt] 
    \midrule

    \includegraphics[height=\qualH]{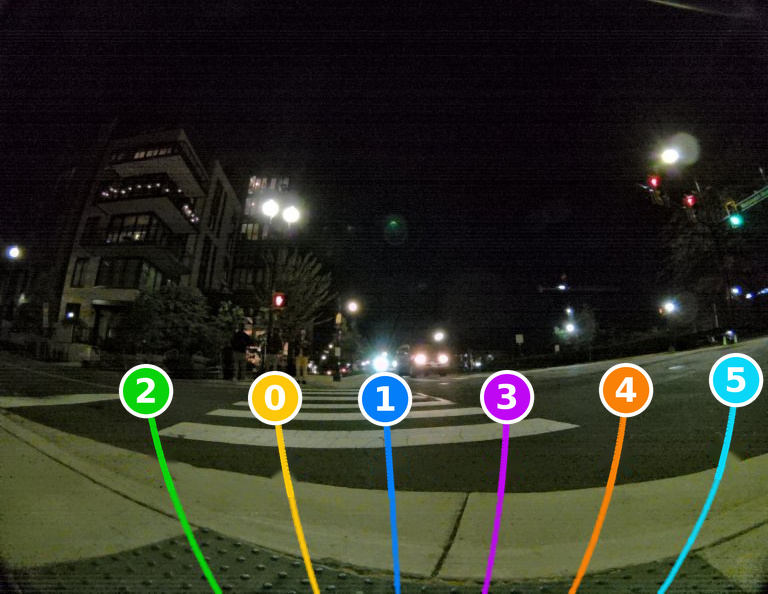} &
    \textbf{MOSU:} [0, 2, 1, 3, 4, 5], "Stop", "The robot is in front of a crosswalk.... Traffic lights ahead show red, and vehicles with bright headlights are further down the road ... The robot should stop until the light and the traffic allow crossing."\par\vspace{3pt}
    \textbf{MOSU2: Stage 1:} True, "Items: the traffic light (signal state), vehicles (motion) and pedestrians. Needed: which light applies to the robot, and whether the vehicles are moving."\par\vspace{3pt}
    \textbf{Stage 2 Decision:} Slow, 0, "the pedestrian light at (8, 1) is on, so the robot has the right of way. The vehicles are stopped about 11 m away on the right, and the second light at (5, -10) is on their side of the road.... Slow speed is used because it is dark, the headlights are bright, and a vehicle or pedestrian could still start moving." &
    \includegraphics[height=\qualH]{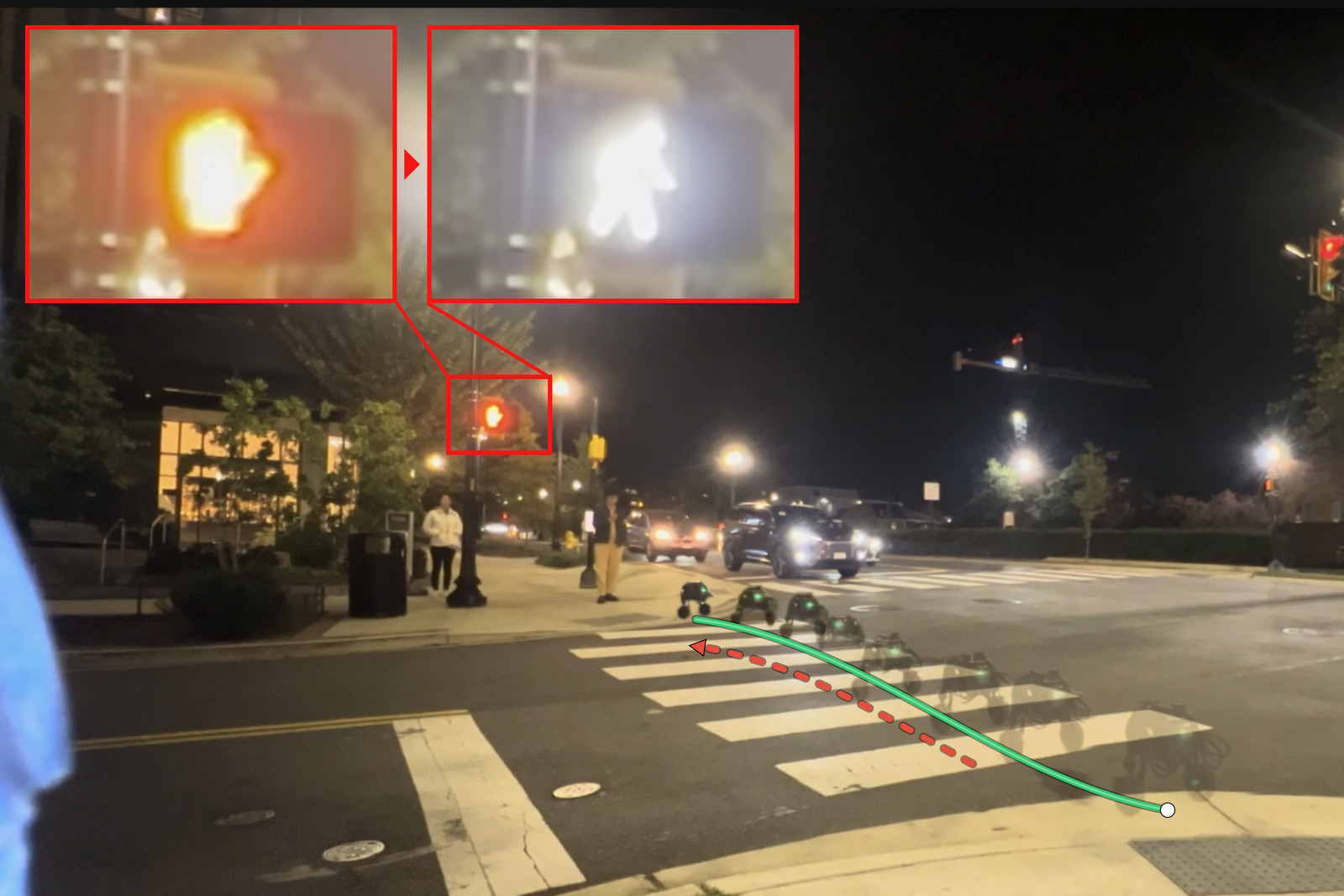} \\
    \addlinespace[3pt] 
    \midrule

    \includegraphics[height=\qualH]{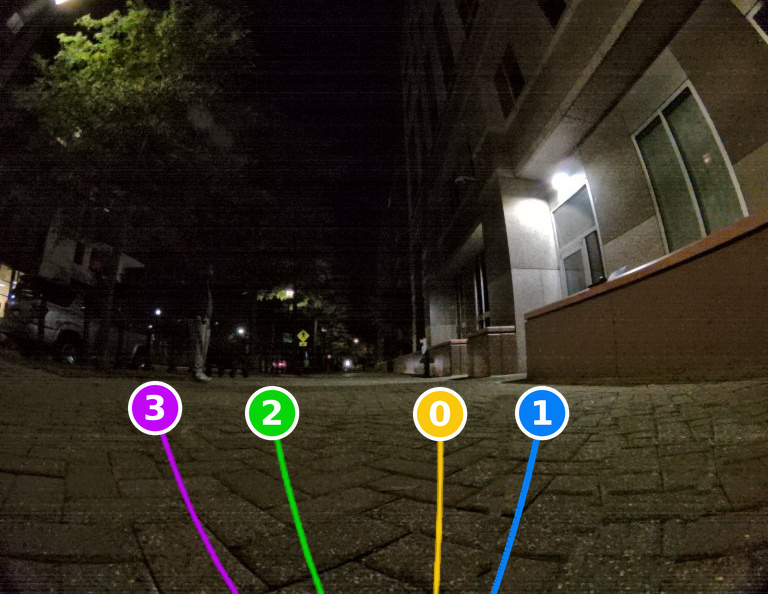} &
    \textbf{MOSU:} [2, 0, 1, 3], "Normal" "This is a brick-paved sidewalk at night with a building..."\par\vspace{3pt}
    \textbf{MOSU2: Stage 1:} False, Normal, 0, "Path 0 goes straight to the goal and keeps the most distance from the wall on the right and the left edge. Normal speed is fine because the route is clear. I would still watch the brick surface in the dark."  &
    \includegraphics[height=\qualH]{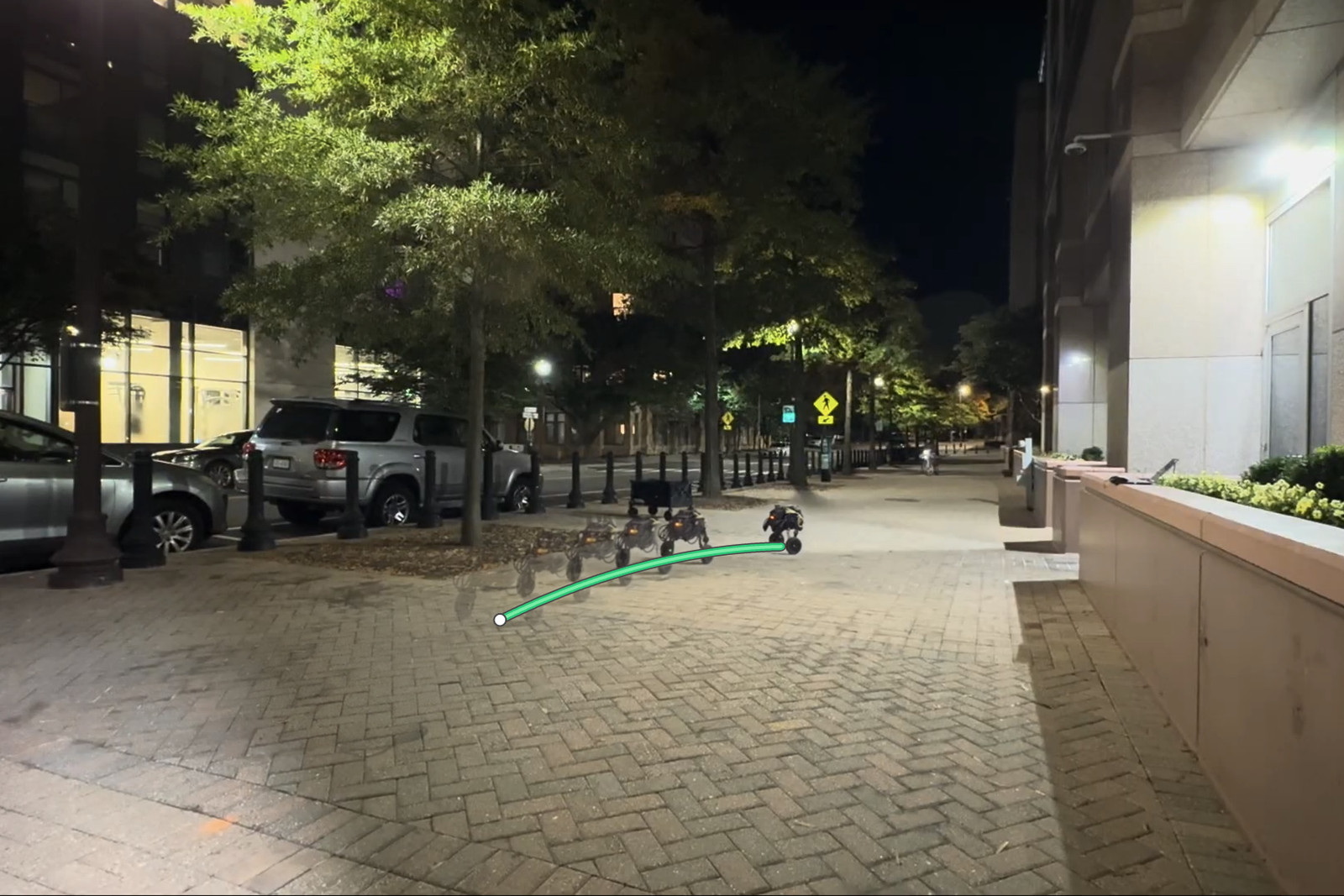} \\
    \bottomrule
  \end{tabularx}
  \caption{\textbf{Qualitative evaluation of MOSU2 across seven representative interactions from the five scenario groups}: Left column displays the onboard egocentric camera observation (First-Person View). Middle column shows the textual reasoning output from the two-stage reasoning, including screening decisions, map retrieval, and chosen motion. Right column shows the external observer view (Third-Person View) demonstrating physical execution.}
  \label{fig:three_column_qualitative}
\end{figure*}

First, we qualitatively show the construction of the geometric and dynamic maps in Fig.~\ref{fig:local_map_scenarios}. While the robot is moving, it maintains the geometric map in the form of an occupancy map. In the second stage of VLM reasoning, the requested semantic items are segmented in the RGB image and projected onto the geometric map to obtain the position and shape of each queried item. Given the input point clouds and RGB images, the middle column of Fig.~\ref{fig:local_map_scenarios} shows the geometric map, in which the orange items are those queried by the second-stage VLM, and the light green and light gray areas are free space and other occupied cells, respectively. The right column shows the dynamic map, in which we track the objects and store the velocities of the occupied cells. 

We then compare MOSU2 with MOSU to evaluate navigation decisions under social, traffic, and irregular environmental conditions. As shown in Fig.~\ref{fig:social_traffic_challenge}, the evaluation covers five scenario groups: human gestures; construction/hazard sites; signalized crosswalks with cars, pedestrians and bicycles; roads shared with vehicles, cyclists, or pedestrians; and sidewalk baselines without any traffic or social conditions. These scenarios examine complementary capabilities, including interpreting human gestures, following temporary restrictions, and coordinating robot motion with traffic signals and surrounding road users. MOSU performs contextual trajectory evaluation using a single RGB image and a single VLM query. MOSU2 screens the current observation with a VLM and invokes the second stage only when further investigation is requested. The second stage reasons over consecutive RGB observations and retrieves geometric and dynamic measurements from the local map according to the semantic items requested by the first stage. The evaluation addresses three questions: (Q1) whether MOSU2 improves navigation decisions in the scenarios; (Q2) whether it improves the recognition of conditions that require a specific response, and whether correct recognition leads to correct actions; and (Q3) how much reasoning latency and token usage it adds. Table~\ref{tab:mosu_mosu2_comparison} reports the navigation success rate (Nav. Success Rate), reasoning latency (Latency), condition-detection accuracy (Det. Accuracy), and token usage (Tokens) of both methods in each scenario group. Fig.~\ref{fig:three_column_qualitative} shows the first-person inputs to the VLMs, the VLM reasoning and outputs, and the third-person view of the robot-environment interactions corresponding to the scenarios. 

\paragraph{Metrics and protocol.}
A trial is successful when the robot completes the test without collision or human intervention and its behavior satisfies the requirement of the scenario: it stops in response to the stop gesture, passes the construction site on the side indicated by the signs, enters the crosswalk only when the pedestrian signal permits crossing, and yields to road users whose paths conflict with its own. Remaining stopped when proceeding is permitted and safe also counts as a failure. The navigation success rate is the fraction of successful trials in a scenario group. The condition-detection accuracy is computed over decision steps rather than trials: a step counts as correct when the output identifies the condition that characterizes the scenario group, such as the stop gesture, the construction site, the signal state, a conflicting road user, or the absence of any social or traffic condition that governs the robot at the sidewalk baseline. For MOSU, we evaluate the output of its single VLM query, and for MOSU2, the output of the last executed stage. For MOSU2, the first stage is performed at each step; when further investigation is requested, the second stage executes and retrieves the requested information to make the navigation decision. Because VLM inference runs in the cloud, the latency also includes the network round trip. We also report the input and output tokens per decision step. The token counts of MOSU and MOSU2 are both from the same VLM (ChatGPT-realtime). Because the evaluation consists of real-robot trials, the number of trials in each group is limited, so the group-level differences indicate trends rather than statistically established effects.

\paragraph{Navigation performance (Q1).}

MOSU2 improves navigation success in four of the five scenario groups and matches MOSU at the sidewalk baseline. The unweighted mean navigation success rate increases from 61.8\% for MOSU to 94.2\% for MOSU2, representing an absolute improvement of 32.4 percentage points. At signalized crosswalks and on roads shared with vehicles, cyclists, or pedestrians, MOSU2 achieves 100\% navigation success compared to 42\% and 56\% for MOSU, respectively. These dynamic settings require reasoning over traffic signal states and surrounding road-user trajectories, where MOSU's single-frame visual scoring frequently leads to overly conservative stops or premature entry. In construction site environments featuring temporary signage and barriers, MOSU2 increases navigation success from 50\% to 100\%, demonstrating robust path selection around physical restrictions. For human stop gestures, navigation success increases from 61\% to 71\%, reflecting improved responsiveness to social cues, though minor edge cases remain when gesture visibility is obscured. MOSU fails in this group mainly because its prompt does not specifically address waving or pausing gestures, so it tends to slowly steer around a person instead of stopping. At the sidewalk baseline, both methods achieve 100\% navigation success.

\paragraph{Condition detection and navigation decisions (Q2).}
The unweighted mean condition-detection accuracy rises from 72.2\% for MOSU to 91.4\% for MOSU2, an increase of 19.2 percentage points. The most pronounced improvement occurs at signalized crosswalks, where detection accuracy improves from 41\% to 93\%. As shown in the sixth row of Fig.~\ref{fig:three_column_qualitative}, MOSU detects car headlights and other traffic lights and is confused by them, whereas MOSU2 uses the retrieved position of the relevant traffic light to detect the traffic condition more accurately. Detection accuracy also increases from 50\% to 89\% in shared-road environments with moving vehicles, bicycles, and pedestrians. The third to fifth rows of Fig.~\ref{fig:three_column_qualitative} show that MOSU2 can estimate the velocities of surrounding objects; when a car is yielding to the robot, MOSU2 keeps the robot moving toward the goal. In contrast, MOSU cannot estimate velocity reliably, so it moves very conservatively. MOSU2 also detects stop gestures more accurately, improving accuracy from 70\% to 89\%. These gains indicate that MOSU2 recognizes the conditions that require a specific response more reliably than MOSU.

The construction site results highlight the crucial distinction between first-stage detection and second-stage analysis: both MOSU and MOSU2 achieve 100\% detection accuracy, yet MOSU2 doubles navigation success from 50\% to 100\%. This disparity demonstrates that correct scene classification is insufficient on its own; effective autonomy requires grounding detected conditions in measured geometry to understand the shape and semantics of the construction/hazard area. Conversely, at the sidewalk baseline, MOSU2's detection accuracy is 86\%, compared with 100\% for MOSU, because the first stage does not always recognize that no social or traffic condition governs the robot there; navigation success nevertheless remains at 100\% because the traversability analysis foundation is the same as MOSU. Thus, MOSU2 improves overall context recognition and decision reliability without requiring perfect condition classification in every single frame.

\paragraph{Reasoning cost (Q3).}
The performance gains of MOSU2 come at a higher reasoning cost. Context screening is performed at each decision step, with additional second-stage reasoning over recent RGB observations and local-map evidence when further investigation is requested. The unweighted mean decision latency increases from 1.72\,s for MOSU to 3.83\,s for MOSU2, approximately 2.2 times as long. The increase is concentrated in the scenario groups that invoke the second stage: at the construction site, the signalized crosswalk, and the shared road, MOSU2 takes 4.72--5.16\,s per decision, compared with 1.68--1.90\,s for MOSU. For the stop gesture and the sidewalk baseline, in which the first stage typically resolves the scene without invoking the second stage, the latency of MOSU2 (2.52\,s and 1.80\,s) remains much closer to that of MOSU (1.51\,s and 1.74\,s). MOSU2 also uses on average about 3,560 input and 310 output tokens per decision step, compared with 820 and 60 for MOSU, that is, about 4.3 and 5.3 times as many. The results therefore characterize a trade-off between higher navigation reliability and a higher reasoning cost per decision.

\section{Conclusion}
We presented MOSU2, an extension of MOSU for long-range outdoor navigation that combines route guidance, traversability assessment, and contextual reasoning. Building on MOSU's GPS-based routing and multi-modal trajectory generation and evaluation, MOSU2 introduces a VLM-retrievable local map that associates semantic regions with measured geometry and estimated motion. This information complements consecutive RGB observations when the second stage interprets social interactions, traffic conditions, and irregular situations. A two-stage reasoning architecture determines when additional investigation is needed: in the first stage, a VLM screens the current scene and supports routine subgoal-directed navigation, while the second stage retrieves relevant evidence and uses it to select a trajectory and velocity mode. Together, these components connect contextual understanding with environmental measurements and robot-specific traversability.

In real-world experiments across five scenario groups, MOSU2 achieved 100\% navigation success in four groups and 71\% in the stop-gesture group, whereas MOSU succeeded in 42--61\% of the trials in the four groups other than the sidewalk baseline. MOSU2 detected the relevant conditions with 86--100\% accuracy, at the cost of a higher reasoning latency when the second stage is invoked (4.7--5.2\,s, compared with 1.7--1.9\,s for MOSU) and about four times as many input tokens on average.

\section{Limitations and Future Work}

MOSU2 depends on the quality of its geometric, semantic, and dynamic observations. Occlusions, sparse LiDAR returns, segmentation errors, and calibration or odometry inaccuracies can affect the local map. In particular, associating segments by semantic category and spatial proximity can produce incorrect correspondences when several nearby objects share the same category. Changes in the visible portion of an object can also shift its observed centroid and introduce errors into the velocity estimate. Future work will investigate more robust temporal association and motion estimation, together with uncertainty estimates that allow the reasoning process to distinguish reliable measurements from incomplete or ambiguous evidence.

The conditional two-stage architecture depends on the first stage recognizing situations that require further investigation. A missed cue can prevent the second stage from being invoked, while unnecessary escalation increases reasoning cost. Even routine decisions require first-stage inference, and an escalated decision additionally requires map retrieval and second-stage inference. Conditional invocation therefore does not by itself establish a latency advantage over the single-query MOSU baseline. Future work will investigate uncertainty-aware escalation, lightweight screening models, and mechanisms for revisiting decisions as new observations arrive. Evaluation across a broader range of rare and unfamiliar situations will be necessary to characterize the trade-off between reasoning cost and missed contextual requirements.

Even when additional evidence is available, VLM interpretations can remain ambiguous or incorrect. Gestures, signal relevance, and vehicle behavior may admit several explanations, and motion measurements alone cannot establish another road user's intent. With reasoning latencies up to 5.16\,s (Table~\ref{tab:mosu_mosu2_comparison}), the scene can also change before a decision is executed. Future work will investigate explicit handling of conflicting evidence, checks on observation freshness, and independent verification of selected motions against updated geometric and dynamic observations.

\section{Funding}
The authors received no financial support for the research, authorship, and/or publication of this article.

\section{Declaration of conflicting interests}
The authors declared no potential conflicts of interest with respect
to the research, authorship, and/or publication of this article.

\FloatBarrier 



\balance 
\bibliographystyle{SageH}
\bibliography{references}

@inproceedings{gnd,
  title={Gnd: Global navigation dataset with multi-modal perception and multi-category traversability in outdoor campus environments},
  author={Liang, Jing and Das, Dibyendu and Song, Daeun and Shuvo, Md Nahid Hasan and Durrani, Mohammad and Taranath, Karthik and Penskiy, Ivan and Manocha, Dinesh and Xiao, Xuesu},
  booktitle={2025 IEEE International Conference on Robotics and Automation (ICRA)},
  pages={2383--2390},
  year={2025},
  organization={IEEE}
}

@inproceedings{gao2019global,
  title={A global path planning algorithm for robots using reinforcement learning},
  author={Gao, Penggang and Liu, Zihan and Wu, Zongkai and Wang, Donglin},
  booktitle={2019 IEEE International Conference on Robotics and Biomimetics (ROBIO)},
  pages={1693--1698},
  year={2019},
  organization={IEEE}
}

@article{chen2021adoption,
  title={The adoption of self-driving delivery robots in last mile logistics},
  author={Chen, Cheng and Demir, Emrah and Huang, Yuan and Qiu, Rongzu},
  journal={Transportation research part E: logistics and transportation review},
  volume={146},
  pages={102214},
  year={2021},
  publisher={Elsevier}
}

@article{davids2002urban,
  title={Urban search and rescue robots: from tragedy to technology},
  author={Davids, Angela},
  journal={IEEE Intelligent systems},
  volume={17},
  number={2},
  pages={81--83},
  year={2002},
  publisher={IEEE}
}

@article{gasparetto2015path,
  title={Path planning and trajectory planning algorithms: A general overview},
  author={Gasparetto, Alessandro and Boscariol, Paolo and Lanzutti, Albano and Vidoni, Renato},
  journal={Motion and Operation Planning of Robotic Systems: Background and Practical Approaches},
  pages={3--27},
  year={2015},
  publisher={Springer}
}

@article{ozturk2022review,
  title={A review of path planning algorithms in maritime autonomous surface ships: Navigation safety perspective},
  author={Ozturk, Ulku and Akdaug, Melih and Ayabakan, Tarik},
  journal={Ocean Engineering},
  volume={251},
  pages={111010},
  year={2022},
  publisher={Elsevier}
}

@inproceedings{van2015world,
  title={The world’s first GPS MOOC and worldwide laboratory using smartphones},
  author={Van Diggelen, Frank and Enge, Per},
  booktitle={Proceedings of the 28th international technical meeting of the satellite division of the institute of navigation (ION GNSS+ 2015)},
  pages={361--369},
  year={2015}
}

@article{kim2024learning,
  title={Learning Semantic Traversability with Egocentric Video and Automated Annotation Strategy},
  author={Kim, Yunho and Lee, Jeong Hyun and Lee, Choongin and Mun, Juhyeok and Youm, Donghoon and Park, Jeongsoo and Hwangbo, Jemin},
  journal={IEEE Robotics and Automation Letters},
  year={2024},
  publisher={IEEE}
}

@inproceedings{liang2024mtg,
  title={Mtg: Mapless trajectory generator with traversability coverage for outdoor navigation},
  author={Liang, Jing and Gao, Peng and Xiao, Xuesu and Sathyamoorthy, Adarsh Jagan and Elnoor, Mohamed and Lin, Ming C and Manocha, Dinesh},
  booktitle={2024 IEEE International Conference on Robotics and Automation (ICRA)},
  pages={2396--2402},
  year={2024},
  organization={IEEE}
}

@inproceedings{liang2024dtg,
  title={Dtg: Diffusion-based trajectory generation for mapless global navigation},
  author={Liang, Jing and Payandeh, Amirreza and Song, Daeun and Xiao, Xuesu and Manocha, Dinesh},
  booktitle={2024 IEEE/RSJ International Conference on Intelligent Robots and Systems (IROS)},
  pages={5340--5347},
  year={2024},
  organization={IEEE}
}

@ARTICLE{song2024vlm,
  author={Song, Daeun and Liang, Jing and Payandeh, Amirreza and Raj, Amir Hossain and Xiao, Xuesu and Manocha, Dinesh},
  journal={IEEE Robotics and Automation Letters}, 
  title={VLM-Social-Nav: Socially Aware Robot Navigation Through Scoring Using Vision-Language Models}, 
  year={2025},
  volume={10},
  number={1},
  pages={508-515},
  doi={10.1109/LRA.2024.3511409}}

@ARTICLE{song2024tgs,
  author={Song, Daeun and Liang, Jing and Xiao, Xuesu and Manocha, Dinesh},
  journal={IEEE Robotics and Automation Letters}, 
  title={VL-TGS: Trajectory Generation and Selection Using Vision Language Models in Mapless Outdoor Environments}, 
  year={2025},
  volume={10},
  number={6},
  pages={5791-5798},
  doi={10.1109/LRA.2025.3559822}}

@article{kong2025autospatial,
  title={AutoSpatial: Visual-Language Reasoning for Social Robot Navigation through Efficient Spatial Reasoning Learning},
  author={Kong, Yangzhe and Song, Daeun and Liang, Jing and Manocha, Dinesh and Yao, Ziyu and Xiao, Xuesu},
  journal={arXiv preprint arXiv:2503.07557},
  year={2025}
}

@article{cvae,
  title={Learning structured output representation using deep conditional generative models},
  author={Sohn, Kihyuk and Lee, Honglak and Yan, Xinchen},
  journal={Advances in neural information processing systems},
  volume={28},
  year={2015}
}

@inproceedings{bresenham1998algorithm,
  author    = {Bresenham, Jack E.},
  title     = {Algorithm for computer control of a digital plotter},
  booktitle = {Seminal graphics: pioneering efforts that shaped the field},
  volume    = {1},
  pages     = {45--50},
  year      = {1998},
  publisher = {Association for Computing Machinery (ACM)},
  address   = {New York, NY, USA},
  doi       = {10.1145/280811.280913},
}

@article{guo2020deep,
  title={Deep learning for 3d point clouds: A survey},
  author={Guo, Yulan and Wang, Hanyun and Hu, Qingyong and Liu, Hao and Liu, Li and Bennamoun, Mohammed},
  journal={IEEE transactions on pattern analysis and machine intelligence},
  volume={43},
  number={12},
  pages={4338--4364},
  year={2020},
  publisher={IEEE}
}

@article{sohail2024advancing,
  title={Advancing 3D point cloud understanding through deep transfer learning: A comprehensive survey},
  author={Sohail, Shahab Saquib and Himeur, Yassine and Kheddar, Hamza and Amira, Abbes and Fadli, Fodil and Atalla, Shadi and Copiaco, Abigail and Mansoor, Wathiq},
  journal={Information Fusion},
  pages={102601},
  year={2024},
  publisher={Elsevier}
}

@inproceedings{sridhar2024nomad,
  title={Nomad: Goal masked diffusion policies for navigation and exploration},
  author={Sridhar, Ajay and Shah, Dhruv and Glossop, Catherine and Levine, Sergey},
  booktitle={2024 IEEE International Conference on Robotics and Automation (ICRA)},
  pages={63--70},
  year={2024},
  organization={IEEE}
}

@article{shah2023vint,
  title={ViNT: A foundation model for visual navigation},
  author={Shah, Dhruv and Sridhar, Ajay and Dashora, Nitish and Stachowicz, Kyle and Black, Kevin and Hirose, Noriaki and Levine, Sergey},
  journal={arXiv preprint arXiv:2306.14846},
  year={2023}
}

@article{nasiriany2024pivot,
  title={Pivot: Iterative visual prompting elicits actionable knowledge for vlms},
  author={Nasiriany, Soroush and Xia, Fei and Yu, Wenhao and Xiao, Ted and Liang, Jacky and Dasgupta, Ishita and Xie, Annie and Driess, Danny and Wahid, Ayzaan and Xu, Zhuo and others},
  journal={arXiv preprint arXiv:2402.07872},
  year={2024}
}

@misc{GeminiTeam2024,
  author = {Gemini Team},
  title = {Gemini: A Family of Highly Capable Multimodal Models},
  year = {2024},
  eprint = {2312.11805},
  archivePrefix = {arXiv},
  primaryClass = {cs.CL},
  url = {https://arxiv.org/abs/2312.11805},
}

@article{li2018pointcnn,
  title={Pointcnn: Convolution on x-transformed points},
  author={Li, Yangyan and Bu, Rui and Sun, Mingchao and Wu, Wei and Di, Xinhan and Chen, Baoquan},
  journal={Advances in neural information processing systems},
  volume={31},
  year={2018}
}

@article{mavrogiannis2023core,
  title={Core challenges of social robot navigation: A survey},
  author={Mavrogiannis, Christoforos and Baldini, Francesca and Wang, Allan and Zhao, Dapeng and Trautman, Pete and Steinfeld, Aaron and Oh, Jean},
  journal={ACM Transactions on Human-Robot Interaction},
  volume={12},
  number={3},
  pages={1--39},
  year={2023},
  publisher={ACM, New York, NY}
}

@phdthesis{liang2025robot,
  title={Robot Navigation in Complex Scenarios},
  author={Liang, Jing},
  year={2025},
  school={University of Maryland, College Park}
}

@article{gul2019comprehensive,
  title={A comprehensive study for robot navigation techniques},
  author={Gul, Faiza and Rahiman, Wan and Nazli Alhady, Syed Sahal},
  journal={Cogent Engineering},
  volume={6},
  number={1},
  pages={1632046},
  year={2019},
  publisher={Taylor \& Francis}
}

@article{desouza2002vision,
  title={Vision for mobile robot navigation: A survey},
  author={DeSouza, Guilherme N and Kak, Avinash C},
  journal={IEEE transactions on pattern analysis and machine intelligence},
  volume={24},
  number={2},
  pages={237--267},
  year={2002},
  publisher={IEEE}
}

@article{alverhed2024autonomous,
  title={Autonomous last-mile delivery robots: a literature review},
  author={Alverhed, Elin and Hellgren, Simon and Isaksson, Hanna and Olsson, Lisa and Palmqvist, Hanna and Flod{e}n, Jonas},
  journal={European Transport Research Review},
  volume={16},
  number={1},
  pages={4},
  year={2024},
  publisher={Springer}
}

@article{zemke2020build,
  title={How to build a better robot... for quick-service restaurants},
  author={Zemke, Dina Marie V and Tang, Jason and Raab, Carola and Kim, Jungsun},
  journal={Journal of Hospitality \& Tourism Research},
  volume={44},
  number={8},
  pages={1235--1269},
  year={2020},
  publisher={SAGE Publications Sage CA: Los Angeles, CA}
}

@incollection{berezina2019robots,
  title={Robots, artificial intelligence, and service automation in restaurants},
  author={Berezina, Katerina and Ciftci, Olena and Cobanoglu, Cihan},
  booktitle={Robots, artificial intelligence, and service automation in travel, tourism and hospitality},
  pages={185--219},
  year={2019},
  publisher={Emerald Publishing Limited}
}

@article{kalita2020exploration,
  title={Exploration of extreme environments with currentand emerging robot systems},
  author={Kalita, Himangshu and Thangavelautham, Jekan},
  journal={Current Robotics Reports},
  volume={1},
  number={3},
  pages={97--104},
  year={2020},
  publisher={Springer}
}

@inproceedings{orca,
  title={Reciprocal collision avoidance with acceleration-velocity obstacles},
  author={Van Den Berg, Jur and Snape, Jamie and Guy, Stephen J and Manocha, Dinesh},
  booktitle={2011 IEEE International Conference on Robotics and Automation},
  pages={3475--3482},
  year={2011},
  organization={IEEE}
}

@article{graspe,
  title={Graspe: Graph based multimodal fusion for robot navigation in outdoor environments},
  author={Weerakoon, Kasun and Sathyamoorthy, Adarsh Jagan and Liang, Jing and Guan, Tianrui and Patel, Utsav and Manocha, Dinesh},
  journal={IEEE Robotics and Automation Letters},
  volume={8},
  number={12},
  pages={8090--8097},
  year={2023},
  publisher={IEEE}
}

@inproceedings{terrapn,
  title={Terrapn: Unstructured terrain navigation using online self-supervised learning},
  author={Sathyamoorthy, Adarsh Jagan and Weerakoon, Kasun and Guan, Tianrui and Liang, Jing and Manocha, Dinesh},
  booktitle={2022 IEEE/RSJ International Conference on Intelligent Robots and Systems (IROS)},
  pages={7197--7204},
  year={2022},
  organization={IEEE}
}

@article{wijayathunga2023challenges,
  title={Challenges and solutions for autonomous ground robot scene understanding and navigation in unstructured outdoor environments: A review},
  author={Wijayathunga, Liyana and Rassau, Alexander and Chai, Douglas},
  journal={Applied Sciences},
  volume={13},
  number={17},
  pages={9877},
  year={2023},
  publisher={MDPI}
}

@article{eder2023traversability,
  title={Traversability analysis for off-road environments using locomotion experiments and earth observation data},
  author={Eder, Matthias and Prinz, Raphael and Sch{\"o}ggl, Florian and Steinbauer-Wagner, Gerald},
  journal={Robotics and Autonomous Systems},
  volume={168},
  pages={104494},
  year={2023},
  publisher={Elsevier}
}

@inproceedings{wellhausen2021rough,
  title={Rough terrain navigation for legged robots using reachability planning and template learning},
  author={Wellhausen, Lorenz and Hutter, Marco},
  booktitle={2021 IEEE/RSJ International Conference on Intelligent Robots and Systems (IROS)},
  pages={6914--6921},
  year={2021},
  organization={IEEE}
}

@inproceedings{liang2025mosu,
  title={Mosu: Autonomous long-range robot navigation with multi-modal scene understanding},
  author={Liang, Jing and Weerakoon, Kasun and Song, Daeun and Kirubaharan, Senthurbavan and Xiao, Xuesu and Manocha, Dinesh},
  booktitle={International Symposium on Experimental Robotics},
  year={2026},
  organization={Springer}
}

@inproceedings{payandeh2025social,
  title={Social-LLaVA: Enhancing social robot navigation through human-language reasoning},
  author={Payandeh, Amirreza and Song, Daeun and Nazeri, Mohammad and Liang, Jing and Mukherjee, Praneel and Raj, Amir Hossain and Kong, Yangzhe and Manocha, Dinesh and Xiao, Xuesu},
  booktitle={2025 IEEE/RSJ International Conference on Intelligent Robots and Systems (IROS)},
  pages={17192--17198},
  year={2025},
  organization={IEEE}
}

@article{elevationmap,
    author = {Fankhauser, P{\'{e}}ter and Bloesch, Michael and Hutter, Marco},
    doi = {10.1109/LRA.2018.2849506},
    title = {Probabilistic Terrain Mapping for Mobile Robots with Uncertain Localization},
    journal = {IEEE Robotics and Automation Letters (RA-L)},
    volume = {3},
    number = {4},
    pages = {3019--3026},
    year = {2018}
  }

@inproceedings{liang2021crowd,
  title={Crowd-steer: Realtime smooth and collision-free robot navigation in densely crowded scenarios trained using high-fidelity simulation},
  author={Liang, Jing and Patel, Utsav and Sathyamoorthy, Adarsh Jagan and Manocha, Dinesh},
  booktitle={Proceedings of the twenty-ninth international conference on international joint conferences on artificial intelligence},
  pages={4221--4228},
  year={2021}
}

@inproceedings{weerakoon2025behav,
  title={Behav: Behavioral rule guided autonomy using vlms for robot navigation in outdoor scenes},
  author={Weerakoon, Kasun and Elnoor, Mohamed and Seneviratne, Gershom and Rajagopal, Vignesh and Arul, Senthil Hariharan and Liang, Jing and Jaffar, Mohamed Khalid M and Manocha, Dinesh},
  booktitle={2025 IEEE International Conference on Robotics and Automation (ICRA)},
  pages={7044--7051},
  year={2025},
  organization={IEEE}
}

@article{fang2026obstacles,
  title={From obstacles to etiquette: Robot social navigation with vlm-informed path selection},
  author={Fang, Zilin and Xiao, Anxing and Hsu, David and Lee, Gim Hee},
  journal={IEEE Robotics and Automation Letters},
  year={2026},
  publisher={IEEE}
}

@inproceedings{carion2026sam,
  title={Sam 3: Segment anything with concepts},
  author={Carion, Nicolas and Gustafson, Laura and Hu, Yuan-Ting and Debnath, Shoubhik and Hu, Ronghang and Suris Coll-Vinent, Didac and Ryali, Chaitanya and Alwala, Kalyan Vasudev and Khedr, Haitham and Huang, Andrew and others},
  booktitle={International conference on learning representations},
  volume={2026},
  pages={138846--138923},
  year={2026}
}

@article{wu2023brief,
  title={A brief overview of ChatGPT: The history, status quo and potential future development},
  author={Wu, Tianyu and He, Shizhu and Liu, Jingping and Sun, Siqi and Liu, Kang and Han, Qing-Long and Tang, Yang},
  journal={IEEE/CAA Journal of Automatica Sinica},
  volume={10},
  number={5},
  pages={1122--1136},
  year={2023},
  publisher={IEEE}
}

@inproceedings{liu2025citywalker,
  title={Citywalker: Learning embodied urban navigation from web-scale videos},
  author={Liu, Xinhao and Li, Jintong and Jiang, Yicheng and Sujay, Niranjan and Yang, Zhicheng and Zhang, Juexiao and Abanes, John and Zhang, Jing and Feng, Chen},
  booktitle={2025 IEEE/CVF Conference on Computer Vision and Pattern Recognition (CVPR)},
  pages={6875--6885},
  year={2025},
  organization={IEEE}
}

@article{schmittle2025lrn,
  author    = {Schmittle, Matthew and Baijal, Rohan and Hatch, Nathan and Scalise, Rosario and Guaman Castro, Mateo and Talia, Sidharth and Khetarpal, Khimya and Srinivasa, Siddhartha and Boots, Byron},
  title     = {Long Range Navigator (LRN) : Extending robot planning horizons beyond metric maps},
  journal   = {arXiv},
  year      = {2025},
}

@inproceedings{roth2024viplanner,
  title={Viplanner: Visual semantic imperative learning for local navigation},
  author={Roth, Pascal and Nubert, Julian and Yang, Fan and Mittal, Mayank and Hutter, Marco},
  booktitle={2024 IEEE International Conference on Robotics and Automation (ICRA)},
  pages={5243--5249},
  year={2024},
  organization={IEEE}
}

@inproceedings{chen2023open,
  title={Open-vocabulary queryable scene representations for real world planning},
  author={Chen, Boyuan and Xia, Fei and Ichter, Brian and Rao, Kanishka and Gopalakrishnan, Keerthana and Ryoo, Michael S and Stone, Austin and Kappler, Daniel},
  booktitle={2023 IEEE International Conference on Robotics and Automation (ICRA)},
  pages={11509--11522},
  year={2023},
  organization={IEEE}
}

@inproceedings{
        rana2023sayplan,
        title={SayPlan: Grounding Large Language Models using 3D Scene Graphs for Scalable Task Planning},
        author={Krishan Rana and Jesse Haviland and Sourav Garg and Jad Abou-Chakra and Ian Reid and Niko Suenderhauf},
        booktitle={7th Annual Conference on Robot Learning},
        year={2023},
        url={https://openreview.net/forum?id=wMpOMO0Ss7a}
      }

@article{zhang2026supernav,
  title={SuperNav: An Agentic Navigation System for Any Task in Any Scene},
  author={Zhang, Jinkai and Xu, Jingyi and Yu, Yuanhong and Guo, Jiarui and Hu, Ruizhen and Bao, Hujun and Zhou, Xiaowei and Peng, Sida},
  journal={arXiv preprint arXiv:2610.12126},
  year={2026}
}

@inproceedings{wei2026dualvln,
  title={Ground slow, move fast: A dual-system foundation model for generalizable vision-language navigation},
  author={Wei, Meng and Wan, Chenyang and Peng, Peng and Yu, Xiqian and Yang, Yuqiang and Feng, Delin and Cai, Wenzhe and Zhu, Chenming and Wang, Tai and Pang, Jiangmiao and others},
  booktitle={International Conference on Learning Representations},
  volume={2026},
  pages={12380--12396},
  year={2026}
}

@article{zhang2026walk,
  title={Walk with me: Long-horizon social navigation for human-centric outdoor assistance},
  author={Zhang, Lingfeng and Hao, Xiaoshuai and Bu, Xizhou and Tang, Yingbo and Li, Hongsheng and Lu, Jinghui and Wei, Xiu-shen and Ma, Jiayi and Liu, Yu and Zhang, Jing and others},
  journal={arXiv preprint arXiv:2604.26839},
  year={2026}
}

@article{wang2026hydranav,
  title={Hydra-Nav: Object Navigation via Adaptive Dual-Process Reasoning},
  author={Wang, Zixuan and Fang, Huang and Wang, Shaoan and Luo, Yuanfei and Dong, Heng and Li, Wei and Gan, Yiming},
  year={2026},
  journal={arXiv pre-print},
  url={https://arxiv.org/abs/2602.09972}
}

\end{document}